\documentclass[journal]{new-aiaa}
\usepackage[utf8]{inputenc}
\usepackage{float}
\usepackage{graphicx}
\usepackage{subcaption}
\usepackage{amsmath}
\usepackage[version=4]{mhchem}
\usepackage{siunitx}
\usepackage{booktabs}
\usepackage{amsmath}
\usepackage{longtable,tabularx}
\title{Deep Learning–Based Multi-Level Classification for Aviation Safety}                                                                                                                                                                                                                                                                                                                     
\author{Elaheh Sabziyan Varnousfaderani\footnote{Graduate Research Assistant, College of Aeronautics and Engineering, AIAA Student Member, esabziya@kent.edu}, Syed A. M. Shihab \footnote{Assistant Professor, College of Aeronautics and Engineering, AIAA Member, sshihab@kent.edu}, and Jonathan King\footnote{Undergraduate Research Assistant, College of Aeronautics and Engineering, jking130@kent.edu}}
\affil{Kent State University, Kent, OH, USA 44242}

\begin{document}

\maketitle

\begin{abstract}
Bird strikes pose a significant threat to aviation safety, often resulting in loss of life, severe aircraft damage, and substantial financial costs. Existing bird strike prevention strategies primarily rely on avian radar systems that detect and track birds in real time. A major limitation of these systems is their inability to identify bird species, an essential factor, as different species exhibit distinct flight behaviors, and altitudinal preference. To address this challenge, we propose an image-based bird classification framework using Convolutional Neural Networks (CNNs), designed to work with camera systems for autonomous visual detection. The CNN is designed to identify bird species and provide critical input to species-specific predictive models for accurate flight path prediction. In addition to species identification, we implemented dedicated CNN classifiers to estimate flock formation type and flock size. These characteristics provide valuable supplementary information for aviation safety. Specifically, flock type and size offer insights into collective flight behavior, and trajectory dispersion . Flock size directly relates to the potential impact severity, as the overall damage risk increases with the combined kinetic energy of multiple birds. We implemented and evaluated two classification architectures: the Cascade Classification Approach (CCA) and the Unified Classification Approach (UCA) for classifying bird species. CCA follows a three-stage process: it first classifies inputs as either bird or aircraft, then assigns bird instances to one of three size categories (small, medium, or large), and finally identifies the species using a size-specific classifier. In contrast to CCA, UCA uses a single CNN model to directly distinguish between aircraft and multiple bird species in one step. To benchmark our CNN-based framework, we compared its performance with three traditional machine learning classifiers: Support Vector Machine, Random Forest, and K-Nearest Neighbors. Experimental results demonstrate that the CNN consistently outperformed the baseline models across all evaluated scenarios. Furthermore, UCA achieved slightly higher overall accuracy than CCA. These findings highlight the potential of CNN-based classification to revolutionize proactive bird strike prevention. For flock type classification, we propose a two-stage CNN-based framework in which horizontal flock formations are first classified from bottom-view images and Column formations are further resolved using side-view images. We also implement a dedicated multi-class CNN to estimate flock size from visual data. Experimental results show that both the flock type and flock size classifiers achieve high accuracy, demonstrating the feasibility of extracting reliable group-level flight characteristics to support bird strike risk assessment.
\end{abstract}

\section{Introduction}
\subsection{Bird Strikes}

Bird strikes, collisions between aircraft and birds, have posed a significant threat to aviation safety since the beginning of aviation. These incidents can occur during various phases of flight, including takeoff, landing, and cruise, and are referred to as bird strikes \cite{BS4}. Bird strikes not only endanger the lives of passengers and crew onboard but also lead to substantial financial losses for the aviation, which cost millions of dollars in aircraft damages annually.

The impact of bird strikes has been well-documented throughout aviation history, with both nonfatal and fatal incidents recorded. The first nonfatal bird strike occurred in 1905 when Orville Wright’s Wright Flyer III struck a bird over a cornfield near Dayton, Ohio \cite{dolbeer2013history}. The first fatality due to a bird strike was recorded in 1912, when Calbraith Rodgers’s aircraft collided with a gull, leading to his death \cite{cleary2005wildlife}. More recent high-profile incidents underscore the continued risks posed by bird strikes. These include the emergency water landing of US Airways Flight 1549 in the Hudson River in 2009, following a collision with a flock of geese, and the fatal crash of a Dornier 228 in Kathmandu, Nepal, in 2012 after striking a black kite. Additionally, in 2019, Ural Airlines Flight 178 was forced to land in a cornfield near Moscow after ingesting a flock of gulls into its engines during takeoff \cite{faa2023wildlife}. In 2024, a Jeju Air Boeing 737-800 attempted an emergency landing at Muan International Airport in South Korea, but the effort was unsuccessful due to a suspected bird strike and poor weather conditions. Only two crew members survived the crash, with all other lives lost \cite{spirochkin2025design}.

\begin{figure}[H]
    \centering
        \includegraphics[width=15.5cm, height = 7 cm]{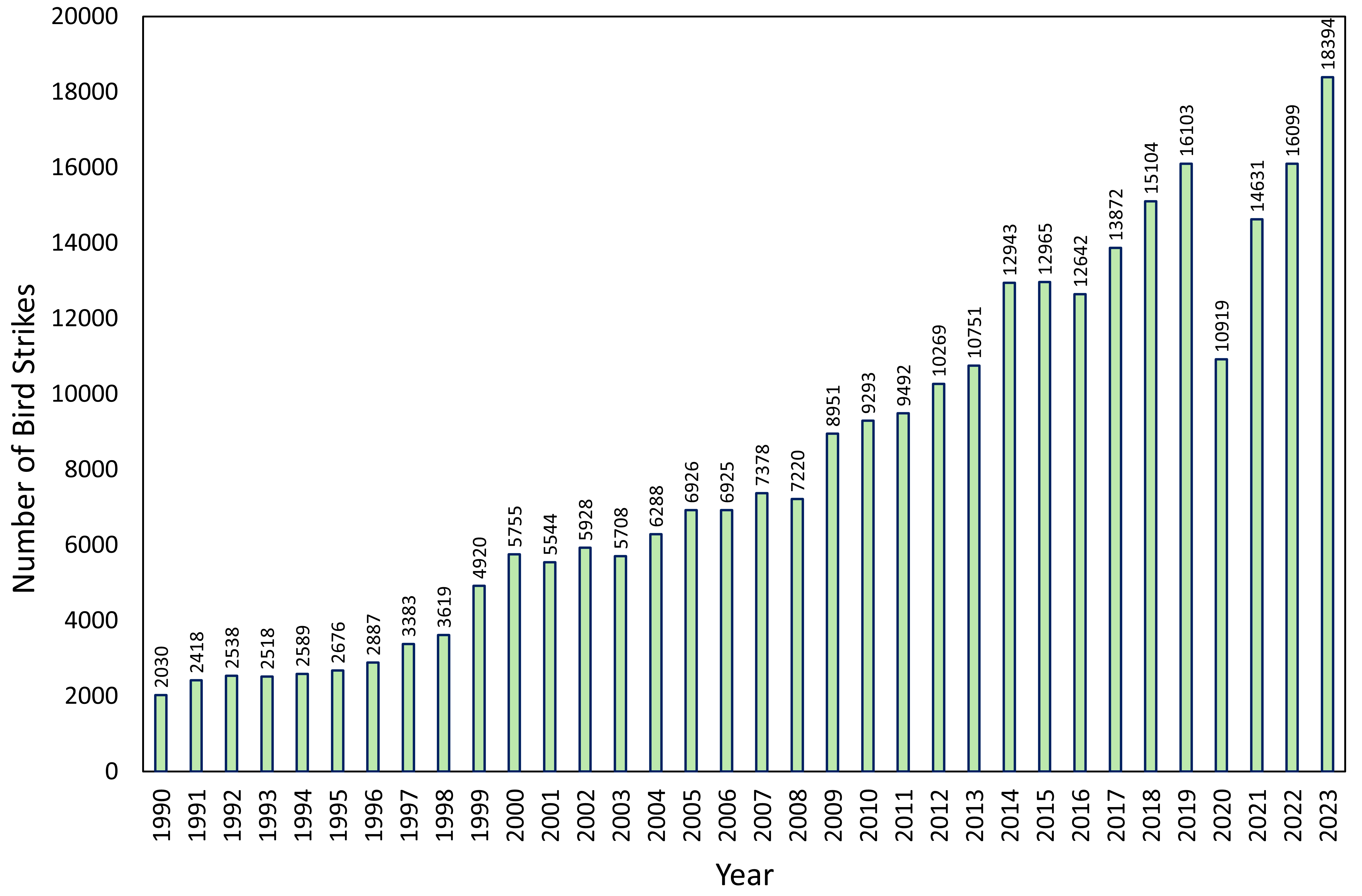}
        \caption{Total number of annual bird strikes with civil aircraft in the United States, 1990-2023 \cite{faa2023wildlife}}
        \label{Annual-Bird-Strikes}
\end{figure}

\begin{figure}[H]
    \centering
        \includegraphics[width=16.5cm, height = 8 cm]{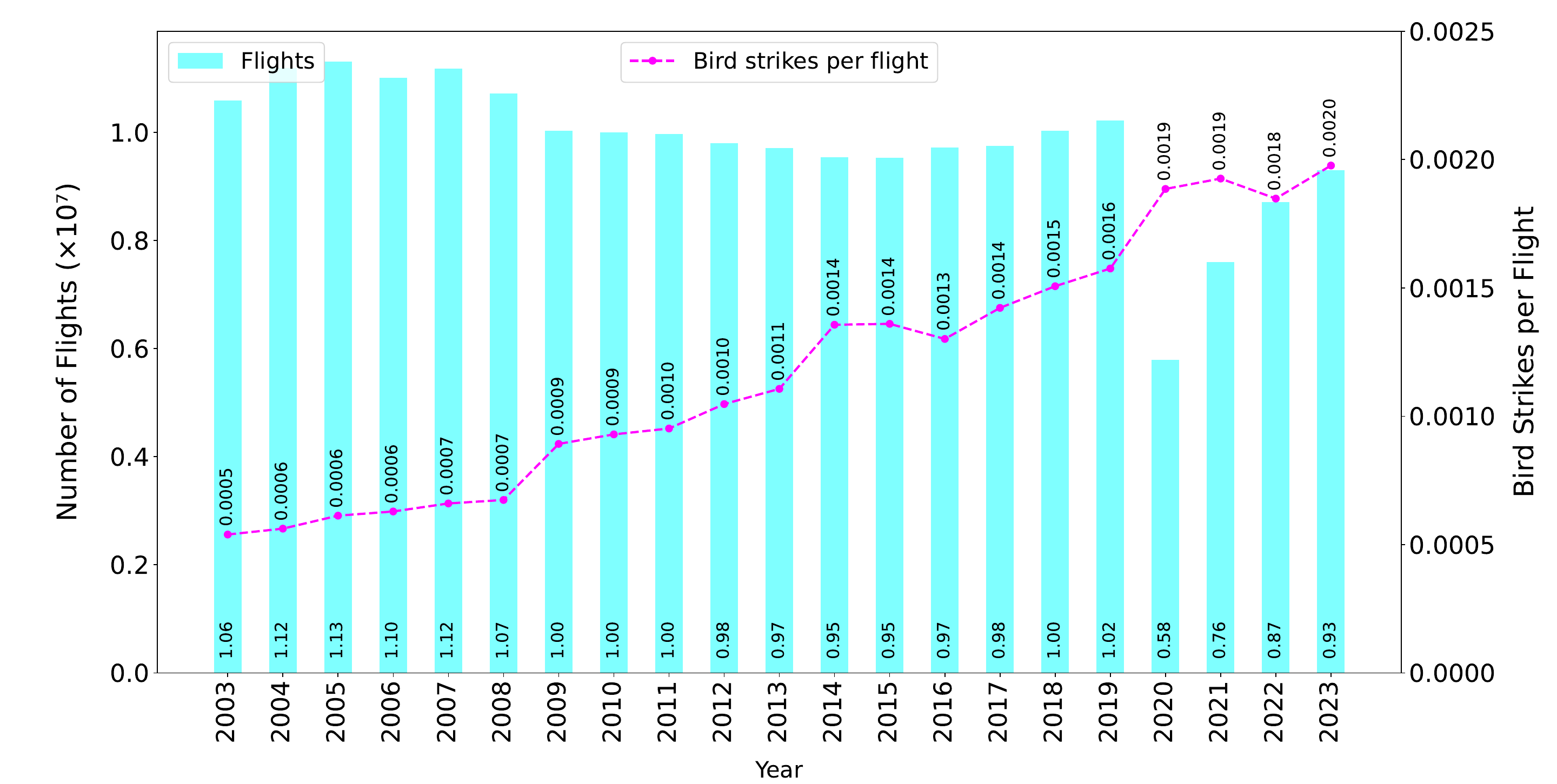}
        \caption{Annual number of commercial flights and bird strike rates (2003–2023) (Adapted from \cite{varnousfaderani2025bird})}
        \label{Rate}
\end{figure}

The frequency of bird strikes has shown a troubling upward trend over the years, with tens of thousands of cases reported annually in the United States alone \cite{faa2023wildlife}. This increase is attributed to factors such as rising bird populations, increased air traffic, quieter aircraft designs, and improved reporting systems. Data from the Federal Aviation Administration (FAA) wildlife strike database highlight this trend, shown in Fig. \ref{Annual-Bird-Strikes}, with a significant dip observed in 2020 due to reduced air traffic during the COVID-19 pandemic. Figure \ref{Rate}, however, shows that even though the total number of bird strikes dropped significantly during the COVID-19 pandemic, the number of bird strikes per flight actually increased. This underscores the importance of developing effective bird strike prevention measures.

\subsection{How Bird Strikes Would Be Actively Prevented?}

Monitoring and tracking bird movements within and around airport airspace are essential for understanding their flight patterns and actively preventing bird strikes. The role of advanced surveillance sensors in tracking and collecting precise, real-time data on bird movements to develop predictive models and implement mitigation systems effectively is vital. Among the various sensors, avian radars play a critical role in this process. Avian radars have been specifically designed to detect and monitor individual birds in localized areas, making them particularly effective around airports \cite{nohara2007overview, beason2013beware, gauthreaux2019detecting, phillips2018efficacy}. The incident involving flight 1549 in 2009, which resulted from a collision with a flock of geese, could potentially have been avoided if avian radar systems were installed in the airport’s airspace to detect the birds and provide vital data to pilots and ground control \cite{nohara2009could}. These systems are now recognized as essential tools in mitigating bird strikes and ensuring flight safety. Weather radars also provide valuable data, particularly in detecting large-scale movements of migratory birds \cite{wang2023quantification}. They can be connected with avian radar systems to form a cohesive network, and enhance overall detection capabilities \cite{shi2023prospects, schekler2023automatic}. However, avian radars excel at identifying and tracking individual birds within smaller, high-priority zones around airports \cite{chen2012flying, chen2022review}. Notable avian radar systems include Merlin radar (USA), Accipiter radar (Canada), and Robin radar (Netherlands) \cite{chen2022review, Merlin, Accipiter, Robin}.

Despite these advancements, avian radars have a notable limitation: they cannot identify bird species. This shortcoming restricts the development of species-specific predictive models that could enhance flight path predictions and mitigation measures. Addressing this limitation requires a large datasets of bird images and the development of an image-based classifier. Such a classifier would identify bird species accurately, and allow predictive models to tailor their forecasts based on the distinct behaviors of each species. To enhance aviation safety and reduce the impact of bird strikes, it is crucial to equip airports with advanced avian radar systems integrated with cameras. By combining these technologies, airports can significantly improve their bird strike prevention strategies, protecting both flights and wildlife.

\subsection{Outline of the Paper}
The remainder of this paper is organized as follows. Section \ref{Literature Review} begins with a review of the literature on bird strikes, followed by a discussion of existing classification methods. Section \ref{Problem Statement} defines the problem addressed in this study. Our proposed methodology is detailed in Section \ref{Solution}. Experimental results and their analysis are presented in Section \ref{Result}. Section \ref{Discussion} discusses key insights gained from the implementation. Finally, Section \ref{Conclusion} concludes the paper and highlights potential future work.

\section{Literature Review}
\label{Literature Review}
Bird strikes have been a focus of extensive research aimed at understanding the factors that affect the probability of bird strikes such as season, bird type, aircraft type, phases of flight, and environmental condition. Researchers have investigated high-risk bird species \cite{high-risk}, analyzed the conditions that increase bird strike probabilities \cite{metz2020bird, factor2, shao2020key, ning2014bird}, and assessed the economic damage caused by such incidents \cite{costs}. Studies have also utilized historical data to predict future strike occurrences and identify geographic hotspots prone to bird strikes \cite{smojver2011bird, dolbeer2021wildlife}. While these efforts have enhanced knowledge in this domain, they have largely centered on reactive analysis rather than proactive prevention.

Various strategies have been proposed to reduce the risk of bird strikes. Habitat modification near airports is a common approach, targeting the removal of bird-attracting elements such as food sources, water bodies, and vegetation that may serve as nesting sites \cite{L3}. Techniques to control bird behavior include the use of deterrents such as pyrotechnics, lasers, and long-range acoustic devices, as well as changing aircraft appearances through specific colors to discourage birds \cite{L1, L2}. However, these measures require periodic variation, as birds can become habituated to repeated techniques, reducing their effectiveness over time. Using multiple methods has been shown to improve outcomes \cite{harris1998evaluation, klug2023review}.

Another significant area of research has been the development of bird movement prediction models to enable aircraft and bird deconfliction. Using data from avian and weather radars, these models attempt to forecast bird flight paths and inform adjustments to aircraft operations. Linear regression models have been the very first predictive models, but they are reported to be inadequate for real-world application due to their low accuracy \cite{metz2021analysis, metz2021air}. Several existing preventive models primarily focus on ground-based deconfliction strategies, such as delaying departures to avoid predicted bird trajectories \cite{metz2021air, sabziyan2024safe}. While effective to some extent, these strategies carry economic consequences, such as financial costs associated with delayed operations, which underscores the need for more advanced and precise prediction methodologies. The next advanced model for predicting bird movement discussed in the literature is the Long Short-Term Memory (LSTM) network, implemented in \cite{sabziyan2023bird}, which demonstrated higher accuracy compared to the existing linear regression model. The predictive model is implemented using publicly available data on pigeons and is specifically designed for them. Predicting the movement of another bird species would require a separate model tailored to the characteristics of its flight. Incorporating tactical deconfliction models, commonly used for aircraft-aircraft collision avoidance, offers promising possibilities for reducing bird strike risks. These models optimize decision variables such as aircraft velocity, heading, and rate of climb to achieve objectives like minimizing fuel consumption, flight time, or deviations from planned routes \cite{fuel1, fuel2, time, zhao2021multiple, deviation1, deviation2}. To prevent bird strikes tactically, \cite{sabziyan2024collision} implemented an optimization-based aircraft-bird tactical deconfliction model which uses predicted flight path of the bird with 95\% confidence interval. All active preventive model assumed 1) the existing of the avian radar and real-time tracking, 2) birds are classified correctly.  Although the literature highlights various approaches to mitigating bird strikes, the assumption related to bird classification in the papers which developed active preventive models highly need to be relaxed. This is crucial, as each bird species has unique flight characteristics that significantly affect the accuracy and effectiveness of predictive models.

A classifier needs to be developed to specify the type of bird to enhance the accuracy of predictive models and the effectiveness of active preventive measures. Different bird species exhibit distinct flight behaviors, altitudinal preferences, and movement patterns, which directly influence the precision of bird movement predictions. By integrating species-specific data into prediction and deconfliction models, aircraft operations can be adjusted more effectively to prevent collisions. Several studies have attempted to classify bird species using various data modalities, such as videos, sounds, and images. Table \ref{tab:comparison} shows the features previous works have considered. Although video and sound-based classifiers have shown good results, they are not ideal for real-time bird species classification. Capturing a video takes time, even a short 2-second video introduces delay before processing can begin. Their success rates vary widely from 33\% to 97\%. The most accurate results, especially for sound-based methods, came from Support Vector Machines (SVM) and Artificial Neural Networks (ANN), reaching up to 97\% and 96.4\% accuracy. However, in real-world airport environments, sound-based detection is not reliable due to background noise. In addition, there is no guarantee that birds generate sounds all the time. Image-based classifiers work with images that can be captured and processed almost instantly, making them better suited for fast, real-time detection. Image-based classifiers have also been tested but often do not perform well. They also overlooked considering images of bird while flying. This is a major issue because identifying birds in flight is crucial to preventing bird strikes. Many classifiers also miss high-risk species that are most commonly involved in bird strikes, reducing applicability. Therefore, more research is needed to build practical classifiers that can accurately detect high-risk bird species.

\renewcommand{\arraystretch}{1}
\begin{table}[H]
\centering
\caption{Comparison of recently developed classifiers (Adapted from \cite{varnousfaderani2025bird})}
\begin{tabular}{ccccccccc}
\toprule
\textbf{Papers} & \textbf{Sound} & \textbf{Video} & \textbf{Image} & \textbf{Flight Track} & \textbf{Method} & \textbf{Accuracy} & \textbf{\# Birds} \\
\midrule
\cite{raghuram2016bird} & $\checkmark$ & $\times$ & $\times$ & $\times$ & RF & 83\% & 33\\
\cite{lucio2015bird} & $\checkmark$ & $\times$ & $\times$ & $\times$ & SVM & 77.65\% & 46\\
\cite{dharaniya2022bird} & $\checkmark$ & $\times$ & $\times$ & $\times$ & CNN & 70.8\% & 999\\
\cite{qiao2017bird} & $\times$ & $\times$ & $\checkmark$ & $\times$ & SVM+DT & 83.87\% & 15 \\
\cite{fagerlund2007bird} & $\checkmark$ & $\times$ & $\times$ & $\times$ & SVM & 97\% & 14 \\
\cite{pahuja2021sound} & $\checkmark$& $\times$ & $\times$ & $\times$ & ANN  & 96.1\% & 8 \\
\cite{lopes2011automatic} & $\checkmark$ & $\times$ & $\times$ & $\times$ & ANN & 96.4\% & 3 \\
\cite{yoshihashi2017bird} & $\times$ & $\times$ & $\checkmark$ & $\times$ & CNN & 87.5\% & 2  \\
\cite{marini2015visual} & $\checkmark$ & $\times$ & $\checkmark$ & $\times$ & SVM & 37.33\% & 50 \\
\cite{atanbori2018classification} & $\times$ & $\checkmark$ & $\times$ & $\times$ & CNN & 90\% & 10  \\
\cite{rai2019analysis} & $\times$ & $\times$ & $\checkmark$ & $\times$ & KNN & 33\% & 200 \\

\bottomrule
\end{tabular}

\label{tab:comparison}
\end{table}
\indent \indent \textit{Note: RF is Random Forest, DT is Decision Tree, KNN is K-Nearest Neighbors, CNN is Convolutional Neural Network, and ANN is Artificial Neural Network.}

\section{Problem Statement}
\label{Problem Statement}

\begin{figure}[H]
\centering
\includegraphics[width=15.5cm, height =6.5 cm]{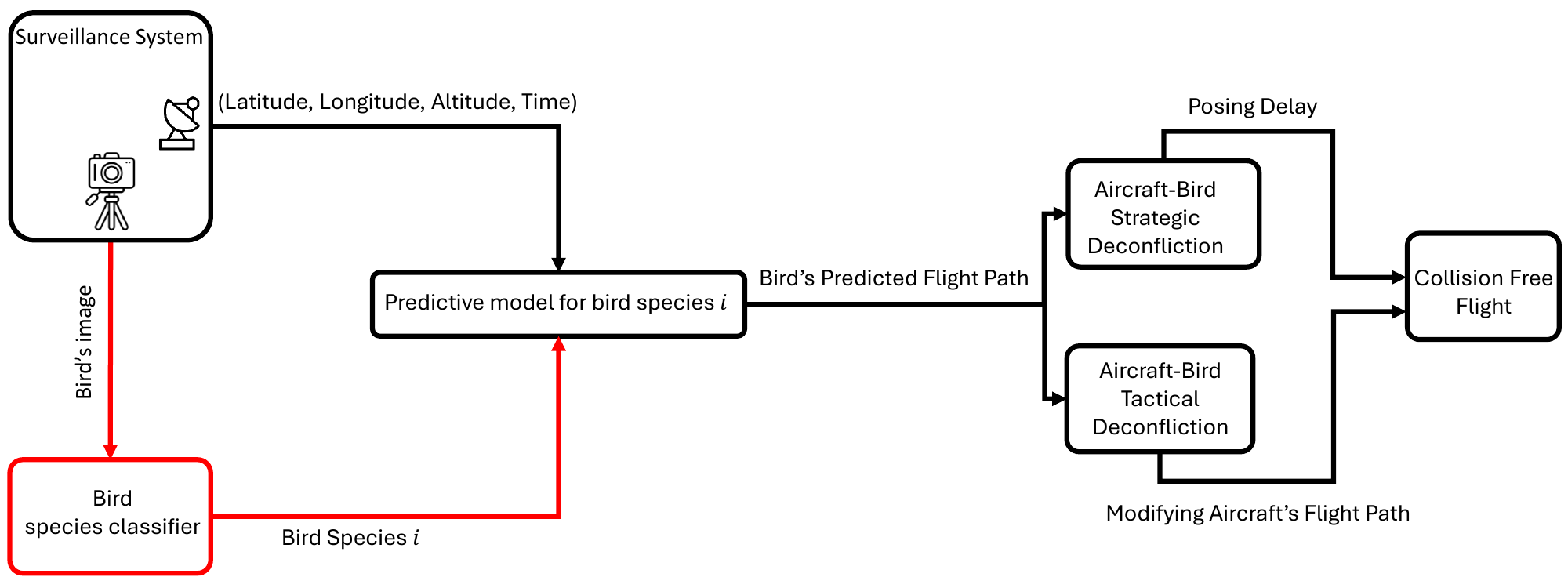}
\caption{Workflow for bird strike prevention using avian sensors (e.g., radars, cameras), a classifier and predictive models (adapted from \cite{sabziyan2025deep})}
\label{PS}
\end{figure}

Bird strikes are a significant concern for aviation safety, causing damage to aircraft, financial losses, and putting human lives at risk. Avian radars have been used near airports to track the movement of birds near airports in real time. Although avian radars provide valuable data on bird movements, including their latitude, longitude, altitude, and time of detection, they have a notable limitation: they do not provide information about the species of the birds being tracked. This is a critical gap, as the behavior of birds including their flight speed, altitude preferences, and migratory paths varies significantly between species. Effective bird strike prevention depends on the ability to predict bird movements accurately, as shown in Fig. \ref{PS}, which requires modeling flight paths specific to each species.

Bird species classification plays a pivotal role in bridging this gap. By classifying birds based on images taken by cameras, we can identify the species and consider their unique behavioral patterns to develop species-specific predictive models. These models are essential for forecasting future bird flight paths accurately. Once these predictive models are in place, the accurate predicted flight path can be utilized in strategic and tactical deconfliction systems. This paper primarily focuses on addressing the gap illustrated in Fig. \ref{PS}, showing with red color, by collecting data on the top 33 bird species frequently involved in bird strikes in the United States from 1990 to 2023 and selecting an appropriate model to classify these species. 

In addition to bird species classification, another important piece of information that can be extracted from visual sensing systems is flock type, and flock size classification when birds are flying in flocks. Identifying the flock type and flock size can provide deeper insight into their collective flight dynamics, including relative spacing, formation geometry, and coordinated motion patterns. This information can enhance the accuracy of flight path prediction models by accounting for inter-bird interactions that are not present in solo flight. Furthermore, in scenarios where bird strike prevention becomes infeasible, knowledge of flock structure and density can be leveraged to assess and mitigate the potential severity of impact on the aircraft, supporting more informed tactical and operational decision-making.

\section{Methodology}
\label{Solution}
As discussed in the preceding section, this paper addresses the limitation of avian radar systems in identifying specific bird species, a critical capability for predicting flight paths and preventing bird strikes. To tackle this problem, we propose two different solutions: (1) the \textbf{Cascade Classification Approach (CCA)} and (2) the \textbf{Unified Classification Approach (UCA)}.

\begin{figure}[H]
\centering
\includegraphics[width=16.5cm, height = 8.5 cm]{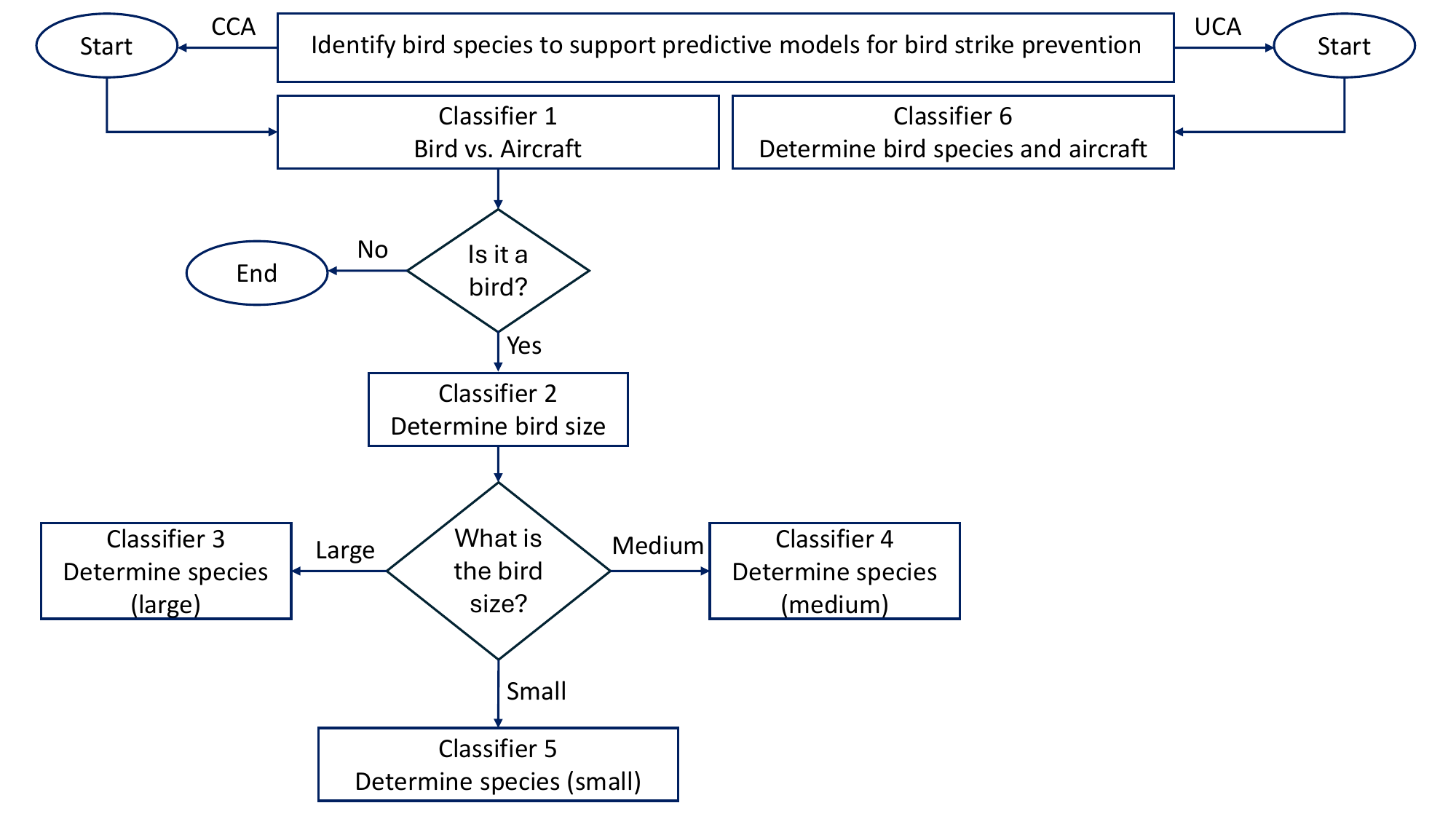}
\caption{Flowchart illustrating the two proposed approaches for bird species identification. The CCA decomposes the problem into three stages: bird detection, size classification, and species classification using size-specific models. The UCA uses a single classifier to simultaneously identify bird species or detect aircraft inputs.}
\label{Two-Solutions}
\end{figure}

In the CCA, as shown in Fig. \ref{Two-Solutions}, the identification task is decomposed into three sequential stages. \textbf{Stage 1} classifies input images as either bird or aircraft. \textbf{Stage 2} determines the size category of the bird (small, medium, or large). \textbf{Stage 3} applies a dedicated species-level classifier based on the identified size, using separate classifiers for each size group. This hierarchical structure aims to improve species-level classification by narrowing the search space according to size.

In contrast, the UCA employs a single classifier that directly categorizes an input image into one of the predefined bird species or the aircraft class. Figure \ref{Two-Solutions} illustrates the steps for the two proposed solutions. We evaluate both approaches to determine which offers better performance in terms of species identification accuracy and system efficiency for bird strike prevention applications. The first step in tackling this issue involves collecting images of aircraft and bird species frequently involved in bird strikes in the United States, captured during flight. Once images are collected for each group, they are used as input for classification models. A CNN, a deep learning (DL) algorithm, is employed as the main model for both CCA and UCA. CNNs are capable of processing input images, learning and assigning importance to various features through weights and biases, and effectively distinguishing between different bird species. The next step involves dividing the preprocessed images into three datasets: training, validation, and test sets. The training and validation sets are used to train the classifiers, with the validation set playing a critical role in preventing overfitting. The performance of the classifiers is then evaluated using the test set by calculating its accuracy. Additionally, we compared the CNN’s performance with three well-known machine learning algorithms: Support Vector Machine, Random Forest, and k-Nearest Neighbors.

In addition to bird species identification, this study also addresses the classification of flock type and flock size using dedicated CNN classifiers. These attributes provide complementary information that can further enhance bird flight path prediction and risk assessment for bird strike prevention.

\begin{figure}[H]
\centering
\includegraphics[width=16.5cm, height = 8 cm]{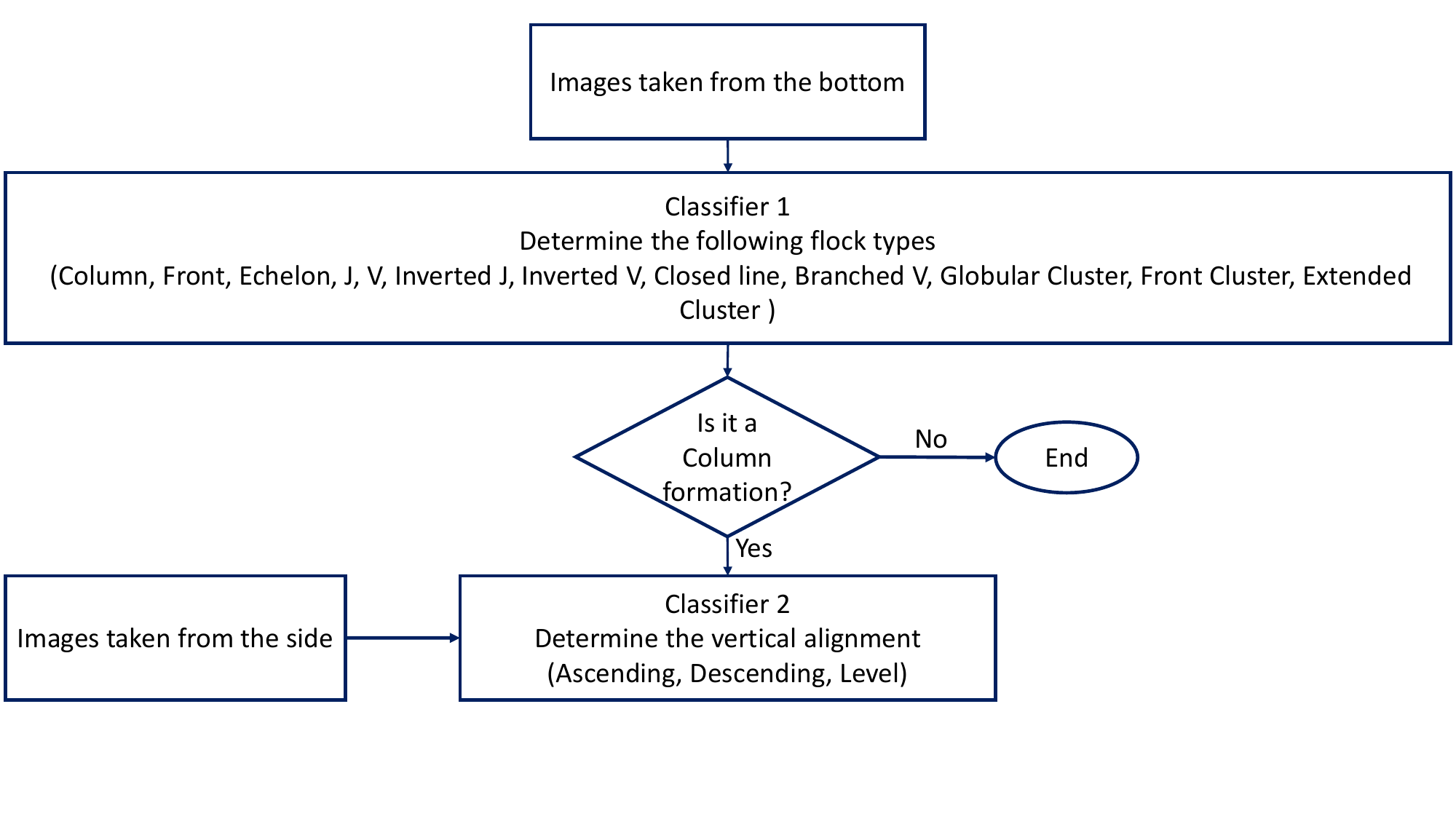}
\caption{Two-stage classification framework for flock type identification. Bottom-view images are first used to classify the horizontal flock formation. If a column formation is detected, side-view images are used to determine the vertical alignment (ascending, descending, or level).}
\label{Flock-Flowchart}
\end{figure}

For flock type classification, a cascade strategy is adopted, as illustrated in the flowchart shown in Fig.~\ref{Flock-Flowchart}. When the flock is observed from a bottom view, twelve distinct flock formations are considered, namely Column, Front, Echelon, J, V, Inverted J, Inverted V, Closed Line, Branched V, Globular Cluster, Front Cluster, and Extended Cluster, as reported in \cite{varnousfaderani2025bird}. A first classifier is trained to identify the flock formation based on bottom-view images. If the detected flock type is a Column formation, a second classifier is activated using side-view images to further determine the vertical alignment of the flock. Specifically, the Column formation is categorized as Ascending, Descending, or Level, reflecting the three possible vertical motion patterns \cite{varnousfaderani2025bird}. This two-stage classification approach enables more accurate characterization of flock geometry and motion.

For flock size classification, a separate multi-class classifier is employed. Five flock size categories are defined: 5--20 birds, 21--40 birds, 41--60 birds, 61--80 birds, and 81--100 birds. These ranges are selected under the assumption that the total kinetic energy and potential impact severity within each size class do not vary significantly, whereas transitions between classes correspond to meaningful changes in risk level. The classified flock size information can therefore be used to support both predictive modeling and impact mitigation strategies in bird strike scenarios.

\subsection{Convolutional Neural Networks}

\begin{figure}[H]
\centering
\includegraphics[width=12cm, height = 4.5 cm]{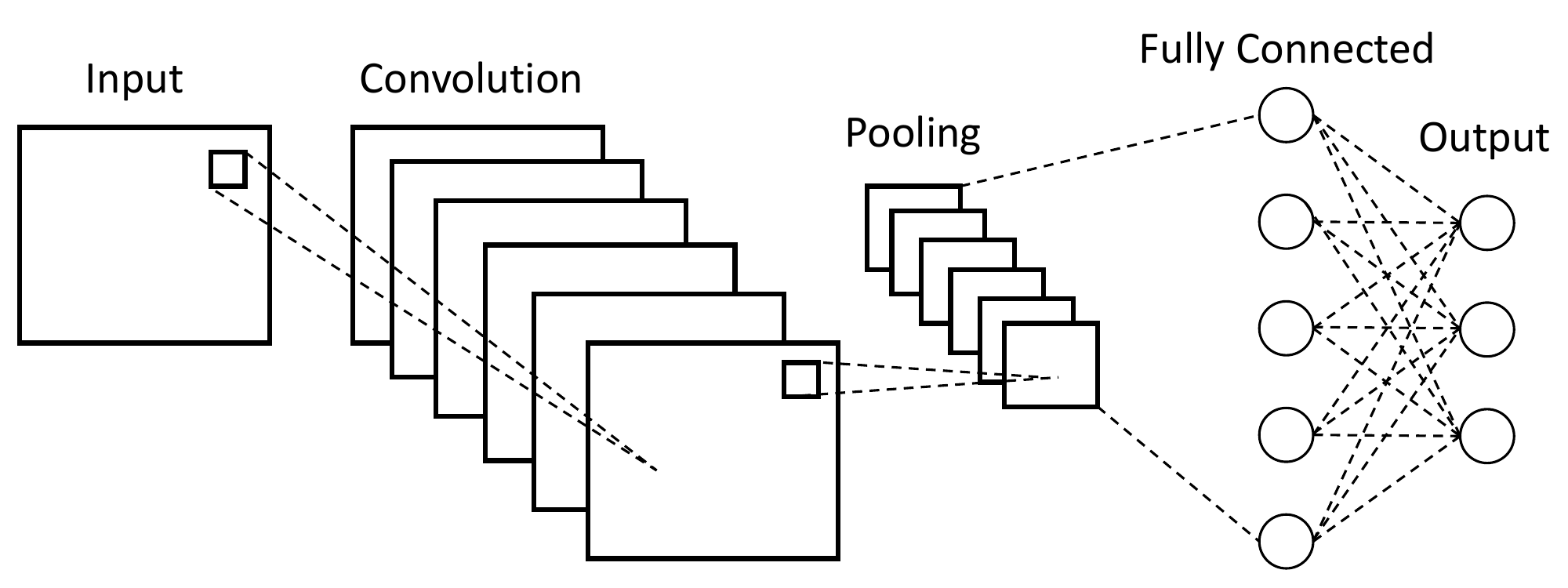}
\caption{Basic structure of convolutional neural network}
\label{Methodology}
\end{figure}

Convolutional neural networks are a type of deep learning architecture designed to process and analyze visual data like images and videos. Their ability to automatically extract and learn features from raw image data makes CNNs an ideal choice for classifying bird species to prevent bird strikes. An input image for a CNN is typically represented as a grid of pixel values. For example, a color image is stored as a 3D array, where the dimensions correspond to height, width, and three color channels (red, green, and blue) \cite{li2021survey}. These pixel values, usually normalized to a scale between 0 and 1, serve as the raw input to the CNN. Within a CNN, the input image is transformed through a series of layers as shown in Fig. \ref{Methodology}:
1) \textbf{Convolutional Layers:} These layers apply small, learnable filters (also called kernels) that slide across the input image to detect spatial patterns such as edges, textures, and shapes. Each filter produces a feature map that highlights the presence of specific features in different regions of the image. In the early layers, convolutional layers extract low-level features, while deeper layers capture increasingly complex patterns.
2) \textbf{Pooling Layers:} Pooling reduces the size of the feature maps generated by convolutional layers, which helps retain only the most important information while making the computation more efficient. Common pooling methods include max pooling (taking the largest value) and average pooling (taking the average value) in a small region.
3) \textbf{Fully Connected Layers:} After the final pooling layer, the multi-dimensional feature maps are flattened into a one-dimensional vector. It transforms the features into a format suitable for input into fully connected layers. Fully connected layers aggregate these features and use them for classification. For bird species identification, this stage assigns a probability to each species class, indicating the model's confidence in the classification.

Finally, the class with the highest probability is the candidate for output. The image is first converted into numerical arrays (pixel values), which are passed through the CNN to identify patterns, reduce irrelevant information, and ultimately classify the image. This entire process allows CNNs to take unprocessed visual data and output meaningful predictions with high accuracy \cite{li2021survey}. 

One notable advancement in CNN architectures is Residual Networks (ResNet), which address the problem of vanishing gradients in deep networks. ResNet introduces residual connections, also known as skip connections, that allow gradients to flow directly through the network by bypassing one or more layers. It enables the construction of much deeper networks such as ResNet-50 or ResNet-101 without degradation in performance \cite{he2016deep}. In our study, we chose ResNet over custom CNN models due to its superior accuracy in classifying bird species. A key reason for this performance gain is that ResNet architectures are pre-trained on large-scale datasets like ImageNet, which allows them to leverage well-initialized weights. These pretrained weights provide a strong starting point, leading to faster convergence during training and improved generalization, especially when data is limited. 

\subsection{Benchmark Models: SVM, RF, KNN}

To evaluate the performance of our CNN-based approach, we compared it against several classical machine learning models commonly used for image classification: SVM, RF, and KNN. These models are trained on features extracted from the images using conventional feature extraction techniques.

\textbf{Support Vector Machine} is a powerful supervised learning algorithm primarily used for classification tasks. The core idea of SVM is to find the optimal hyperplane that maximally separates data points of different classes in a high-dimensional space. This is achieved by maximizing the margin, which is the distance between the hyperplane and the nearest data points from each class, known as support vectors. SVM is particularly effective in high-dimensional spaces and remains robust even when the number of features exceeds the number of samples. Furthermore, by using kernel functions such as the radial basis function (RBF) or polynomial kernels, SVMs can efficiently handle non-linear decision boundaries by implicitly mapping input features into higher-dimensional spaces \cite{cortes1995support, scholkopf2002learning}.

\textbf{Random Forest} is an ensemble learning method that builds many decision trees during training and makes predictions by taking the majority vote of those trees for classification. It uses a technique called bagging, where each tree is trained on a random subset of the data, and adds extra randomness by selecting a random subset of features when splitting nodes. This helps reduce overfitting and makes the model more accurate than a single decision tree. RF works well with large and high-dimensional datasets, and it can handle complex, non-linear relationships in the data \cite{breiman2001random}. 

\textbf{K-Nearest Neighbors} is a non-parametric, instance-based learning algorithm that classifies a new data point based on the majority vote of its k-nearest neighbors in the feature space. The distance between data points is usually measured using Euclidean, Manhattan, or Minkowski distance metrics. KNN is intuitive and easy to implement, making it a strong baseline for classification tasks. However, its performance can be sensitive to the choice of k, distance metric, and feature scaling, since all features contribute equally to the distance computation. KNN also suffers from high computational complexity for large datasets \cite{cover1967nearest, altman1992introduction}.

\section{Results}
\label{Result}
This section begins by introducing the bird species most frequently involved in bird strikes in the United States from 1990 to 2023, along with the criteria used to classify their sizes, and outlines the experimental configuration  as well as performance metrics (Section \ref{DC}). The results of the two proposed approaches, CCA and UCA, are then presented in Sections \ref{PCCA} and \ref{PUCA}, respectively. Finally, Section \ref{Comparison} provides a comparative analysis of the performance of these two methods.

\subsection{Species Profiling, Size Classification, Experimental Setup, and Performance Metrics}
\label{DC}

\renewcommand{\arraystretch}{0.5}
\begin{table}[H]
\centering
\caption{Size classification of individual bird species }
\begin{tabular}{llll}
\toprule
\textbf{Species} & \textbf{Strikes from 1990-2023} & \textbf{Weight (g)} & \textbf{Class} \\
\midrule
Mourning Dove & 14,962 & 86-170 & medium \\
Barn Swallow & 9,914 & 17-20 & small \\
Killdeer & 9,881 & 75-128 & medium \\
American Kestrel & 9,104 & 80-165 & medium \\
Horned Lark & 8,263 & 28-48 & small \\
European Starling & 6,208 & 60-96 & medium \\
Eastern Meadowlark & 4,340 & 90-150 & medium \\
Rock Pigeon & 4,334 & 265-380 & medium \\
Red-tailed Hawk & 4,048 & 690-1460 & large \\
Cliff Swallow & 2,933 & 19-34 & small \\
Western Meadowlark & 2,527 & 89-115 & medium \\
American Robin & 2,318 & 77-85 & medium \\
Ring-billed Gull & 2,250 & 300-700 & medium \\
Canada Goose & 2,142 & 3000-9000 & large \\
Herring Gull & 2,087 & 800-1250 & large \\
Barn Owl & 2,047 & 400-700 & medium \\
Chimney Swift & 1,658 & 17-30 & small \\
Savannah Sparrow & 1,555 & 15-28 & small \\
Pacific Golden- plover & 1,506 & 102-108 & medium \\
Mallard & 1,389 & 1000-1300 & large \\
Common Nighthawk & 1,293 & 53-57 & small \\
Tree Swallow & 1,248 & 16-25 & small \\
Laughing Gull & 1,204 & 203-371 & medium \\
Turkey Vulture & 1163 & 2000 & large \\
Cattle Egret & 895 & 270-512 & medium \\
Short-eared Owl & 882 & 206-475 & medium \\
Bank Swallow & 825 & 10-19 & small \\
Red-winged Blackbird & 775 & 32-77 & small \\
Yellow-rumped Warbler & 741 & 12-13 & small \\
American Crow & 721 & 316-620 & medium \\
Swainson's Thrush & 681 & 23-45 & small \\
Osprey & 628 & 1400-2000 & large \\
Peregrine & 618 & 530-1600 & large \\

\bottomrule
\end{tabular}

\label{table:comparison}
\end{table}

\subsubsection{Size Categorization} 
\label{Size}
To develop a formal classification system for bird species based on size, we first gathered weight data for 33 high-risk species from the Cornell Lab of Ornithology \cite{CL}. These species were identified using strike data from the FAA wildlife strike database spanning from 1990 to 2023 \cite{faa2023wildlife}. To determine how to categorize birds into size classes, we consulted with DeTect Inc \cite{Merlin}, the company behind Merlin radar systems, which uses radar cross-section analysis to group birds into three size categories: small, medium, and large. Based on DeTect's guidance, we adopted their standard classification thresholds: small (0–70 grams), medium (71–800 grams), and large (801–1700 grams). Species with weights exceeding 1700 grams, such as the Canada Goose and Turkey Vulture, were still categorized as large due to their substantial radar visibility. When a species’ weight range spanned multiple categories, we assigned it to the size class that covered the majority of its weight range. The full list of species, along with their strike frequency, weight range, and assigned classification, is presented in Table \ref{table:comparison}.

\subsubsection{Experimental Configuration}
\label{Configuration}
In this paper, we implemented CNN-based classifiers for bird species, along with three additional classification models namely SVM, RF, and KNN which serve as benchmark models. To ensure the classification reflects real-world conditions, we used images of flying birds. For the CNN classifier, the dataset was divided into three parts: 80\% for training, 10\% for validation, and 10\% for testing. Since the benchmark models do not require a validation set, we allocated 90\% of the data for training and 10\% for testing to maintain consistency in test set size across all models. The dataset used in our experiments was balanced, with an equal number of samples for each class. Also, the input image size considered was (512, 512, 3).

We used Area Under the Curve (AUC) and accuracy as the evaluation metrics during CNN training. The best model was selected based on the highest AUC score on the validation set, with accuracy also considered as an important performance indicator. Among the various ResNet versions tested, ResNet50V2 demonstrated the best performance. To further optimize the model’s accuracy, we also experimented with different configurations for the number of neurons in the fully connected layers, testing values of 64, 128, 256, and 512. The batch size used during training was 32. The model was compiled with the following configuration: the optimizer was set to Adam, the loss function used was categorical cross-entropy, and the evaluation metrics included both AUC and accuracy to track model performance. To reduce unnecessary computation, an early stopping mechanism was implemented: if the AUC on the validation set did not improve for 35 consecutive epochs, training was stopped. Otherwise, training continued for a maximum of 100 epochs. Table \ref{table:experiment_parameters} shows a summary of experimental parameters and configurations.

\renewcommand{\arraystretch}{0.6}
\begin{table}[H]
\centering
\caption{Summary of experimental parameters and configurations.}
\begin{tabular}{@{}ll@{}}
\toprule
\textbf{Parameter}                     & \textbf{Value/Description}                                         \\ \midrule
Model Types                  & CNN                           \\
Dataset Split (CNN)               & 80\% Training, 10\% Validation, 10\% Testing          \\

Image Size (CNN) &  512×512×3 \\
Evaluation Metrics            & AUC, Accuracy                          \\
Selection Criteria          & Highest AUC score on validation set (higher accuracy if AUC ties) \\
CNN Architecture              & ResNet50V2 (best-performing model)                                \\
Neurons in Fully Connected Layer & 256                       \\
Batch Size                    & 32                                                                \\
Optimizer                    & Adam                                    \\
Loss Function                 & Categorical cross-entropy                                         \\
Early Stopping Criteria       & Stop if AUC does not improve for 35 epochs, max 100 epochs        \\ \bottomrule
\end{tabular}

\label{table:experiment_parameters}
\end{table}

\subsubsection{Performance Metrics}
\label{Evaluation}
In this subsection, we present the performance metrics to evaluate the performance of classifiers. In a binary or multi-class classification, True Positive (TP) is the number of correctly predicted positive instances, False Positive (FP) is the number of negative instances incorrectly predicted as positive, False Negative (FN) is the number of positive instances incorrectly predicted as negative, and True Negative (TN) is the number of correctly predicted negative instances. According to these values, we report four standard performance metrics: precision, recall, F1-score, and accuracy, which are commonly used to assess classification performance. These metrics are defined as follows:

\textbf{Precision} is the ratio of true positives to the sum of true positives and false positives:
    \[
    \text{Precision} = \frac{TP}{TP + FP}
    \]
    
\textbf{Recall} (also known as sensitivity) is the ratio of true positives to the sum of true positives and false negatives:
    \[
    \text{Recall} = \frac{TP}{TP + FN}
    \]
    
 \textbf{F1-score} is the harmonic mean of precision and recall:
    \[
    \text{F1-score} = 2 \times \frac{\text{Precision} \times \text{Recall}}{\text{Precision} + \text{Recall}}
    \]
    
 \textbf{Accuracy} is the percentage of correctly predicted instances among all instances:
    \[
    \text{Accuracy} = \frac{TP + TN}{TP + TN + FP + FN} \times 100
    \]

\subsection{The Performance of the Cascade Classification Approach}
\label{PCCA}
As outlined earlier, the CCA consists of three sequential classifiers. The first classifier distinguishes between bird and aircraft images. The second classifier applied only to images identified as birds, determines the bird’s size category (small, medium, or large). Finally, based on the predicted size, a dedicated species-level classifier is used to identify the bird species. To evaluate the effectiveness of the CCA, we examine the performance of each stage in the classification pipeline. The following subsections report the performance metrics for each of these classifiers.

\subsubsection{Classifier 1: Determining Bird from Aircraft}

\begin{figure}[H]
    \centering
        \includegraphics[width=10 cm, height = 5 cm]{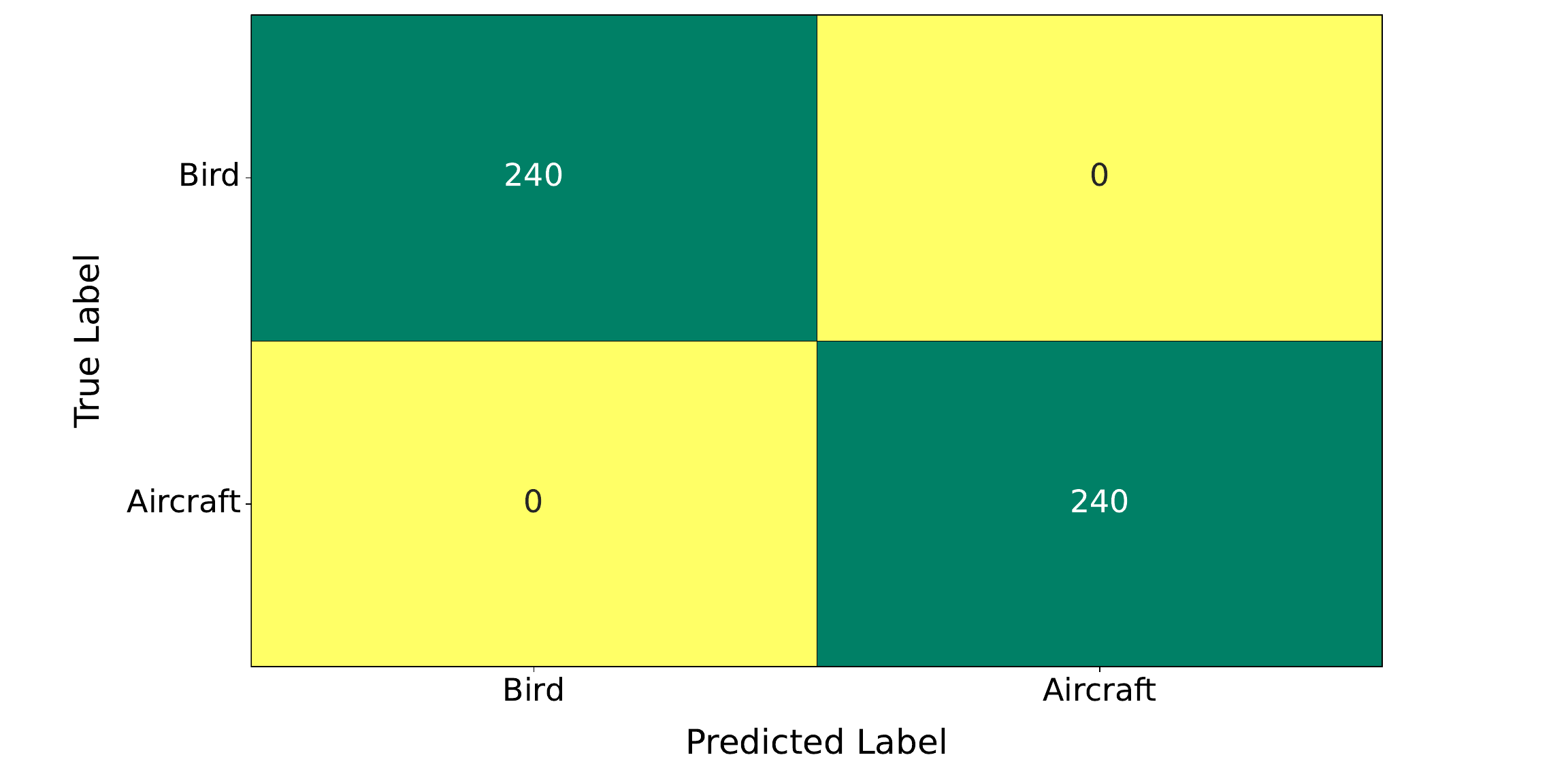}
        \caption{Confusion matrix of the classifier 1 to determine bird and aircraft.}
        \label{CNN-BA}
\end{figure}

To address the critical need for distinguishing birds from aircraft in aviation safety systems, we developed a binary classifier focused on separating bird and aircraft images. Among the 33 bird species most frequently involved in bird strikes in the United States from 1990 to 2023, we were unable to collect sufficient images for the following species: Cattle Egret, Horned Lark, Savannah Sparrow, Swainson's Thrush, Barn Owl, and Pacific Golden-Plover. Therefore, we removed them from our analysis. We created a dataset consisting of 2,400 images for each class (bird and aircraft). Bird images were collected from Cornell Lab of Ornithology \cite{CL}, and aircraft images which encompassed fixed-wing aircraft, helicopters, and and unmanned aerial vehicles (UAVs) are collected from \cite{aircraft_dataset, helicopter_dataset, drone_dataset}, respectively. Then, we implemented a CNN using the ResNet50V2 architecture for classification. The dataset was split into 80\% for training (1920 images per class), 10\% for validation (240 images per class), and 10\% for testing (240 images per class). The resulting model achieved an accuracy of 100\% on the test set, reflecting the controlled dataset and balanced classes used in this study. The confusion matrix, shown in Fig.\ref{CNN-BA} indicates perfect classification with no mislabeling between bird and aircraft categories. Furthermore, the precision, recall, and F1-score for both classes are 1.00, as shown in Table \ref{tab:BA}, confirming the model’s outstanding ability to distinguish between these two categories.

\renewcommand{\arraystretch}{0.5}
\begin{table}[H]
\centering
\caption{Class-wise precision, recall, and F1-score of CNN for identifying bird and aircraft.}
\begin{tabular}{lccc}
\toprule
\textbf{Class} & \textbf{Precision} & \textbf{Recall} & \textbf{F1-score} \\
\midrule
Bird     & 1.00 & 1.00 & 1.00 \\
Aircraft  & 1.00 & 1.00 & 1.00 \\
\midrule
\textbf{Accuracy} & \multicolumn{3}{c}{\textbf{100\%}} \\
\bottomrule
\end{tabular}
\label{tab:BA}
\end{table}

\subsubsection{Classifier 2: Determining Bird Size}
\begin{figure}[H]
    \centering
        \includegraphics[width=10 cm, height = 5 cm]{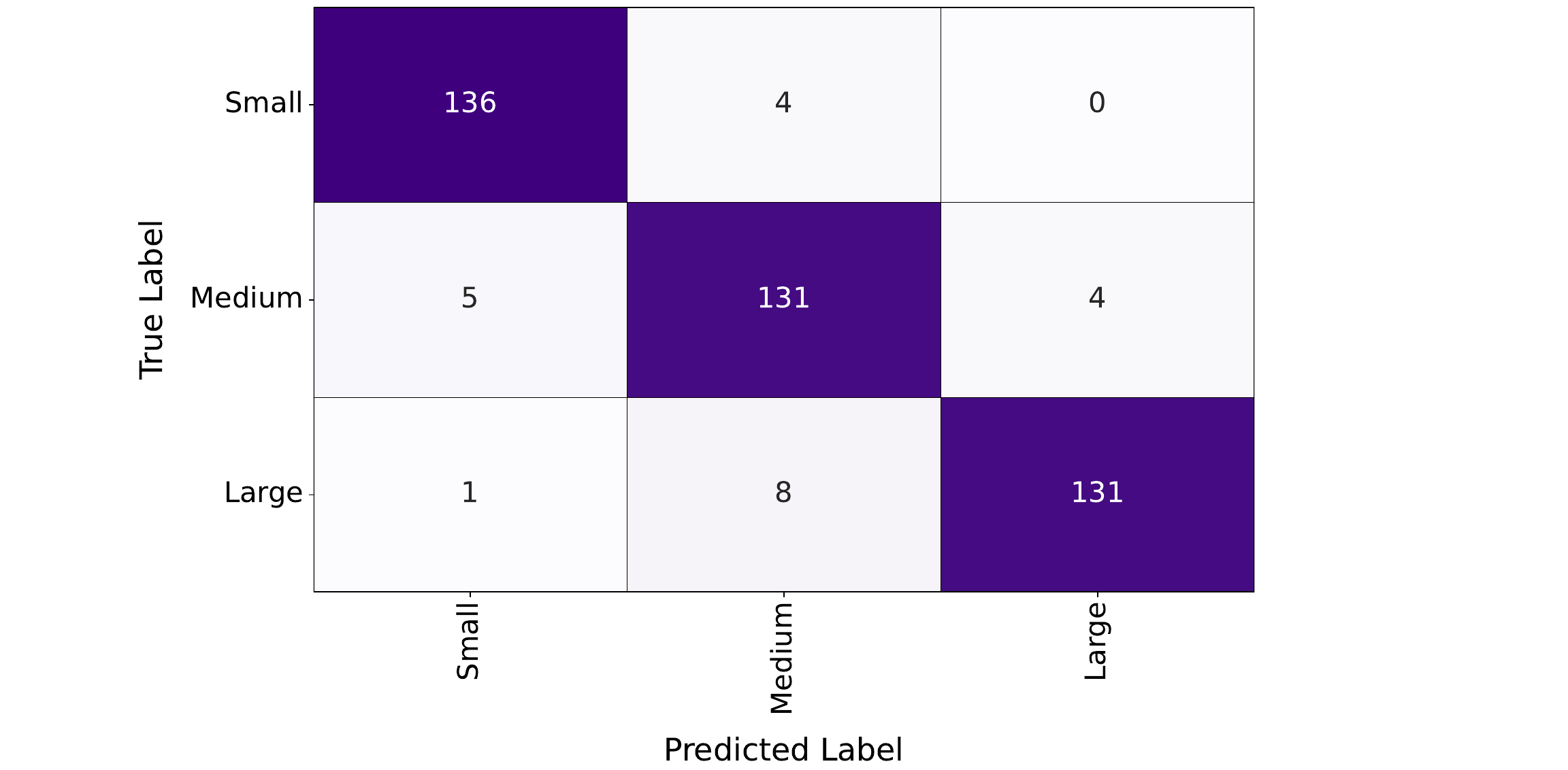}
        \caption{Confusion matrix of the classifier 2 to determine size of the bird.}
        \label{CNN-Size}
\end{figure}

With the size labels established through weight-based categorization (Section \ref{Size}), we developed a CNN classifier to predict bird size (small, medium, or large) directly from images. This classifier is designed to enable automated estimation of bird size from visual data, which is particularly useful in scenarios where avian radar systems are unavailable. To construct the dataset, we collected images of the 33 bird species listed in Table~\ref{table:comparison}, with each image labeled according to the species' assigned size class. For instance, images of Barn Swallow, Bank Swallow, and Tree Swallow were labeled as “small”,  whereas Rock Pigeon and American Kestrel were labeled “medium”,  and Canada Goose and Red-tailed Hawk were labeled “large”. We trained this classifier using the ResNet50V2 architecture. The model learns to associate visual cues such as body proportions, wingspan, and overall mass with the corresponding size categories. Due to limited image availability, we excluded several species (Barn Owl, Pacific Golden-plover, Cattle Egret, Horned Lark, Savannah Sparrow, and Swainson's Thrush) from the training set. We also combined Barn and Cliff Swallows, Eastern and Western Meadowlarks, and Ring-billed and Laughing Gulls, as the species in each pair closely resemble one another. After these adjustments, the final dataset included 24 distinct bird species. From each category (small, medium, and large), we collected 1400 images, resulting in a total of 4200 labeled images. These were divided into training (80\%), validation (10\%), and test (10\%) sets to develop and evaluate the classifier’s performance. The trained model demonstrated strong performance across all three size categories. As illustrated in Fig. \ref{CNN-Size}, the classifier maintained high accuracy in distinguishing between small, medium, and large birds, with minimal misclassification. Table \ref{tab:Size} presents the class-wise precision, recall, and F1-scores, highlighting consistently strong results across categories. The overall classification accuracy achieved on the test set was 94.76\%, confirming the model's effectiveness in visually estimating bird size.

\renewcommand{\arraystretch}{0.5}
\begin{table}[H]
\centering
\caption{Class-wise precision, recall, and F1-score of CNN for identifying size of the bird.}
\begin{tabular}{lccc}
\toprule
\textbf{Class} & \textbf{Precision} & \textbf{Recall} & \textbf{F1-score} \\
\midrule
Small     & 0.96 & 0.97 & 0.96 \\
Medium    & 0.92 & 0.94 & 0.93 \\
Large     & 0.97 & 0.94 & 0.95 \\
\midrule
\textbf{Accuracy} & \multicolumn{3}{c}{\textbf{94.76\%}} \\
\bottomrule
\end{tabular}
\label{tab:Size}
\end{table}

\subsubsection{Classifier 3, 4, 5: Determining Bird Species}
To enhance bird strike mitigation efforts, we developed and evaluated classification models to identify bird species within three size-specific groups: large, medium, and small. Each classifier was designed to differentiate between bird species that frequently appear in bird strike reports in the United States from 1990 to 2023. For each size category, we implemented a CNN based on the ResNet50V2 architecture, and benchmarked its performance against three traditional machine learning models: SVM, RF, and KNN. These benchmark models were included to assess whether simpler, less compute-intensive algorithms could achieve comparable performance. The following subsections detail the dataset preparation, model configurations, training procedures, and comparative performance analysis for each bird size group.

For the benchmark SVM, RF, and KNN we performed hyperparameter tuning using grid search. For the SVM model, we varied the regularization parameter $C$ with values of 0.1, 1, 10, and 100. We tested different kernel functions, including polynomial, linear, RBF, and sigmoid. The gamma parameter was also tuned with both the 'scale' option and explicit values of 0.01 and 0.001. For RF classifier, we tuned several parameters: the number of estimators was varied among 100, 200, and 300 trees; the maximum depth of the trees was set to None, 10, 20, or 30; the minimum number of samples required to split an internal node was set to 2, 5, or 10; and the minimum number of samples required to be at a leaf node was set to 1, 2, or 4. We also evaluated the effect of using bootstrapped datasets versus not using bootstrap sampling. In the case of KNN classifier, we varied the number of neighbors from 1 to 10. We also evaluated different distance weighting strategies, including uniform and distance-based weights, and tested several distance metrics such as Euclidean, Manhattan, and Chebyshev. Additionally, we experimented with all major search algorithms, including auto, ball tree, KD tree, and brute-force.

\noindent \textbf{Classifier 3 (Large Birds):} For the large bird group, we considered seven bird species that are frequently involved in bird strikes: Herring Gull, Turkey Vulture, Mallard, Peregrine, Red-tailed hawk, Canada Goose, and Osprey. A total of 360 images were considered for each class, with 288 images used for training, 36 images for validation, and 36 images for testing. For the benchmark models, we used 324 images per class for training and 36 images per class for evaluation, without a separate validation set, as validation data set was not required for these classifiers. For each model, we performed grid search to identify the optimal hyperparameters. For the SVM, the best parameters were found to be: C = 10, gamma = scale, and kernel = rbf. For the RF model, the optimal parameters included: bootstrap = false, max depth = 20, minimum samples leaf = 1, minimum samples split = 5, and number of estimators = 300. Finally, for the KNN model, the best configuration consisted of: algorithm = auto, metric = Euclidean, number of neighbors = 1, and weights = uniform. When we applied benchmark classifiers to high-resolution image inputs (512×512×3), we encountered memory errors during execution. As image resolution increases, the dimensionality of the input vectors grows proportionally, resulting in substantially higher memory usage. To address this, we limited the input size to 128×128×3 for benchmark models, which allowed the models to run without memory issues while maintaining reasonable input detail. 

\begin{figure}
     \centering
     \begin{subfigure}[h]{0.495\textwidth}
         \centering
         \includegraphics[width = 8 cm, height=5cm]{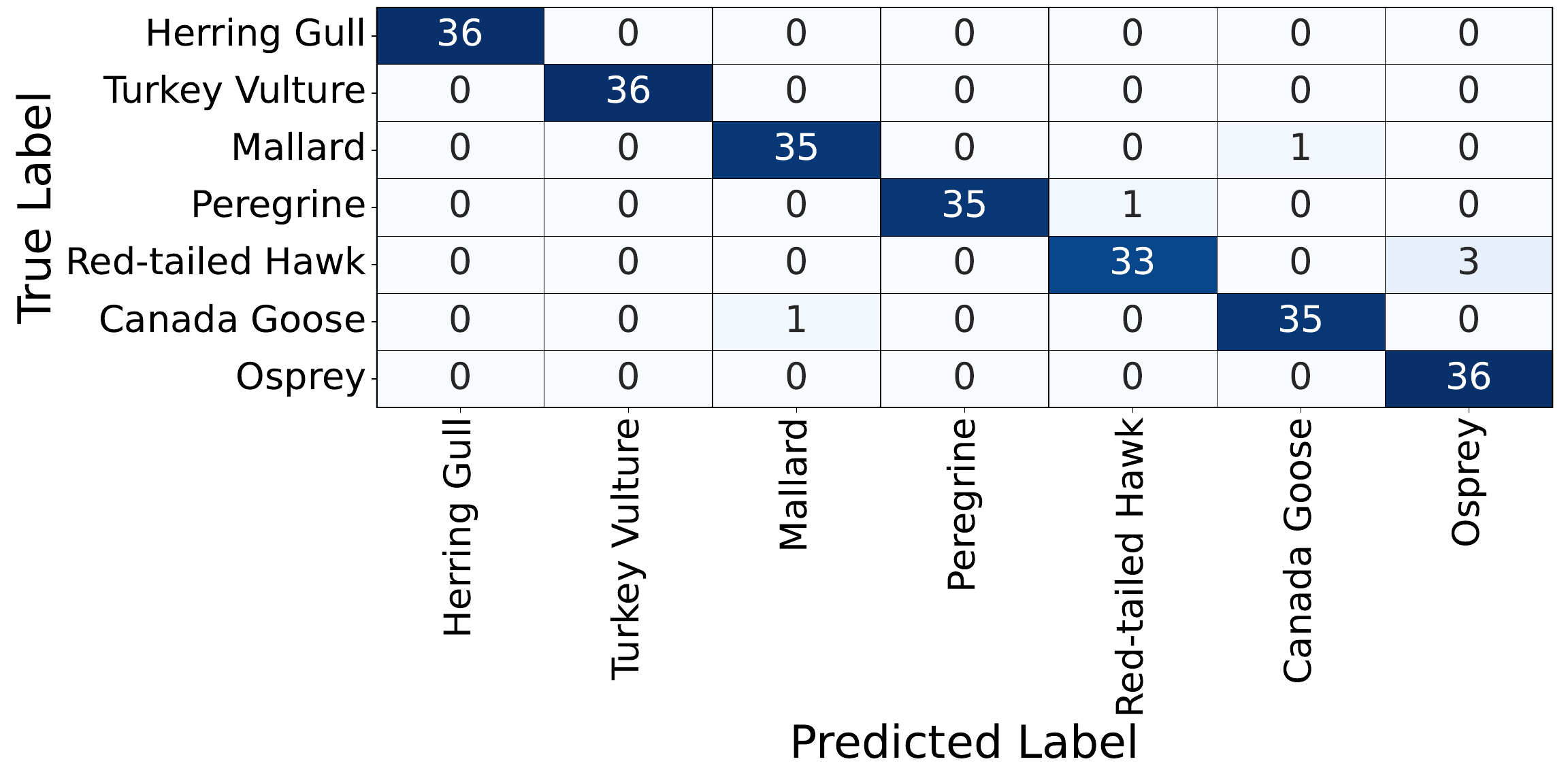}
         \caption{Confusion matrix of the CNN model}
         \label{CNN-Large}
     \end{subfigure}
     \hfill
     \begin{subfigure}[h]{0.495\textwidth}
         \centering
         \includegraphics[width = 8 cm, height=5cm]{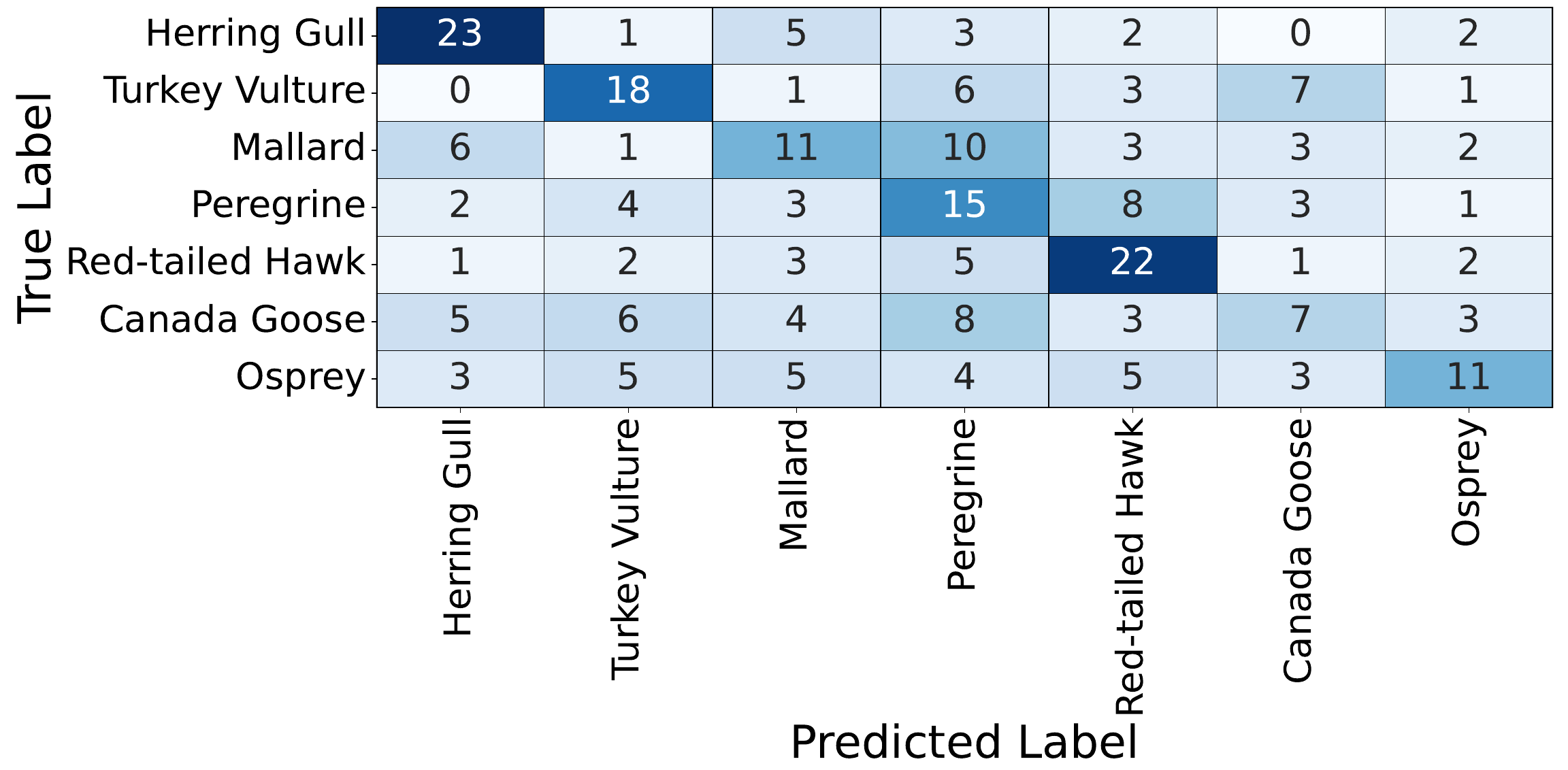}
         \caption{Confusion matrix of the SVM model}
         \label{SVM-Large}
     \end{subfigure}
\vspace{3 mm} 

     \centering
     \begin{subfigure}[h]{0.495\textwidth}
         \centering
         \includegraphics[width = 8 cm, height=5cm]{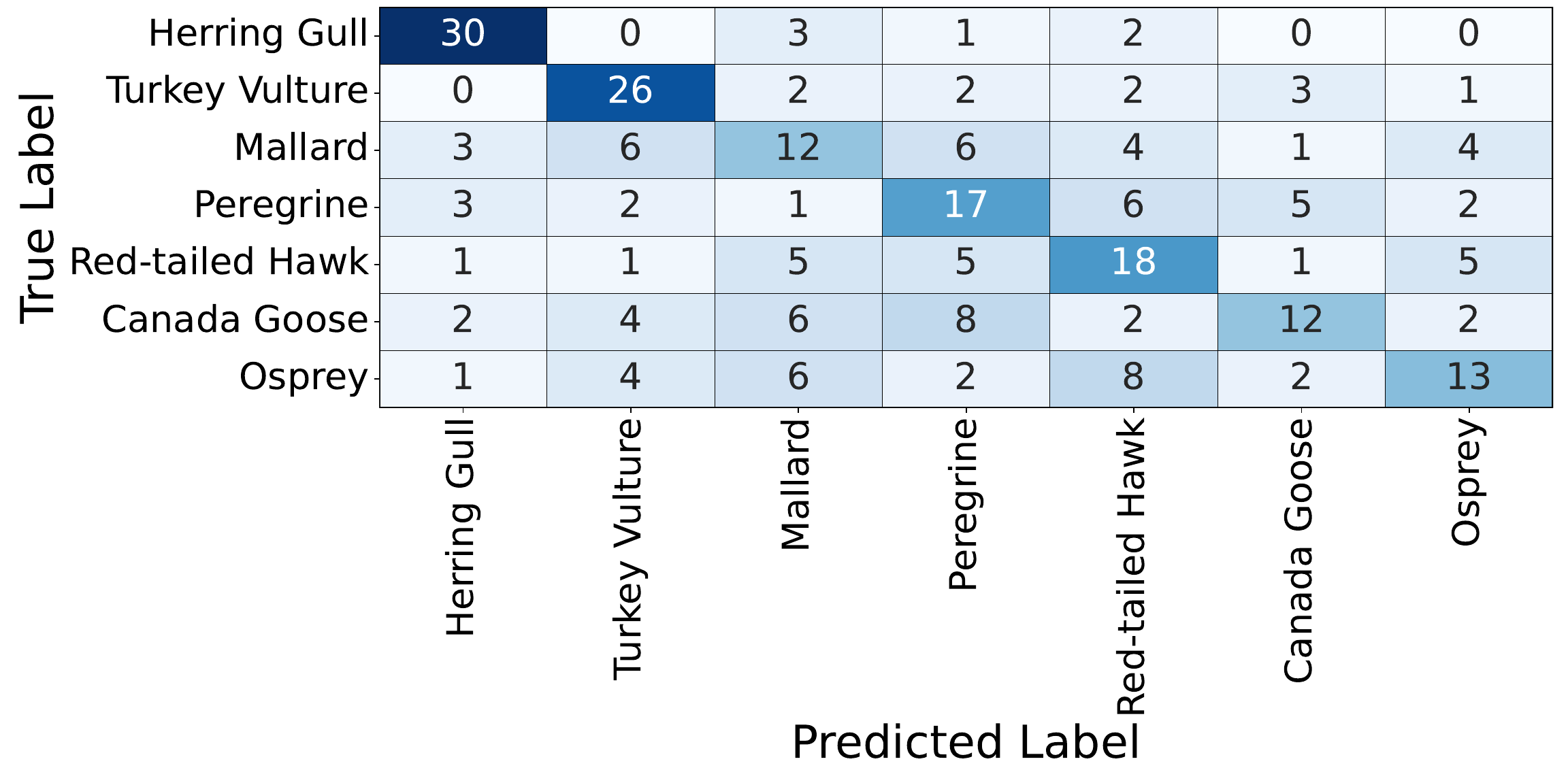}
         \caption{Confusion matrix of the RF model}
         \label{RF-Large}
     \end{subfigure}
     \hfill
     \begin{subfigure}[h]{0.495\textwidth}
         \centering
         \includegraphics[width = 8 cm, height=5cm]{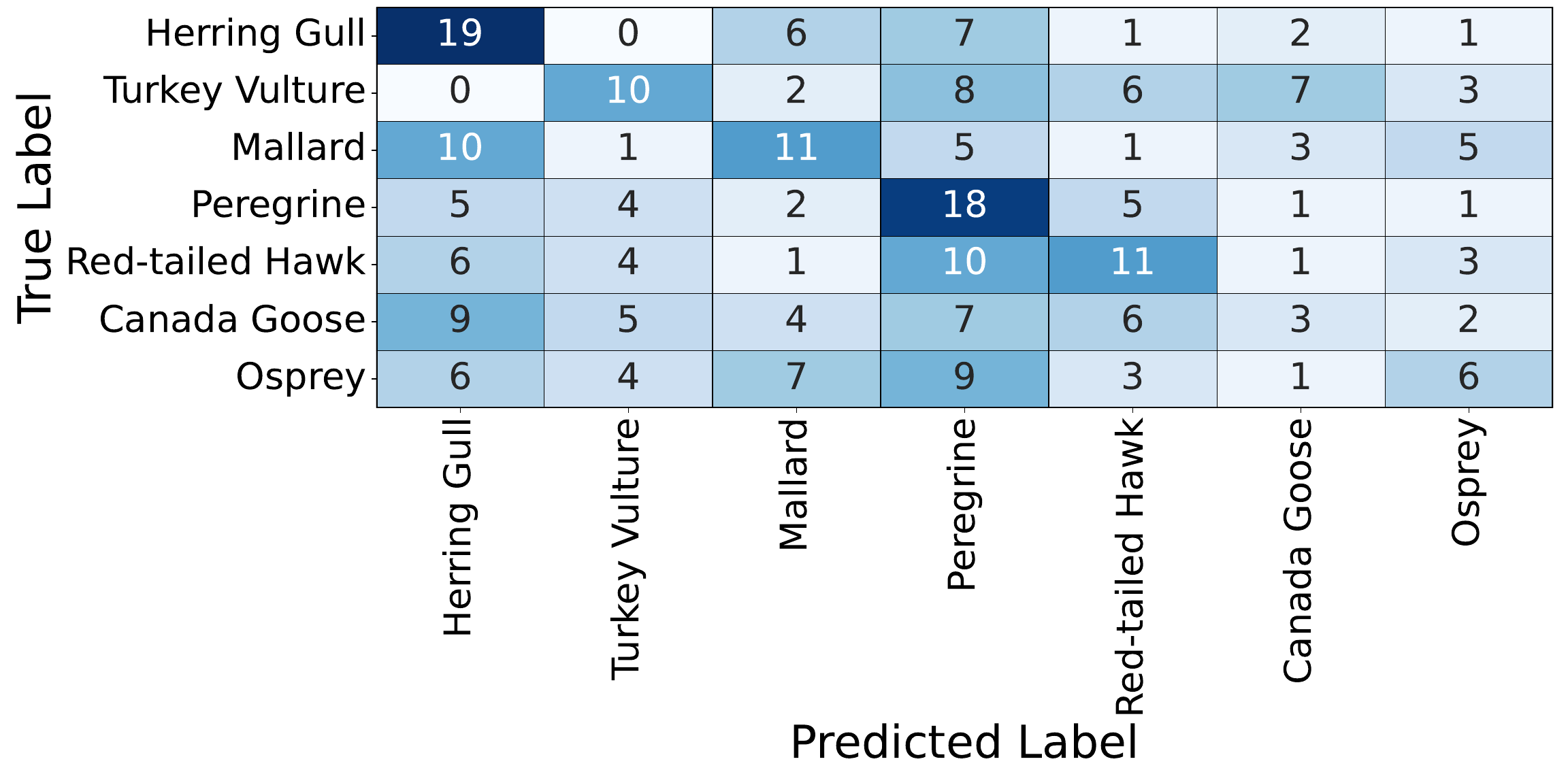}
         \caption{Confusion matrix of the KNN model}
         \label{KNN-Large}
     \end{subfigure}
     \caption{Comparison of confusion matrices from four different classification models: CNN, SVM, RF, and KNN for large size birds (classifier 3).}
     \label{Large}
    
\end{figure}

The confusion matrices of all models with the optimal parameters are shown in Fig. \ref{Large}. From these matrices, it is clear that the CNN model achieves the best accuracy compared to the other models. Table \ref{tab:Large} presents the precision, recall, F1 score, and accuracy for each model and for each class. The results show that, for the large bird classification task, CNN performed the best (accuracy = 97.62\%), followed by RF (accuracy = 50.19\%), SVM (accuracy = 42.46\%), and KNN (accuracy = 34.92\%).

\floatplacement{figure}{H}

\renewcommand{\arraystretch}{0.5}
\begin{table}[H]
\centering
\caption{Class-wise precision, recall, and F1-score for CNN, SVM, RF, and KNN models for large size birds (classifier 3).}
\begin{tabularx}{\textwidth}{l *{12}{>{\centering\arraybackslash}X}}
\toprule
\textbf{Class} 
& \multicolumn{3}{c}{\textbf{CNN}} 
& \multicolumn{3}{c}{\textbf{SVM}} 
& \multicolumn{3}{c}{\textbf{RF}} 
& \multicolumn{3}{c}{\textbf{KNN}} \\
& \rotatebox{90}{Precision} & \rotatebox{90}{Recall} & \rotatebox{90}{F1-score} 
& \rotatebox{90}{Precision} & \rotatebox{90}{Recall} & \rotatebox{90}{F1-score} 
& \rotatebox{90}{Precision} & \rotatebox{90}{Recall} & \rotatebox{90}{F1-score} 
& \rotatebox{90}{Precision} & \rotatebox{90}{Recall} & \rotatebox{90}{F1-score} \\
\midrule
Herring Gull      & 1.00 & 1.00 & 1.00 & 0.57 & 0.64 & 0.61 & 0.75 & 0.83 & 0.79 & 0.35 & 0.53 & 0.42 \\
Turkey Vulture    & 1.00 & 1.00 & 1.00 & 0.49 & 0.50 & 0.49 & 0.60 & 0.72 & 0.66 & 0.36 & 0.28 & 0.31 \\
Mallard           & 0.97 & 0.97 & 0.97 & 0.34 & 0.31 & 0.32 & 0.34 & 0.33 & 0.34 & 0.33 & 0.31 & 0.32 \\
Peregrine         & 1.00 & 0.97 & 0.99 & 0.29 & 0.42 & 0.34 & 0.41 & 0.47 & 0.44 & 0.28 & 0.50 & 0.36 \\
Red-tailed hawk   & 0.97 & 0.92 & 0.94 & 0.48 & 0.61 & 0.54 & 0.43 & 0.50 & 0.46 & 0.33 & 0.31 & 0.32 \\
Canada Goose      & 0.97 & 0.97 & 0.97 & 0.29 & 0.19 & 0.23 & 0.50 & 0.33 & 0.40 & 0.17 & 0.08 & 0.11 \\
Osprey            & 0.92 & 1.00 & 0.96 & 0.50 & 0.31 & 0.38 & 0.48 & 0.36 & 0.41 & 0.29 & 0.17 & 0.21 \\
\midrule
\textbf{Accuracy} & \multicolumn{3}{c}{\textbf{97.62\%}} 
                  & \multicolumn{3}{c}{\textbf{42.46\%}} 
                  & \multicolumn{3}{c}{\textbf{50.19\%}} 
                  & \multicolumn{3}{c}{\textbf{30.95\%}} \\
\bottomrule
\end{tabularx}
\label{tab:Large}
\end{table}

\noindent\textbf{Classifier 4 (Medium Birds):} For medium-sized birds, we initially considered 15 species frequently involved in bird strikes: Mourning Dove, Killdeer, American Kestrel, European Starling, Eastern Meadowlark, Rock Pigeon, Western Meadowlark, American Robin, Ring-billed Gull, Barn Owl, Pacific Golden-plover, Laughing Gull, Cattle Egret, Short-eared Owl, and American Crow. After gathering images, we merged the Eastern and Western Meadowlarks into a single group due to their high visual similarity. Similarly, Ring-billed Gulls and Laughing Gulls were combined. Due to the limited number of available images for the Barn Owl, Pacific Golden-plover, and Cattle Egret, these species were excluded from further analysis. As a result, the classifier is ultimately designed to distinguish between 10 types of birds. For each class, 240 images were collected, with 192 images allocated for training, 24 for validation, and 24 for testing. For the benchmark models, we used 216 images per class for training and 24 images per class for evaluation, without using a separate validation data set, as it was not necessary for these classifiers. Grid search was conducted for each model to determine the optimal hyperparameters. For the SVM model, the best parameters were: C = 10, gamma = scale, and kernel = rbf. For the RF model, the optimal settings included: bootstrap = False, max depth = 10, minimum samples leaf = 2, minimum samples split = 5, and number of estimators = 300. For the KNN model, the best configuration consisted of: algorithm = auto, metric = Euclidean, number of neighbors = 9, and weights = distance. For the aforementioned reason, we limited the input size to 128×128×3 for benchmark models to solve the memory issue.

\begin{figure}
     \centering
     \begin{subfigure}[h]{0.495\textwidth}
         \centering
         \includegraphics[width = 8 cm, height=5cm]{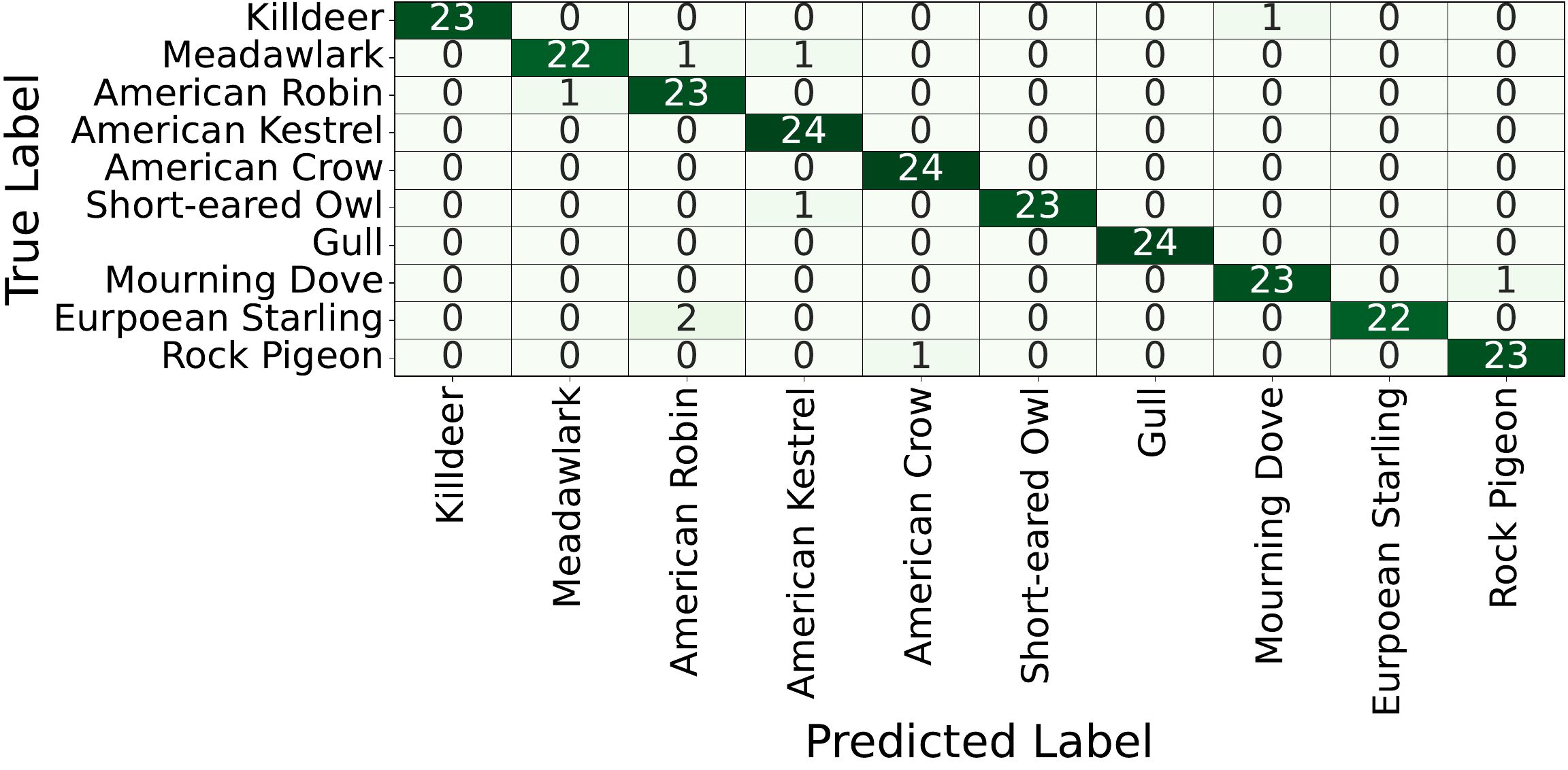}
         \caption{Confusion matrix of the CNN model}
         \label{CNN-Medium}
     \end{subfigure}
     \hfill
     \begin{subfigure}[h]{0.495\textwidth}
         \centering
         \includegraphics[width = 8 cm, height=5cm]{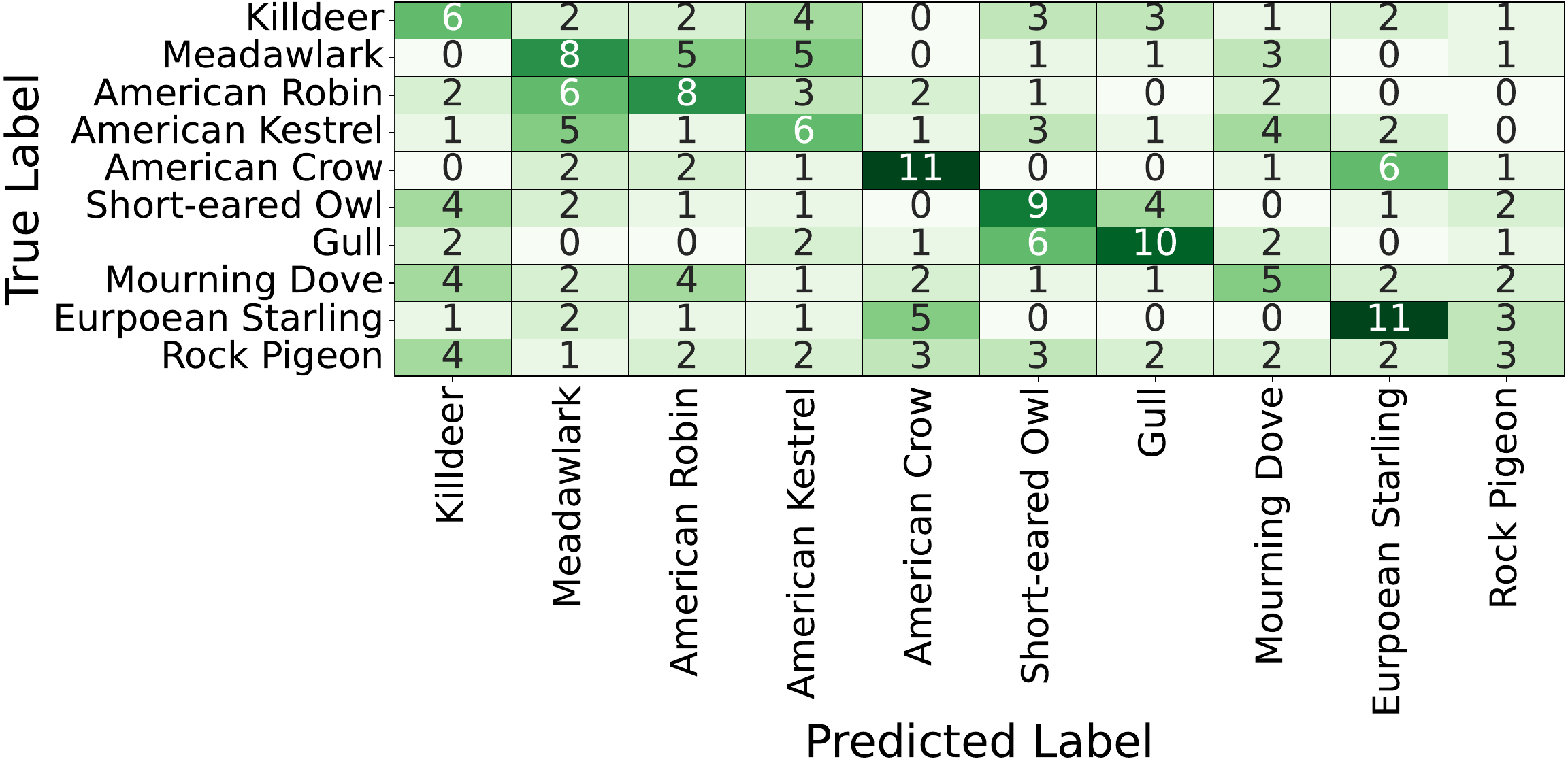}
         \caption{Confusion matrix of the SVM model}
         \label{SVM-Medium}
     \end{subfigure}
\vspace{3 mm} 

     \centering
     \begin{subfigure}[h]{0.495\textwidth}
         \centering
         \includegraphics[width = 8 cm, height=5cm]{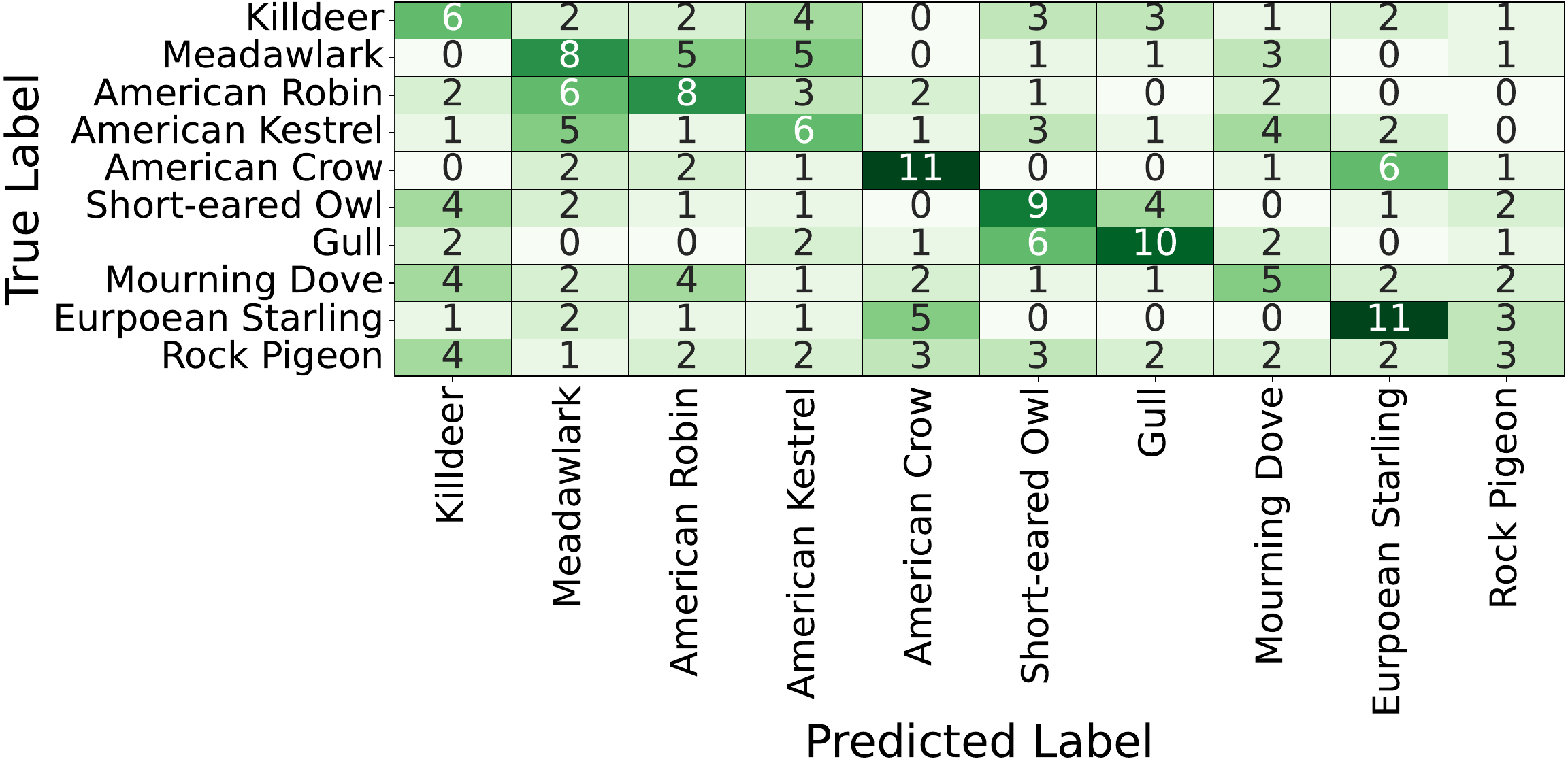}
         \caption{Confusion matrix of the RF model}
         \label{RF-Medium}
     \end{subfigure}
     \hfill
     \begin{subfigure}[h]{0.495\textwidth}
         \centering
         \includegraphics[width = 8 cm, height=5 cm]{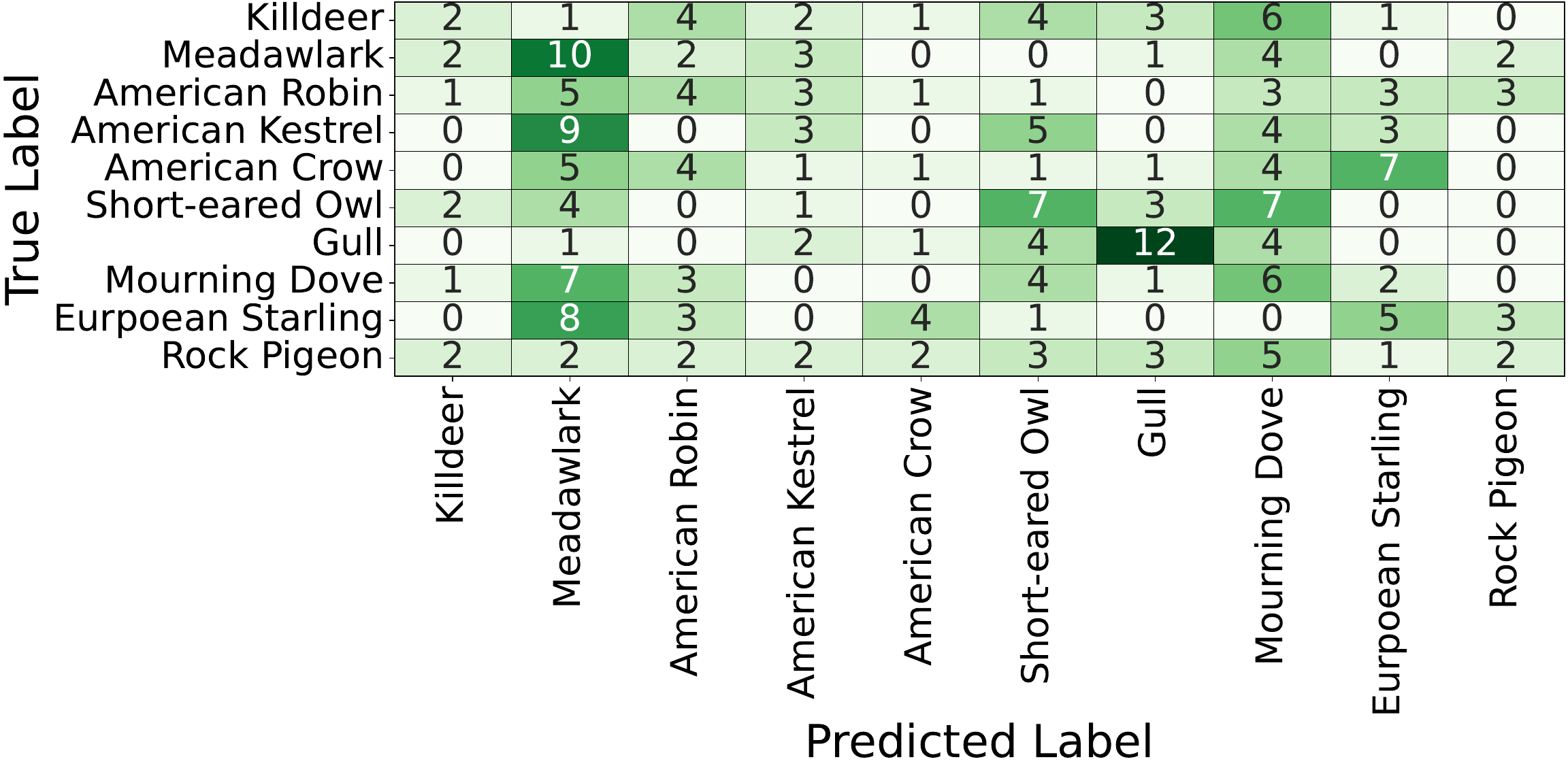}
         \caption{Confusion matrix of the KNN model}
         \label{KNN-Medium}
     \end{subfigure}
     \caption{Comparison of confusion matrices from four different classification models: CNN, SVM, RF, and KNN for medium size birds (classifier 4).}
     \label{Medium}
    
\end{figure}

Table \ref{tab:Medium} summarizes the precision, recall, F1 score, and accuracy for each model and each class. These results demonstrate that, for the medium bird classification task, the CNN outperformed all other models, achieving the best overall accuracy (96.25\%). The performance of the other models was as follows: RF achieved an accuracy of 32.92\%, SVM achieved 32.08\%, and KNN had the lowest accuracy at 21.67\%. The confusion matrices for all models with their optimal parameters are illustrated in Fig. \ref{Medium}. As shown, the CNN model achieves the highest classification accuracy compared to the benchmark models.

\renewcommand{\arraystretch}{0.65}
\begin{table}[H]
\centering
\caption{Class-wise precision, recall, and F1-score for CNN, SVM, RF, and KNN models for medium size birds (classifier 4).}
\begin{tabularx}{\textwidth}{l *{12}{>{\centering\arraybackslash}X}}
\toprule
\textbf{Class} 
& \multicolumn{3}{c}{\textbf{CNN}} 
& \multicolumn{3}{c}{\textbf{SVM}} 
& \multicolumn{3}{c}{\textbf{RF}} 
& \multicolumn{3}{c}{\textbf{KNN}} \\
& \rotatebox{90}{Precision} & \rotatebox{90}{Recall} & \rotatebox{90}{F1-score} 
& \rotatebox{90}{Precision} & \rotatebox{90}{Recall} & \rotatebox{90}{F1-score} 
& \rotatebox{90}{Precision} & \rotatebox{90}{Recall} & \rotatebox{90}{F1-score} 
& \rotatebox{90}{Precision} & \rotatebox{90}{Recall} & \rotatebox{90}{F1-score} \\
\midrule
Killdeer                & 1.00 & 0.96 & 0.98 & 0.25 & 0.25 & 0.25 & 0.30 & 0.25 & 0.27 & 0.20 & 0.08 & 0.12 \\
Meadowlark              & 0.96 & 0.92 & 0.94 & 0.27 & 0.33 & 0.30 & 0.40 & 0.33 & 0.36 & 0.19 & 0.42 & 0.26 \\
American Robin          & 0.88 & 0.96 & 0.92 & 0.31 & 0.33 & 0.32 & 0.26 & 0.33 & 0.29 & 0.18 & 0.17 & 0.17 \\
American Kestrel        & 0.92 & 1.00 & 0.96 & 0.23 & 0.25 & 0.24 & 0.24 & 0.29 & 0.26 & 0.18 & 0.12 & 0.15 \\
American Crow           & 0.96 & 1.00 & 0.98 & 0.44 & 0.46 & 0.45 & 0.56 & 0.58 & 0.57 & 0.10 & 0.04 & 0.06 \\
Short-eared Owl         & 1.00 & 0.96 & 0.98 & 0.33 & 0.38 & 0.35 & 0.30 & 0.33 & 0.31 & 0.23 & 0.29 & 0.26 \\
Gull                    & 1.00 & 1.00 & 1.00 & 0.45 & 0.42 & 0.43 & 0.34 & 0.42 & 0.38 & 0.50 & 0.50 & 0.50 \\
Mourning Dove           & 0.96 & 0.96 & 0.96 & 0.25 & 0.21 & 0.23 & 0.27 & 0.25 & 0.26 & 0.14 & 0.25 & 0.18 \\
European Starling       & 1.00 & 0.92 & 0.96 & 0.42 & 0.46 & 0.44 & 0.41 & 0.38 & 0.39 & 0.23 & 0.21 & 0.22 \\
Rock Pigeon             & 0.96 & 0.96 & 0.96 & 0.21 & 0.12 & 0.16 & 0.20 & 0.12 & 0.15 & 0.20 & 0.08 & 0.12 \\

\midrule
\textbf{Accuracy} & \multicolumn{3}{c}{\textbf{96.25\%}} 
                  & \multicolumn{3}{c}{\textbf{32.08\%}} 
                  & \multicolumn{3}{c}{\textbf{32.92\%}} 
                  & \multicolumn{3}{c}{\textbf{21.67\%}} \\
\bottomrule
\end{tabularx}
\label{tab:Medium}
\end{table}

\floatplacement{figure}{H}

\noindent\textbf{Classifier 5 (Small Birds):} Initially, we focused on 11 small bird species frequently involved in bird strikes in the United States from 1990 to 2023: Barn Swallow, Horned Lark, Cliff Swallow, Chimney Swift, Savannah Sparrow, Common Nighthawk, Tree Swallow, Bank Swallow, Red-winged Blackbird, Yellow-rumped Warbler, and Swainson's Thrush. After collecting images, we combined the Barn and Cliff Swallows into a single category due to their strong visual resemblance. Additionally, the Horned Lark, Savannah Sparrow, and Swainson's Thrush were excluded from the study because of insufficient image data. Consequently, the final classification model was designed to differentiate among 7 bird species. For each category, 200 images were used, 160 for training, 20 for validation, and 20 for testing.

\noindent For the benchmark models, we used 180 images per class for training and 20 images per class for testing, without a separate validation data set since it was not required for these classifiers. A grid search was performed to identify the best hyperparameters for each model. For the SVM, the optimal settings were: C = 10, gamma = scale, and kernel = rbf. The best configuration for the RF model included: bootstrap = True, max depth = 20, minimum samples per leaf = 1, minimum samples per split = 2, and 300 estimators. For the KNN model, the top-performing parameters were: algorithm = auto, distance metric = Manhattan, number of neighbors = 3, and weights = distance. To address memory constraints, we reduced the input image size to 128×128×3 for all benchmark classifiers.

\floatplacement{figure}{H}
\begin{figure}
     \centering
     \begin{subfigure}[h]{0.495\textwidth}
         \centering
         \includegraphics[width = 8 cm, height=5cm]{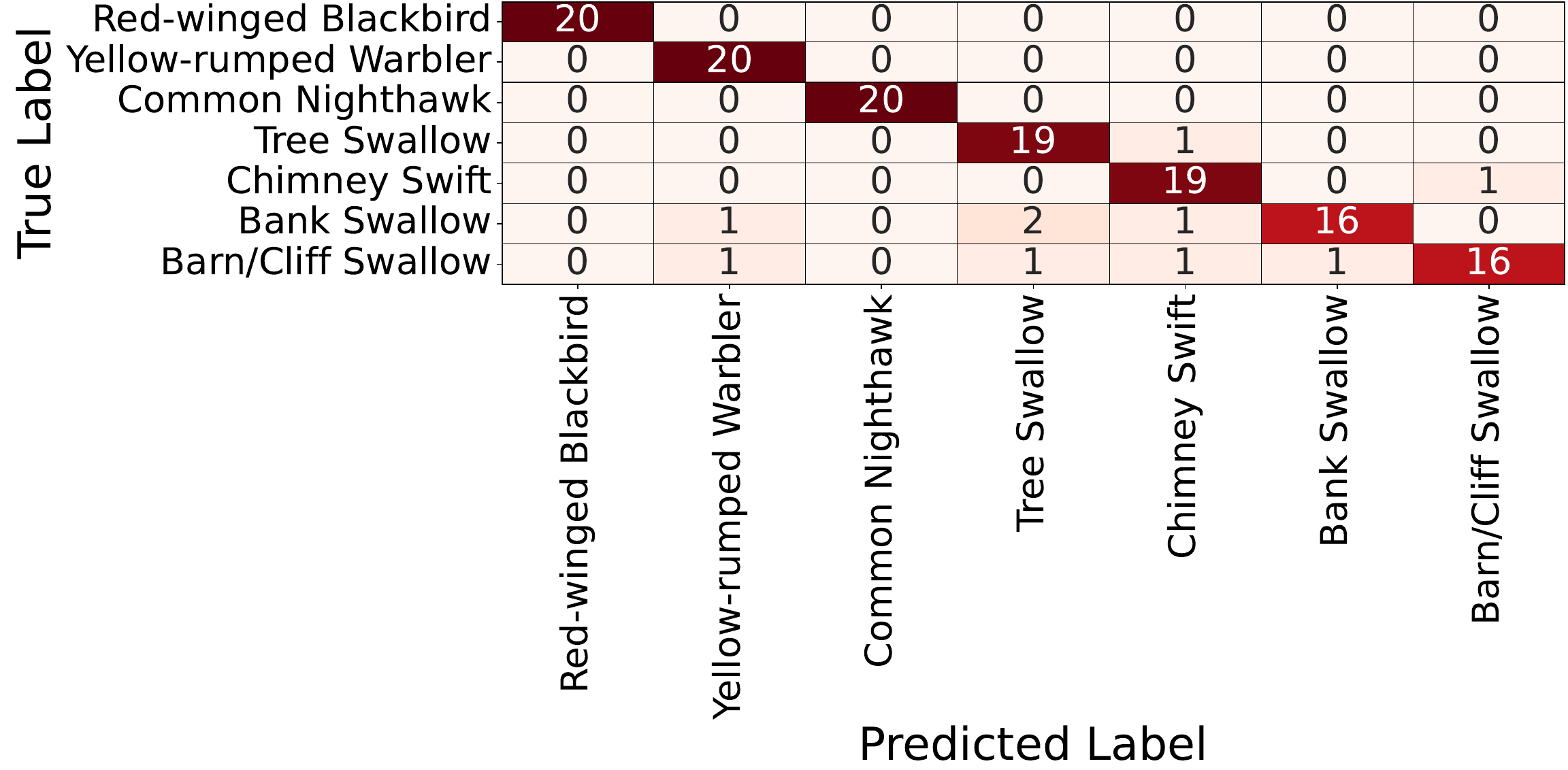}
         \caption{Confusion matrix of the CNN model}
         \label{CNN-Small}
     \end{subfigure}
     \hfill
     \begin{subfigure}[h]{0.495\textwidth}
         \centering
         \includegraphics[width = 8 cm, height=5cm]{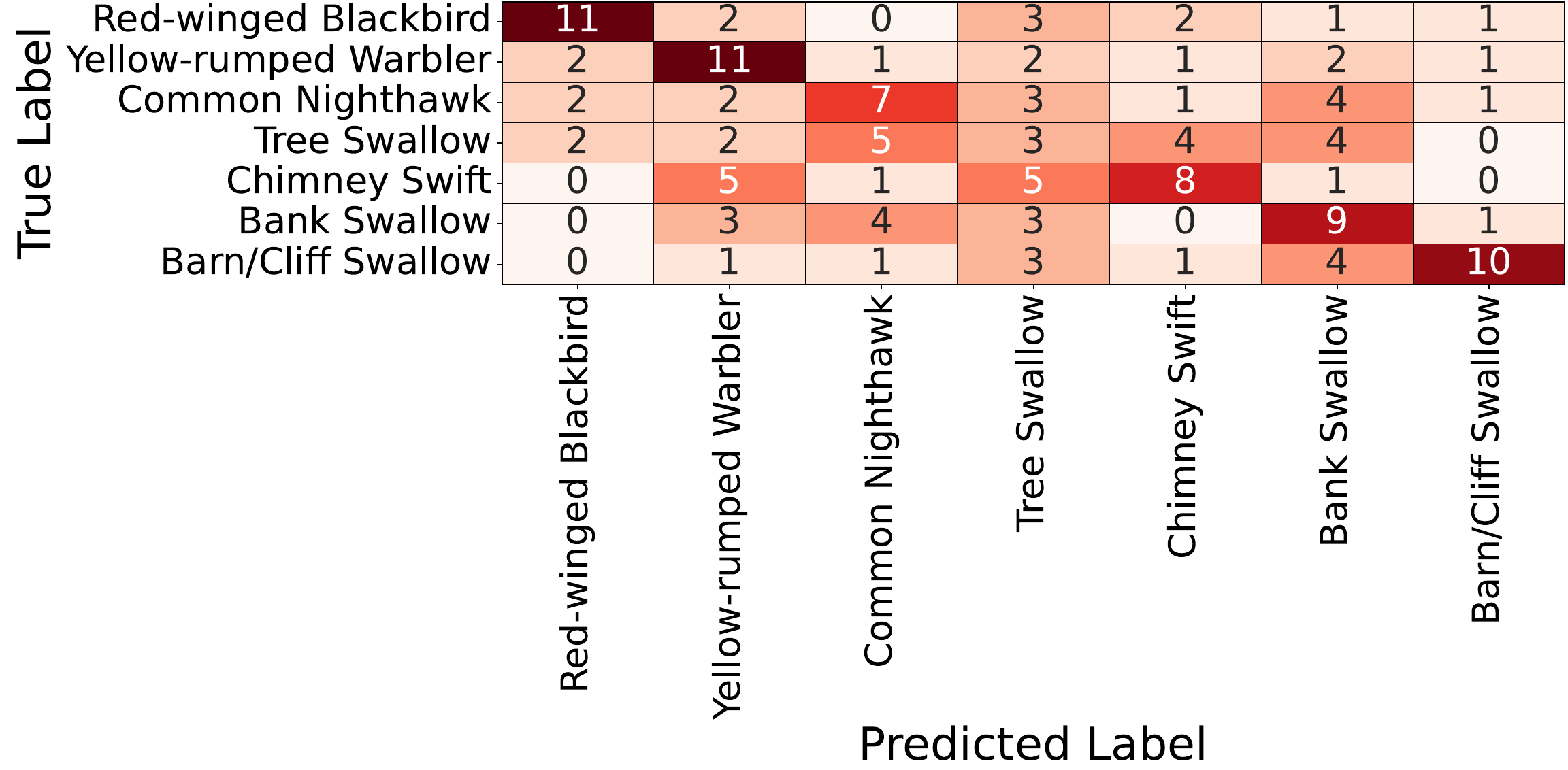}
         \caption{Confusion matrix of the SVM model}
         \label{SVM-Small}
     \end{subfigure}
\vspace{3 mm} 

     \centering
     \begin{subfigure}[h]{0.495\textwidth}
         \centering
         \includegraphics[width = 8 cm, height=5cm]{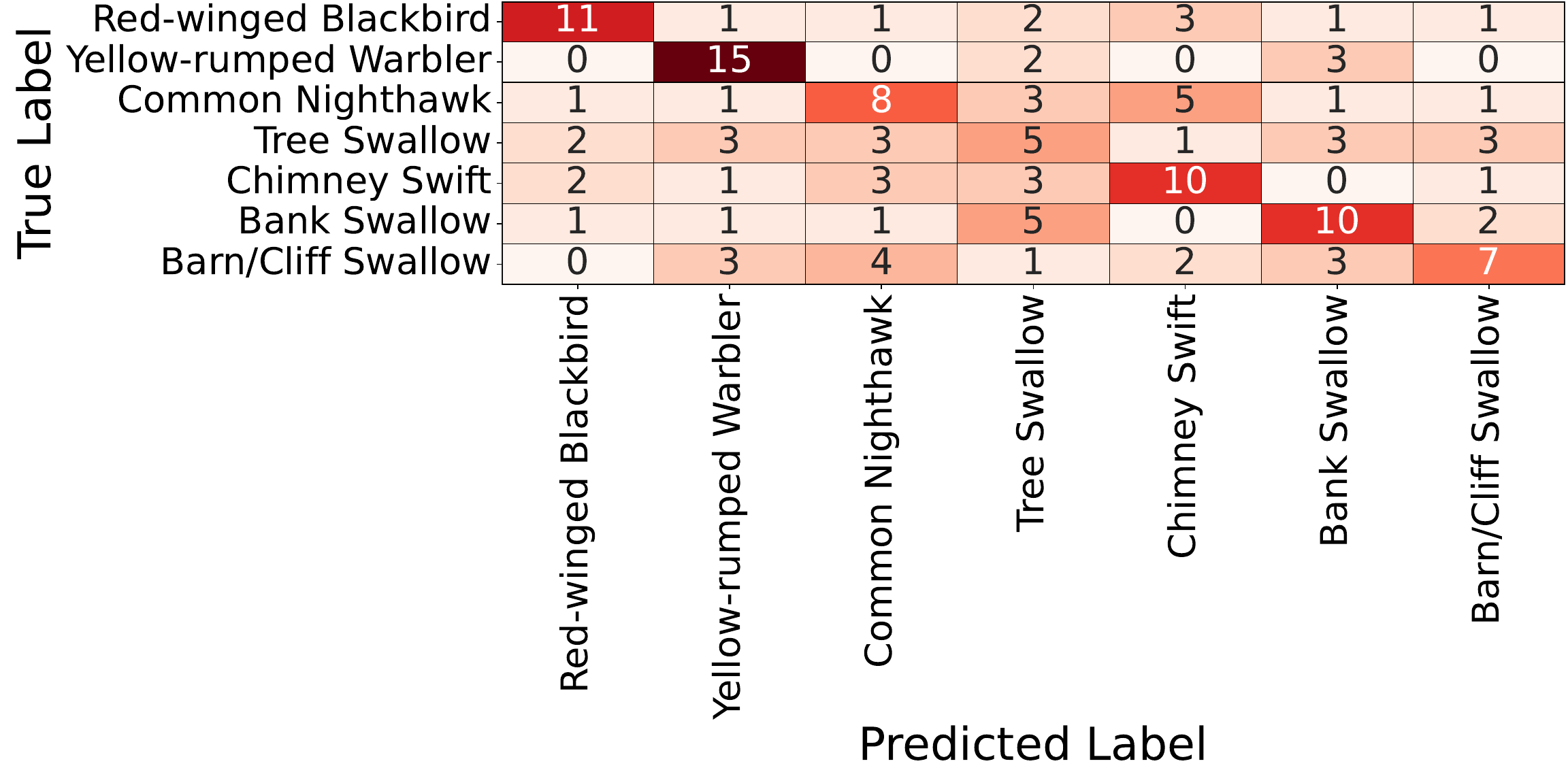}
         \caption{Confusion matrix of the RF model}
         \label{RF-Small}
     \end{subfigure}
     \hfill
     \begin{subfigure}[h]{0.495\textwidth}
         \centering
         \includegraphics[width = 8 cm, height=5cm]{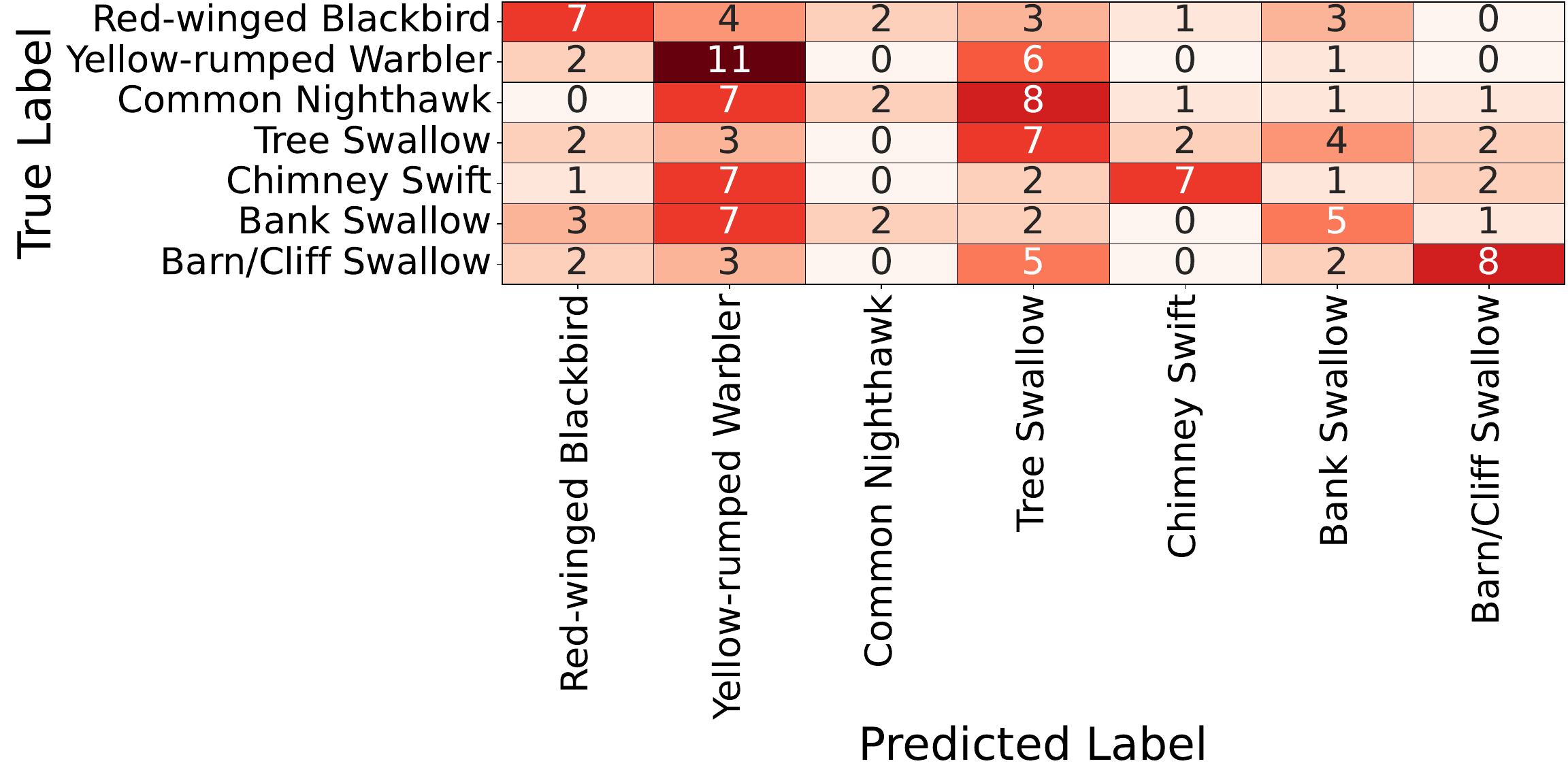}
         \caption{Confusion matrix of the KNN model}
         \label{KNN-Small}
     \end{subfigure}
     \caption{Comparison of confusion matrices from four different classification models: CNN, SVM, RF, and KNN for small size birds (classifier 5).}
     \label{Small}
    
\end{figure}

\renewcommand{\arraystretch}{0.65}
\begin{table}[H]
\centering
\caption{Class-wise precision, recall, and F1-score for CNN, SVM, RF, and KNN models for small size birds (classifier 5).}
\begin{tabularx}{\textwidth}{l *{12}{>{\centering\arraybackslash}X}}
\toprule
\textbf{Class} 
& \multicolumn{3}{c}{\textbf{CNN}} 
& \multicolumn{3}{c}{\textbf{SVM}} 
& \multicolumn{3}{c}{\textbf{RF}} 
& \multicolumn{3}{c}{\textbf{KNN}} \\
& \rotatebox{90}{Precision} & \rotatebox{90}{Recall} & \rotatebox{90}{F1-score} 
& \rotatebox{90}{Precision} & \rotatebox{90}{Recall} & \rotatebox{90}{F1-score} 
& \rotatebox{90}{Precision} & \rotatebox{90}{Recall} & \rotatebox{90}{F1-score} 
& \rotatebox{90}{Precision} & \rotatebox{90}{Recall} & \rotatebox{90}{F1-score} \\
\midrule
Red-winged Blackbird      & 1.00 & 1.00 & 1.00 & 0.65 & 0.55 & 0.59 & 0.65 & 0.55 & 0.59 & 0.41 & 0.35 & 0.38 \\
Yellow-rumped Warbler     & 0.91 & 1.00 & 0.95 & 0.42 & 0.55 & 0.48 & 0.60 & 0.75 & 0.67 & 0.26 & 0.55 & 0.35 \\
Common Nighthawk          & 1.00 & 1.00 & 1.00 & 0.37 & 0.35 & 0.36 & 0.40 & 0.40 & 0.40 & 0.33 & 0.10 & 0.15 \\
Tree Swallow              & 0.86 & 0.95 & 0.90 & 0.14 & 0.15 & 0.14 & 0.24 & 0.25 & 0.24 & 0.21 & 0.35 & 0.26 \\
Chimney Swift             & 0.86 & 0.95 & 0.90 & 0.47 & 0.40 & 0.43 & 0.48 & 0.50 & 0.49 & 0.64 & 0.35 & 0.45 \\
Bank Swallow              & 0.94 & 0.80 & 0.86 & 0.36 & 0.45 & 0.40 & 0.48 & 0.50 & 0.49 & 0.29 & 0.25 & 0.27 \\
Barn/Cliff Swallow        & 0.94 & 0.80 & 0.86 & 0.71 & 0.50 & 0.59 & 0.47 & 0.35 & 0.40 & 0.57 & 0.40 & 0.47 \\
\midrule
\textbf{Accuracy} & \multicolumn{3}{c}{\textbf{92.86\%}} 
                  & \multicolumn{3}{c}{\textbf{42.14\%}} 
                  & \multicolumn{3}{c}{\textbf{47.14\%}} 
                  & \multicolumn{3}{c}{\textbf{33.57\%}} \\
\bottomrule
\end{tabularx}
\label{tab:Small}
\end{table}

\noindent Figure \ref{Small} presents the confusion matrices for all models using their optimal hyperparameters. Among them, the CNN model demonstrates the highest classification accuracy. Detailed performance metrics, including precision, recall, F1 score, and accuracy for each model and class, are provided in Table \ref{tab:Small}. The results indicate that the CNN significantly outperformed the benchmark models in the small bird classification task, achieving the highest overall accuracy of 92.86\%. In comparison, the RF model reached 47.14\% accuracy, the SVM achieved 42.14\%, and the KNN model recorded the lowest accuracy at 33.57\%.

\subsection{The Performance of the Unified Classification Approach}
\label{PUCA}

\begin{figure}[H]
    \centering
        \includegraphics[width=16.5 cm, height = 8 cm]{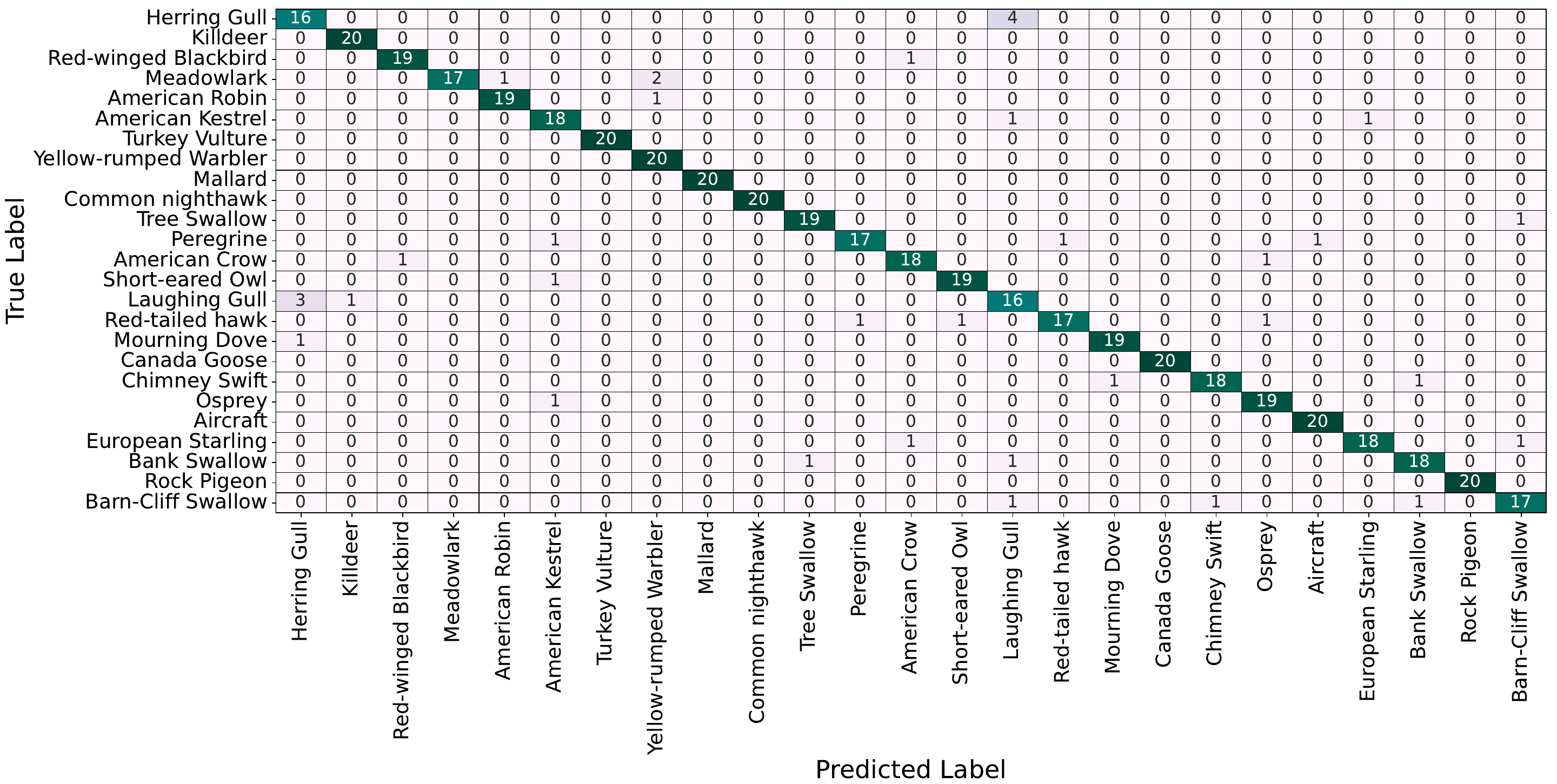}
        \caption{Confusion matrix of CNN for identifying bird species and aircraft (classifier 6).}
        \label{CNN-Unified}
\end{figure}

In UCA, we used a single classifier to identify both bird species and aircraft. Out of the 33 bird species that are most frequently involved in bird strikes, we could not find enough images for the following species: Cattle Egret, Horned Lark, Savannah Sparrow, Swainson's Thrush, Barn Owl, and Pacific Golden-Plover. In addition, some bird species look very similar to each other. For example, Barn Swallow and Cliff Swallow, Eastern and Western Meadowlark, and Ring-billed Gull and Laughing Gull. To avoid confusion, we combined each pair into a single class. As a result, the dataset includes 24 bird classes (images collected from \cite{CL}) and 1 aircraft class (images collected from \cite{aircraft_dataset, helicopter_dataset, drone_dataset}), with 200 images per class, making a total of 5000 images. The dataset was split into 80\% training (160 images per class), 10\% validation (20 images per class), and 10\% testing (20 images per class). The model was trained to classify each image as one of the 24 bird types or as an aircraft. The confusion matrix in Fig. \ref{CNN-Unified} shows how well the model performed on the test set. The precision, recall, and F1-score for each class are provided in Table \ref{tab:Unified}. These results show how accurately the model can distinguish between bird species and aircraft using a single unified classification approach. 

\renewcommand{\arraystretch}{0.6}
\begin{table}[H]
\centering
\caption{Class-wise precision, recall, and F1-score of CNN for identifying bird species and aircraft (classifier 6).}
\begin{tabular}{lccc}
\toprule
\textbf{Class} & \textbf{Precision} & \textbf{Recall} & \textbf{F1-score} \\
\midrule
Herring Gull            & 0.80 & 0.80 & 0.80 \\
Killdeer                & 0.95 & 1.00 & 0.98 \\
Red-winged Blackbird    & 0.95 & 0.95 & 0.95 \\
Meadowlark              & 1.00 & 0.85 & 0.92 \\
American Robin          & 0.95 & 0.95 & 0.95 \\
American Kestrel        & 0.86 & 0.90 & 0.88 \\
Turkey Vulture          & 1.00 & 1.00 & 1.00 \\
Yellow-Rumped Warbler   & 0.87 & 1.00 & 0.93 \\
Mallard                 & 1.00 & 1.00 & 1.00 \\
Common Nighthawk        & 1.00 & 1.00 & 1.00 \\
Tree Swallow            & 0.95 & 0.95 & 0.95 \\
Peregrine               & 0.94 & 0.85 & 0.89 \\
American Crow           & 0.90 & 0.90 & 0.90 \\
Short-eared Owl         & 0.95 & 0.95 & 0.95 \\
Laughing Gull           & 0.70 & 0.80 & 0.84 \\
Red-tailed Hawk         & 0.94 & 0.85 & 0.89 \\
Mourning Dove           & 0.95 & 0.95 & 0.95 \\
Canada Goose            & 1.00 & 1.00 & 1.00 \\
Chimney Swift           & 0.95 & 0.90 & 0.92 \\
Osprey                  & 0.90 & 0.95 & 0.93 \\
Aircraft                & 0.95 & 1.00 & 0.98 \\
European Starling       & 0.95 & 0.90 & 0.92 \\
Bank Swallow            & 0.90 & 0.90 & 0.90 \\
Rock Pigeon             & 1.00 & 1.00 & 1.00 \\
Barn/Cliff Swallow      & 0.89 & 0.85 & 0.87 \\

\midrule
\textbf{Accuracy} & \multicolumn{3}{c}{\textbf{92.80\%}} \\
\bottomrule
\end{tabular}
\label{tab:Unified}
\end{table}

\subsection{Comparison of the Two Approaches}
\label{Comparison}

To assess the effectiveness of the two bird classification strategies, we compared two pipelines: the CCA and the UCA. Because the CCA operates as a sequential decision pipeline, a correct end-to-end prediction requires that all intermediate classification stages make correct decisions. Specifically, for a bird image to be correctly classified at the species level, the system must: (i) correctly identify the presence of a bird (classifier 1), (ii) correctly classify the bird size (classifier 2), and (iii) correctly identify the bird species using the size-specific classifier (classifiers 3, 4, 5). Consequently, the overall accuracy of the CCA is determined by the conditional correctness of each stage rather than by their standalone accuracies.

\noindent
Let $R_1^{\text{bird}}$ denote the recall of the bird-versus-aircraft classifier (classifier 1) for the bird class, $R_2^{i}$ denote the recall of the size classifier (classifier 2) for size category $i \in \{\text{Small}, \text{Medium}, \text{Large}\}$, and $A_3^{i}$ denote the species classification accuracy for size category $i$ (classifiers 3, 4, 5), evaluated on correctly routed samples. The end-to-end accuracy of the CCA for bird images can then be expressed as:
\begin{equation}
\text{Accuracy}_{\text{CCA}}
=
R_1^{\text{bird}}
\sum_{i \in \{\text{S,M,L}\}}
P(i)\, R_2^{i}\, A_3^{i},
\label{Accuracy_CCA}
\end{equation}
where $P(i)$ represents the proportion of bird images belonging to size category $i$ in the dataset. Using the experimental results obtained in this study, the parameters in \eqref{Accuracy_CCA} were instantiated with the corresponding recall and accuracy values of each stage. Applying this formulation yields an overall end-to-end accuracy of 90.77\% for the CCA.

\noindent
This formulation reflects the fact that a misclassification at any intermediate stage propagates downstream and prevents correct species identification. In particular, incorrect size classification routes the input to an inappropriate species classifier, making correct prediction unlikely regardless of the downstream classifier’s standalone accuracy. Therefore, size-specific recalls play a critical role in determining the overall system performance.

\noindent
In contrast, the UCA employs a single end-to-end classifier that directly predicts the bird species or aircraft class without intermediate decisions (classifier 6). As a result, the overall accuracy of the UCA is simply given by:
\begin{equation}
\text{Accuracy}_{\text{UCA}} = A_6,
\end{equation}
where $A_6$ denotes the accuracy of the unified classifier.
Using the experimental results, the accuracy of the UCA was found to be $A_6 = 92.80\%$.

\noindent
Although the UCA achieves a slightly higher overall accuracy, the CCA provides a modular and interpretable structure that allows performance analysis and improvement at individual stages. However, this modularity also implies that the final accuracy of the CCA is inherently dependent on the combined performance of all constituent classifiers.

\subsection{Robustness Study}
\label{RS}

To evaluate the robustness of the bird species classifiers under various real-world environmental distortions, we simulated different conditions on the test images: rain, snow, sensor noise, and darkness. Each condition was applied in varying intensities, and the performance of classifiers (classifiers 3, 4, and 5) was measured in terms of accuracy. 

\begin{itemize}
    \item \textbf{Rain Simulation:} Rainy conditions were simulated by drawing thin, light gray slanted lines (raindrops) over the original image to mimic falling rain. The number of raindrops depended on the specified rain intensity (in percent), with more drops added for heavier rain. These lines were slightly blurred using a Gaussian filter to make them look more natural and less sharp, resembling real rain captured by a camera. Finally, the rain effect was softly blended into the original image using transparency, so the rain appeared realistic without completely obscuring the scene. Figure \ref{fig:rain_levels} illustrates the rain simulation applied to a sample image, ranging from 0\% to 50\% rain intensity. 

\begin{figure}[H]
    \centering
    \begin{subfigure}[b]{1.8cm}
        \includegraphics[width=1.8cm, height=1.25cm]{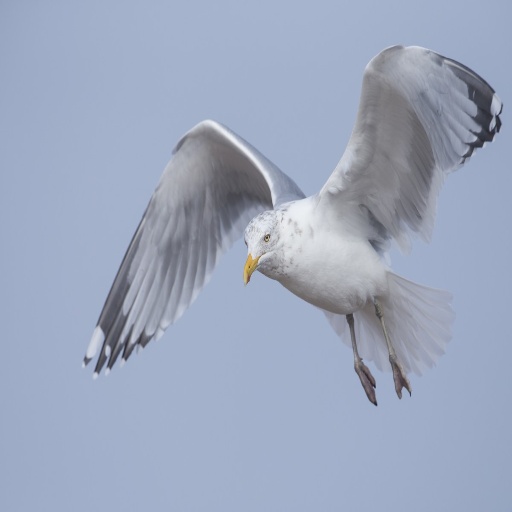}
        \caption{0\%}
    \end{subfigure}\hspace{0.1cm}%
    \begin{subfigure}[b]{1.8cm}
        \includegraphics[width=1.8cm, height=1.25cm]{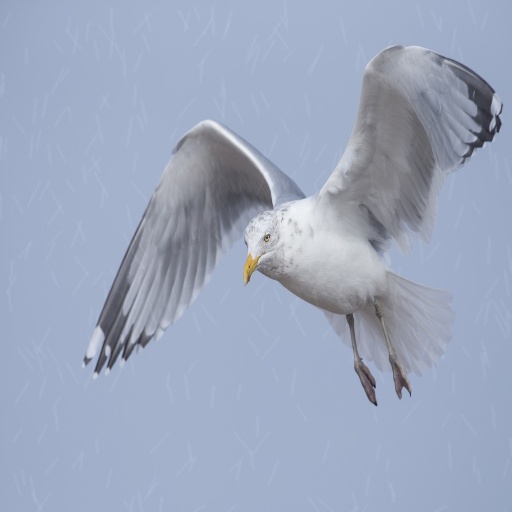}
        \caption{10\%}
    \end{subfigure}\hspace{0.1cm}%
    \begin{subfigure}[b]{1.8cm}
        \includegraphics[width=1.8cm, height=1.25cm]{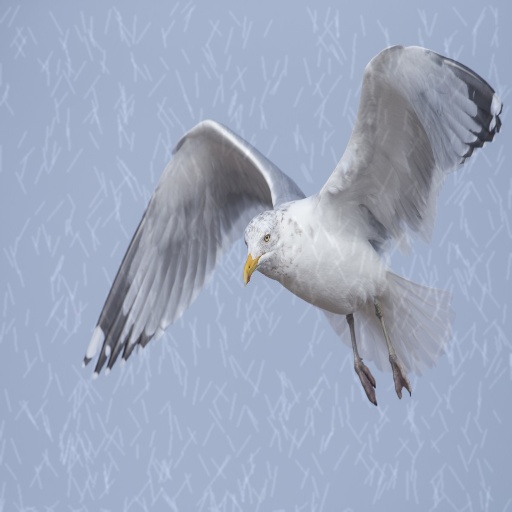}
        \caption{20\%}
    \end{subfigure}\hspace{0.1cm}%
    \begin{subfigure}[b]{1.8cm}
        \includegraphics[width=1.8cm, height=1.25cm]{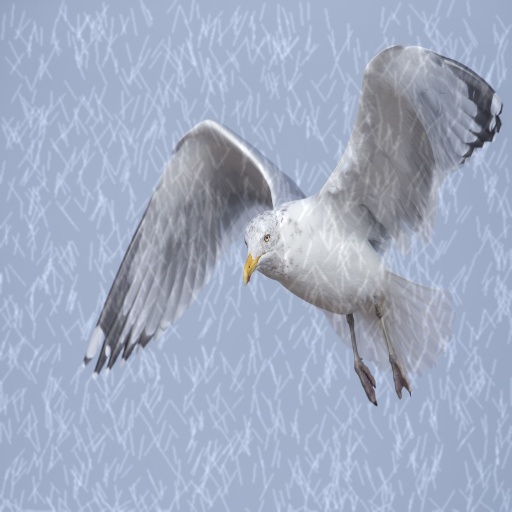}
        \caption{30\%}
    \end{subfigure}\hspace{0.1cm}%
    \begin{subfigure}[b]{1.8cm}
        \includegraphics[width=1.8cm, height=1.25cm]{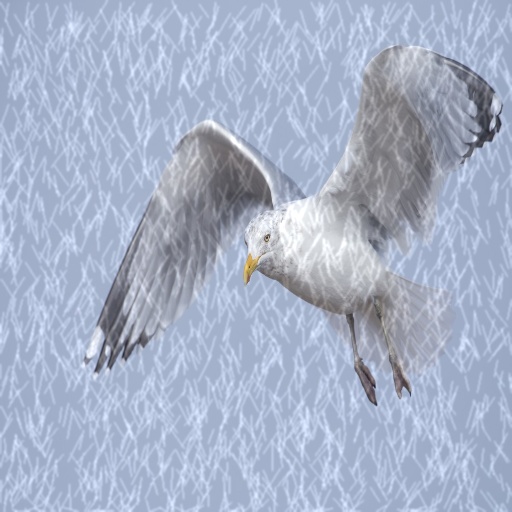}
        \caption{40\%}
    \end{subfigure}\hspace{0.1cm}%
    \begin{subfigure}[b]{1.8cm}
        \includegraphics[width=1.8cm, height=1.25cm]{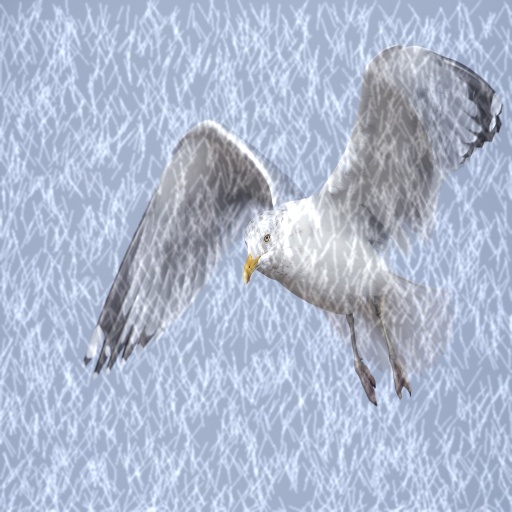}
        \caption{50\%}
    \end{subfigure}
    \caption{Rain simulation levels from 0\% to 50\%.}
    \label{fig:rain_levels}
\end{figure}

Figure \ref{R} showed the effect of these simulated rainy conditions on the classification accuracy of the CNN models for large, medium, and small bird species. We observed that all three classifiers performed well under light rain, with accuracy remaining high at 5\% rain intensity: 96.83\% for large birds, 95.42\% for medium birds, and 92.14\% for small birds. However, as the rain intensity increased, the performance of all classifiers gradually declined. At 50\% intensity, accuracy dropped to 59.13\% for large, 59.58\% for medium, and 60.71\% for small birds.

\begin{figure}[H]
    \centering
        \includegraphics[width=15.5cm, height = 5.25 cm]{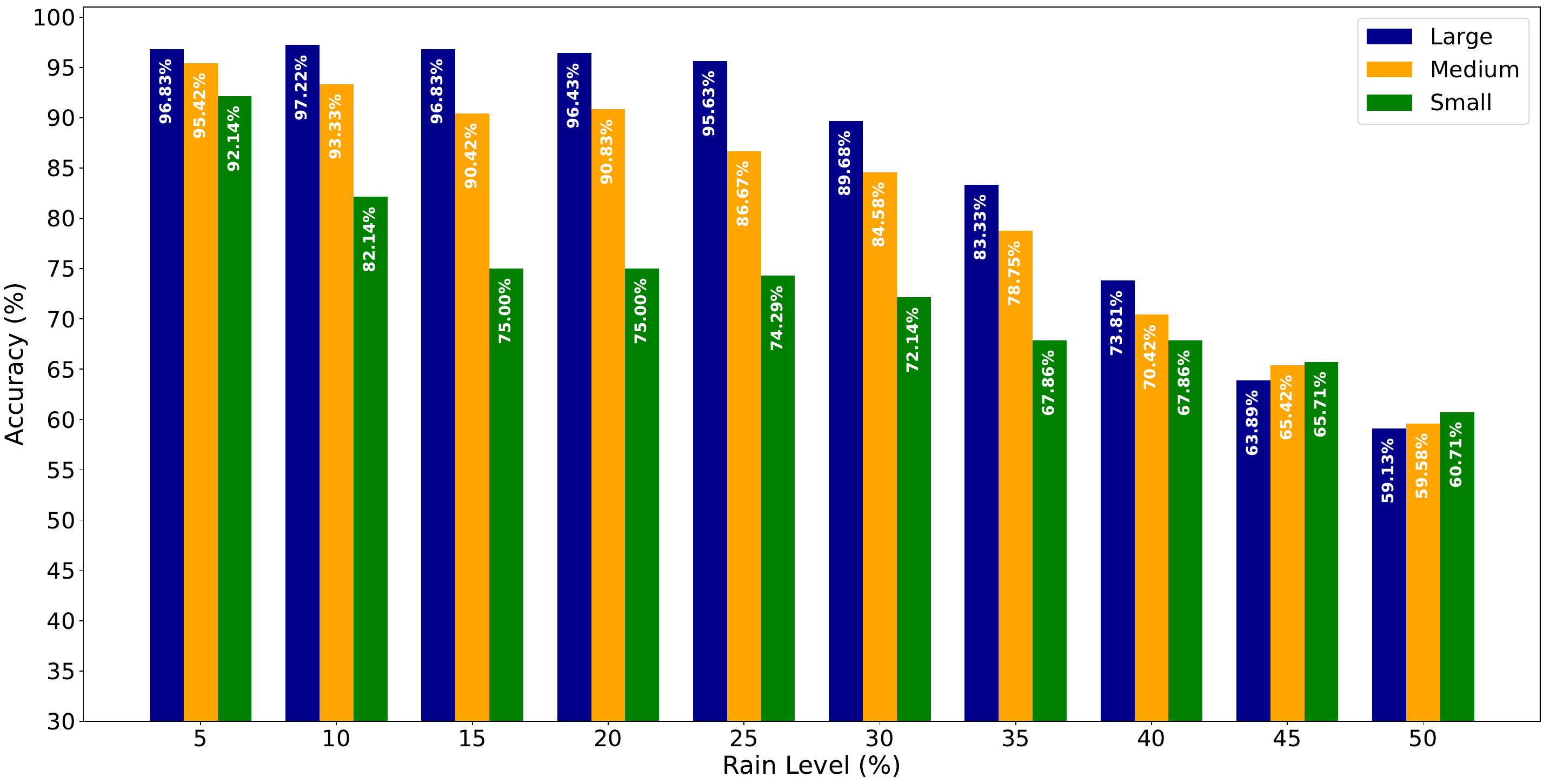}
        \caption{The effect of rain level on the accuracy of the CNN classifiers.}
        \label{R}
\end{figure}

    \item \textbf{Snow Simulation:} Snowy conditions were simulated by adding white circular spots (snowflakes) at random positions across the image to represent falling snow. The number of flakes depended on the snow intensity level, higher intensity meant more flakes. Each flake was assigned a slightly random size and then blurred using a Gaussian filter to create a soft, fluffy appearance like real snowflakes in the air. The snow effect was then blended into the original image using adjustable transparency, with heavier snowfall appearing more opaque. This created a realistic visual simulation of snowfall over the bird images. Figure \ref{fig:snow_levels} displays the visual effects of simulated snowfall on a bird image, ranging from 0\% to 50\% snow intensity. 

\begin{figure}[H]
    \centering
    \begin{subfigure}[b]{1.8cm}
        \includegraphics[width=1.8cm, height=1.25cm]{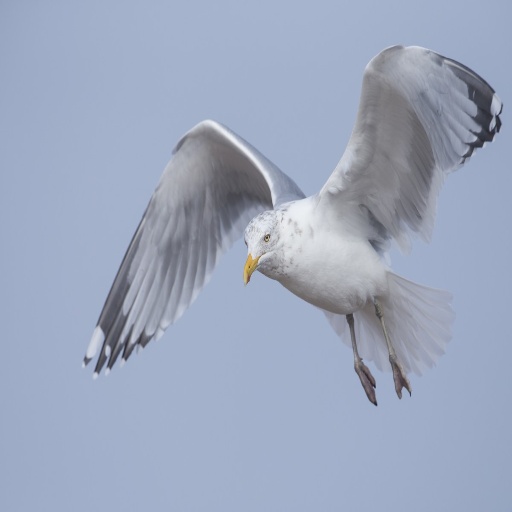}
        \caption{0\% Snow}
    \end{subfigure}\hspace{0.1cm}%
    \begin{subfigure}[b]{1.8cm}
        \includegraphics[width=1.8cm, height=1.25cm]{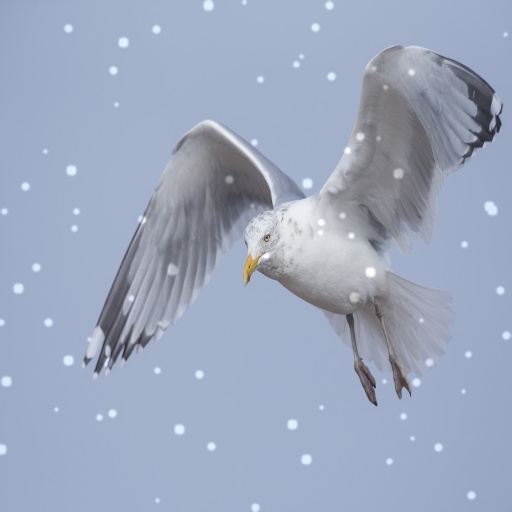}
        \caption{10\% Snow}
    \end{subfigure}\hspace{0.1cm}%
    \begin{subfigure}[b]{1.8cm}
        \includegraphics[width=1.8cm, height=1.25cm]{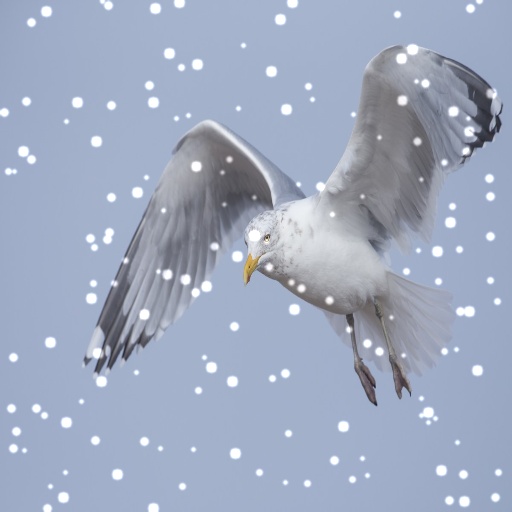}
        \caption{20\% Snow}
    \end{subfigure}\hspace{0.1cm}%
    \begin{subfigure}[b]{1.8cm}
        \includegraphics[width=1.8cm, height=1.25cm]{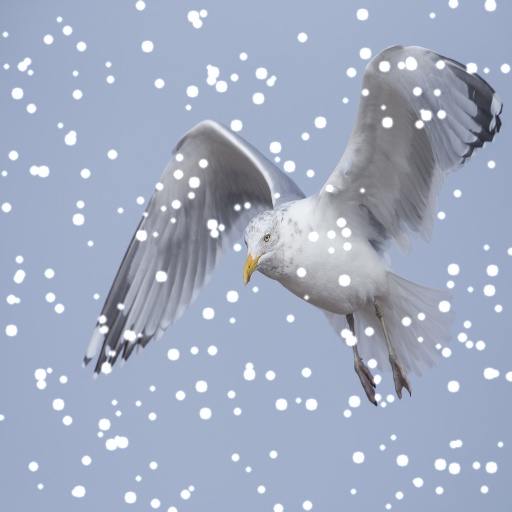}
        \caption{30\% Snow}
    \end{subfigure}\hspace{0.1cm}%
    \begin{subfigure}[b]{1.8cm}
        \includegraphics[width=1.8cm, height=1.25cm]{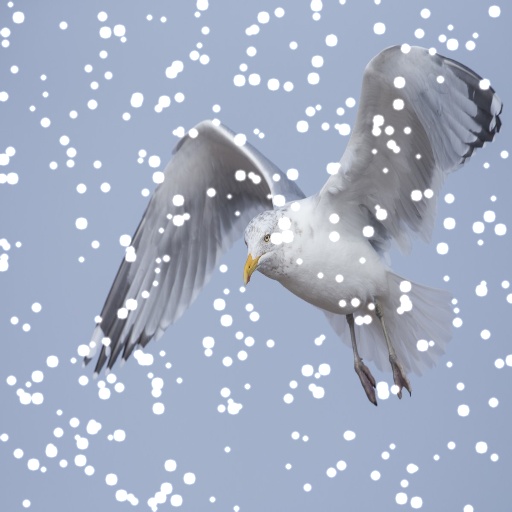}
        \caption{40\% Snow}
    \end{subfigure}\hspace{0.1cm}%
    \begin{subfigure}[b]{1.8cm}
        \includegraphics[width=1.8cm, height=1.25cm]{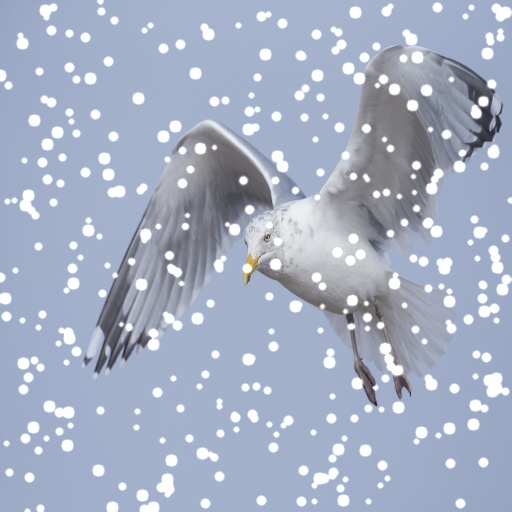}
        \caption{50\% Snow}
    \end{subfigure}
    \caption{Snow simulation levels from 0\% to 50\%.}
    \label{fig:snow_levels}
\end{figure}

Figure \ref{S} quantified the impact of snow on classification performance for large, medium, and small bird species. We observed that the classifiers performed reasonably well under light snow (e.g., 5\% intensity), maintaining accuracies above 89\% for large, medium, and small birds. However, as snow intensity increased, classification accuracy declined steadily across all size categories. At 50\% snow, the accuracies fell to 46.83\% for large birds, 31.67\% for medium birds, and 50.71\% for small birds. Notably, the medium bird classifier showed the sharpest performance drop; it was particularly vulnerable to partial occlusions from snowflakes. The small bird classifier also declined, though slightly less abruptly. These results indicated that moderate to heavy snowfall could significantly degrade visual features critical for species identification.

\begin{figure}[H]
    \centering
        \includegraphics[width=15.5cm, height = 6 cm]{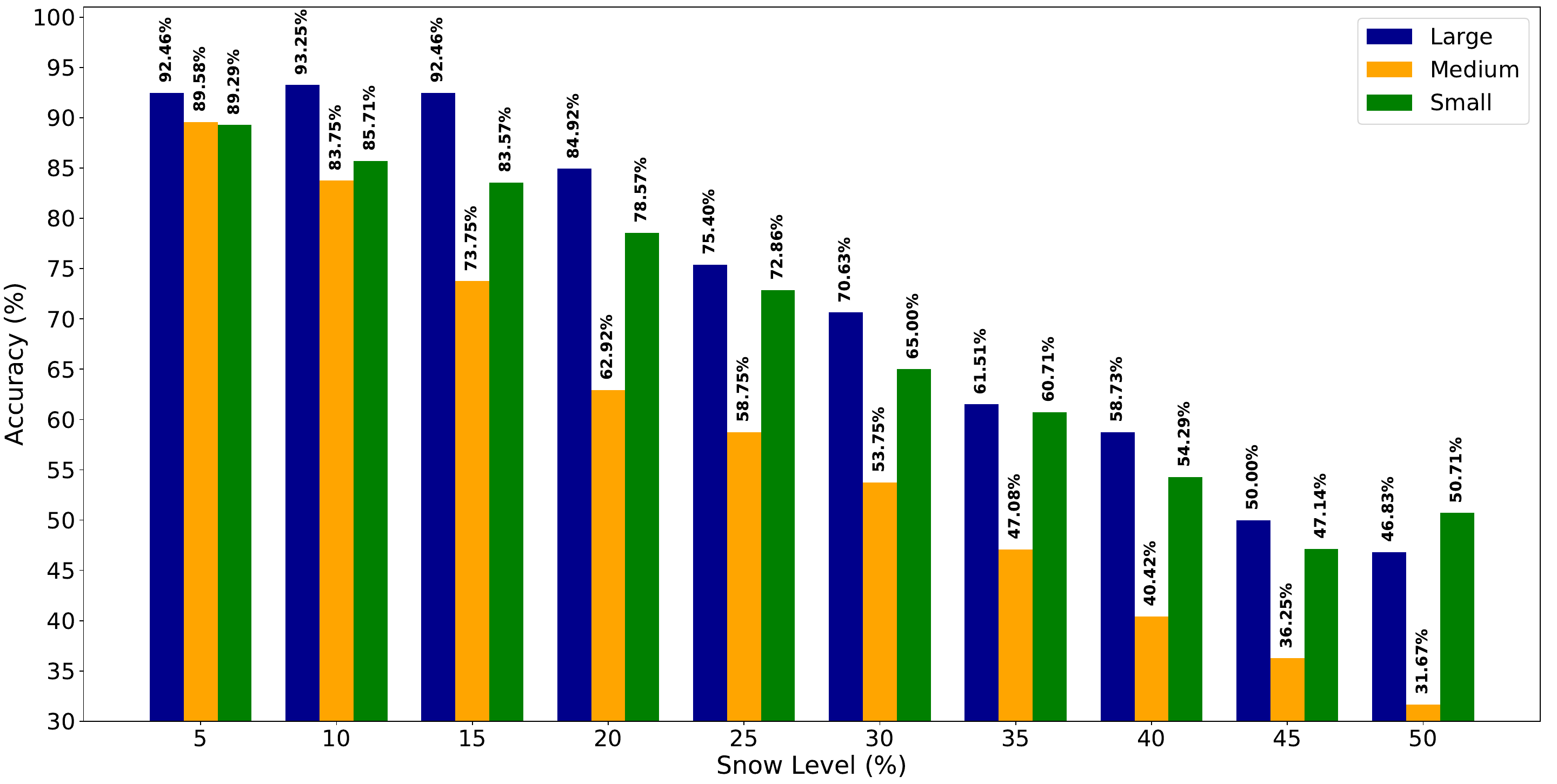}
        \caption{The effect of snow level on the accuracy of the CNN classifiers.}
        \label{S}
\end{figure}

    \item \textbf{Sensor Noise Simulation:} Sensor noise was simulated by adding Gaussian noise to the pixel values of each image to represent imperfections from camera sensors. The noise was generated from a normal distribution with a mean of zero and varying standard deviations, where higher values produced stronger noise effects. This noise appeared as subtle grain or static across the image, especially in uniform areas such as the sky or background. Figure \ref{fig:gaussian_noise_levels} visually demonstrates the effect of increasing Gaussian noise levels on a sample image, with standard deviations ranging from $\sigma = 0.00$ to $\sigma = 0.40$. As the noise level increases, the image becomes progressively more degraded. 

\begin{figure}[H]
    \centering
    \begin{subfigure}[b]{1.75cm}
        \includegraphics[width=1.7cm, height=1.25cm]{snow_0percent.jpg}
        \caption{$\sigma=0.00$}
    \end{subfigure}\hspace{0.01cm}%
    \begin{subfigure}[b]{1.75cm}
        \includegraphics[width=1.7cm, height=1.25cm]{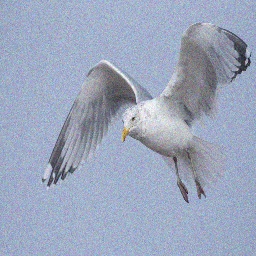}
        \caption{$\sigma=0.05$}
    \end{subfigure}\hspace{0.01cm}%
    \begin{subfigure}[b]{1.75cm}
        \includegraphics[width=1.7cm, height=1.25cm]{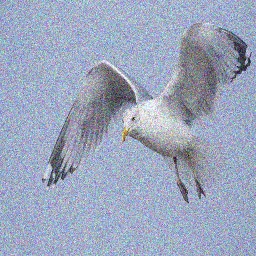}
        \caption{$\sigma=0.10$}
    \end{subfigure}\hspace{0.01cm}%
    \begin{subfigure}[b]{1.75cm}
        \includegraphics[width=1.7cm, height=1.25cm]{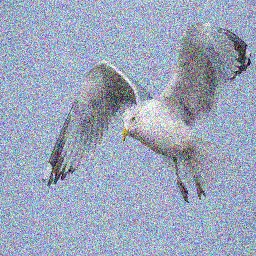}
        \caption{$\sigma=0.15$}
    \end{subfigure}\hspace{0.01cm}%
    \begin{subfigure}[b]{1.75cm}
        \includegraphics[width=1.7cm, height=1.25cm]{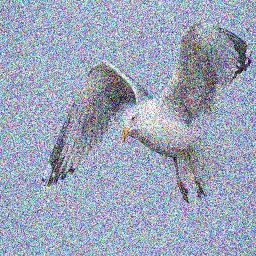}
        \caption{$\sigma=0.20$}
    \end{subfigure}\hspace{0.01cm}%
    \begin{subfigure}[b]{1.75cm}
        \includegraphics[width=1.7cm, height=1.25cm]{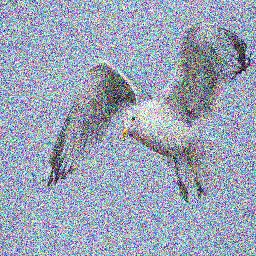}
        \caption{$\sigma=0.25$}
    \end{subfigure}\hspace{0.01cm}%
    \begin{subfigure}[b]{1.75cm}
        \includegraphics[width=1.7cm, height=1.25cm]{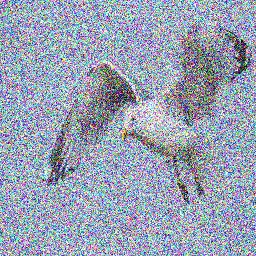}
        \caption{$\sigma=0.30$}
    \end{subfigure}\hspace{0.01cm}%
    \begin{subfigure}[b]{1.75cm}
        \includegraphics[width=1.7cm, height=1.25cm]{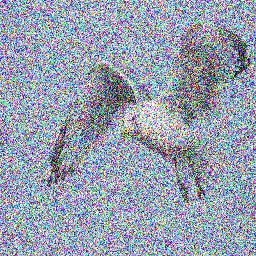}
        \caption{$\sigma=0.35$}
    \end{subfigure}\hspace{0.01cm}%
    \begin{subfigure}[b]{1.75cm}
        \includegraphics[width=1.7cm, height=1.25cm]{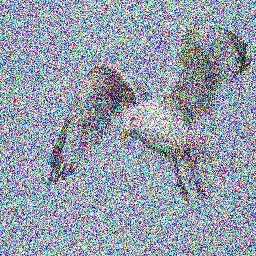}
        \caption{$\sigma=0.40$}
    \end{subfigure}
    \caption{Sensor noise simulation levels from $\sigma = 0.00$ to $\sigma = 0.40$.}
    \label{fig:gaussian_noise_levels}
\end{figure}

Figure \ref{S-N} quantitatively presented how these noise levels affected the performance of the CNN classifiers across large, medium, and small bird species. We observed that the classification accuracy consistently declined as the noise standard deviation increased. For instance, when $\sigma = 0.05$, the classifiers maintained high accuracy, with 94.44\% for large birds, 95\% for medium birds, and 88.57\% for small birds. However, at $\sigma = 0.4$, the accuracy dropped substantially to 59.13\%, 59.58\%, and 60.71\%, respectively.

\begin{figure}[H]
    \centering
        \includegraphics[width=15.5cm, height = 5.25 cm]{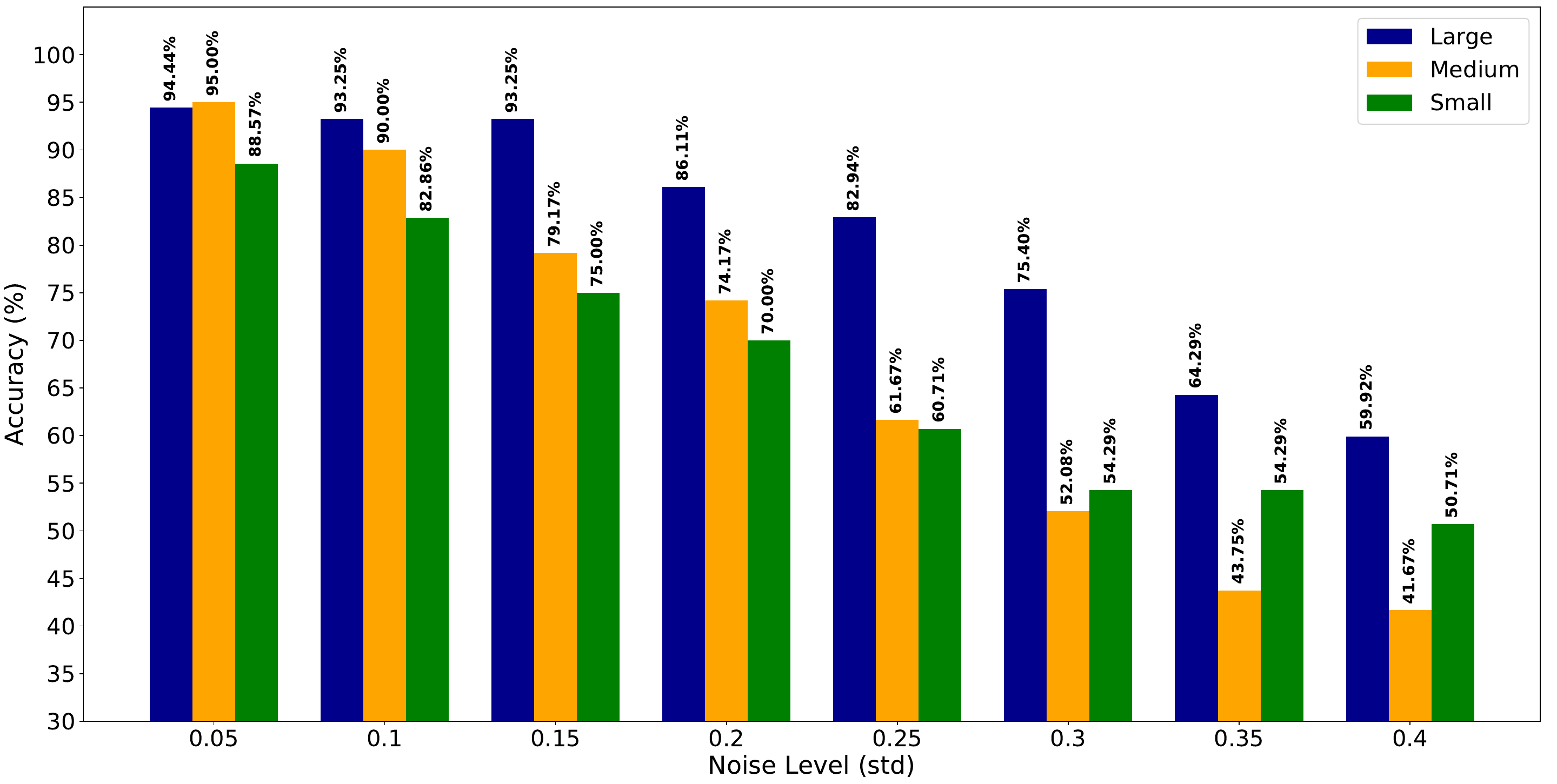}
        \caption{The effect of sensor noise level on the accuracy of the CNN classifiers.}
        \label{S-N}
\end{figure}

    \item \textbf{Darkness Simulation:} Darkness or low-light conditions were simulated by reducing the brightness of the original images. This was done by multiplying the pixel values by a brightness factor less than one, with lower values resulting in darker images. Multiple levels of darkness were tested, ranging from slightly dimmed to severely darkened, to mimic different real-world scenarios such as cloudy environments, evening light, or poor camera exposure. This allowed us to evaluate how reduced visibility affects the classifier's performance under low-light conditions. Figure \ref{fig:darkness_levels} illustrates how darkness was simulated by progressively reducing the brightness of a sample image, from full brightness (100\%) down to 30\%.

\begin{figure}[H]
    \centering
    \begin{subfigure}[b]{1.75cm}
        \includegraphics[width=1.7cm, height=1.25cm]{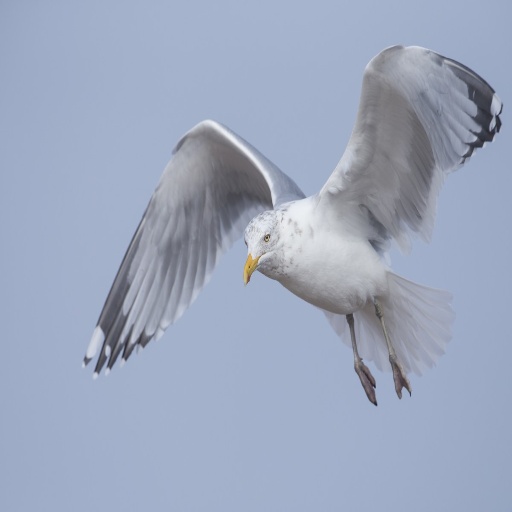}
        \caption{100\%}
    \end{subfigure}\hspace{0.1cm}%
    \begin{subfigure}[b]{1.75cm}
        \includegraphics[width=1.7cm, height=1.25cm]{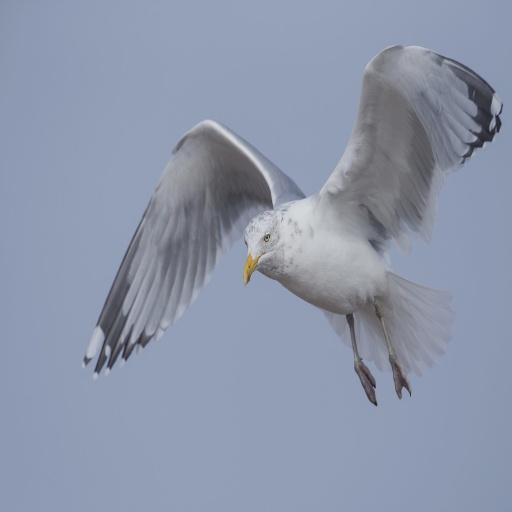}
        \caption{90\%}
    \end{subfigure}\hspace{0.1cm}%
    \begin{subfigure}[b]{1.75cm}
        \includegraphics[width=1.7cm, height=1.25cm]{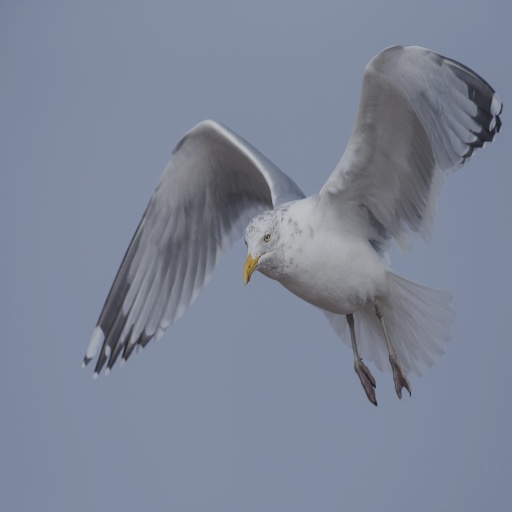}
        \caption{80\%}
    \end{subfigure}\hspace{0.1cm}%
    \begin{subfigure}[b]{1.75cm}
        \includegraphics[width=1.7cm, height=1.25cm]{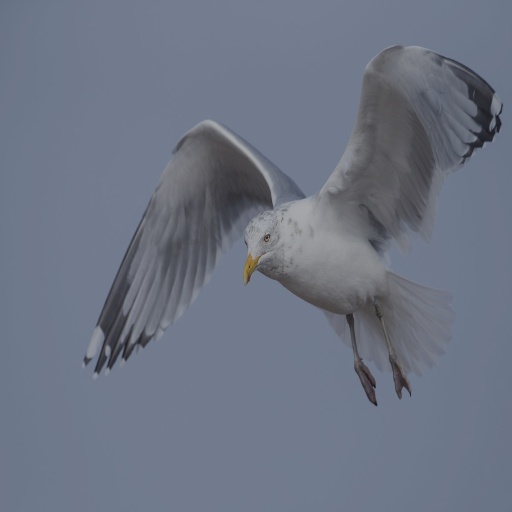}
        \caption{70\%}
    \end{subfigure}\hspace{0.1cm}%
    \begin{subfigure}[b]{1.75cm}
        \includegraphics[width=1.7cm, height=1.25cm]{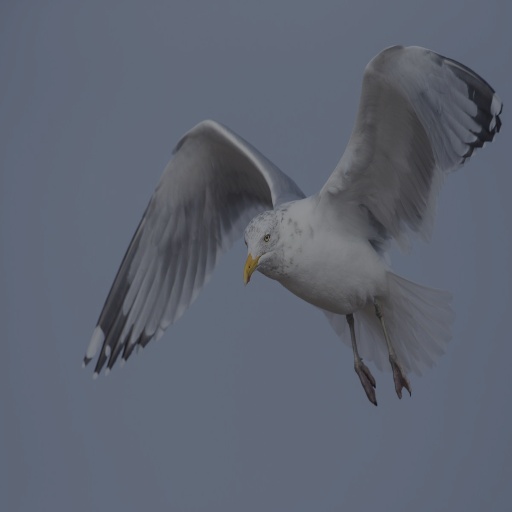}
        \caption{60\%}
    \end{subfigure}\hspace{0.1cm}%
    \begin{subfigure}[b]{1.75cm}
        \includegraphics[width=1.7cm, height=1.25cm]{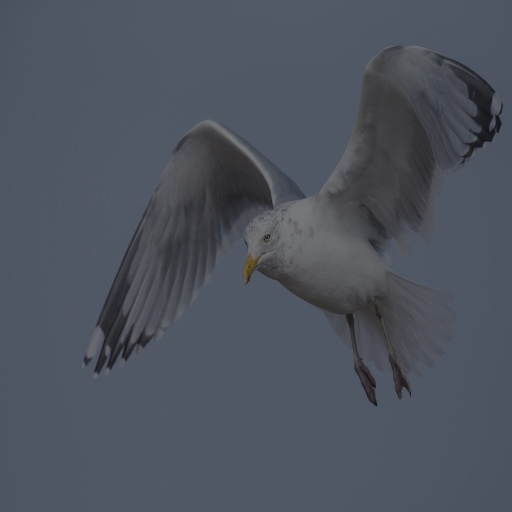}
        \caption{50\%}
    \end{subfigure}\hspace{0.1cm}%
    \begin{subfigure}[b]{1.75cm}
        \includegraphics[width=1.7cm, height=1.25cm]{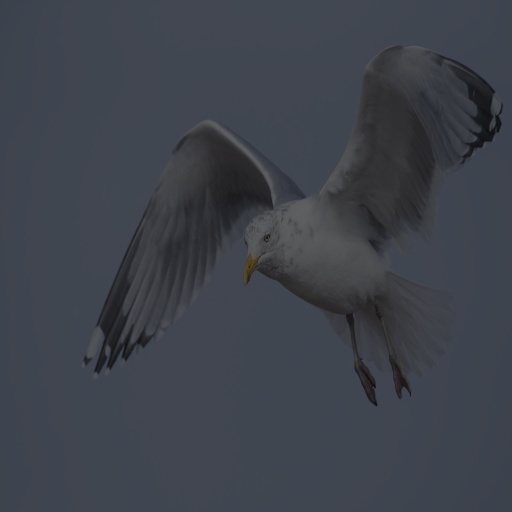}
        \caption{40\%}
    \end{subfigure}\hspace{0.1cm}%
    \begin{subfigure}[b]{1.75cm}
        \includegraphics[width=1.7cm, height=1.25cm]{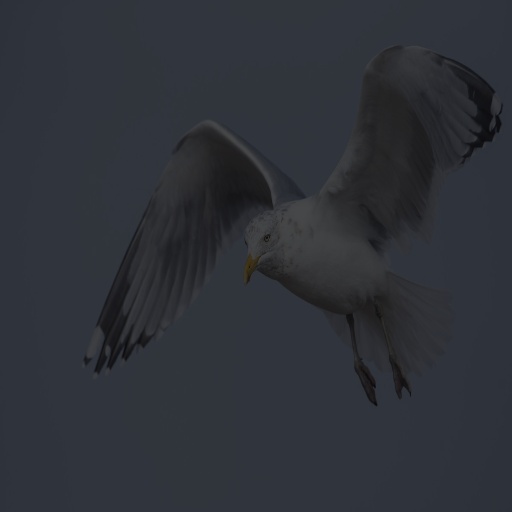}
        \caption{30\%}
    \end{subfigure}
    \caption{Darkness simulation by progressively reducing image brightness from 100\% to 30\%.}
    \label{fig:darkness_levels}
\end{figure}

\begin{figure}[H]
    \centering
        \includegraphics[width=15.5 cm, height = 5.25 cm]{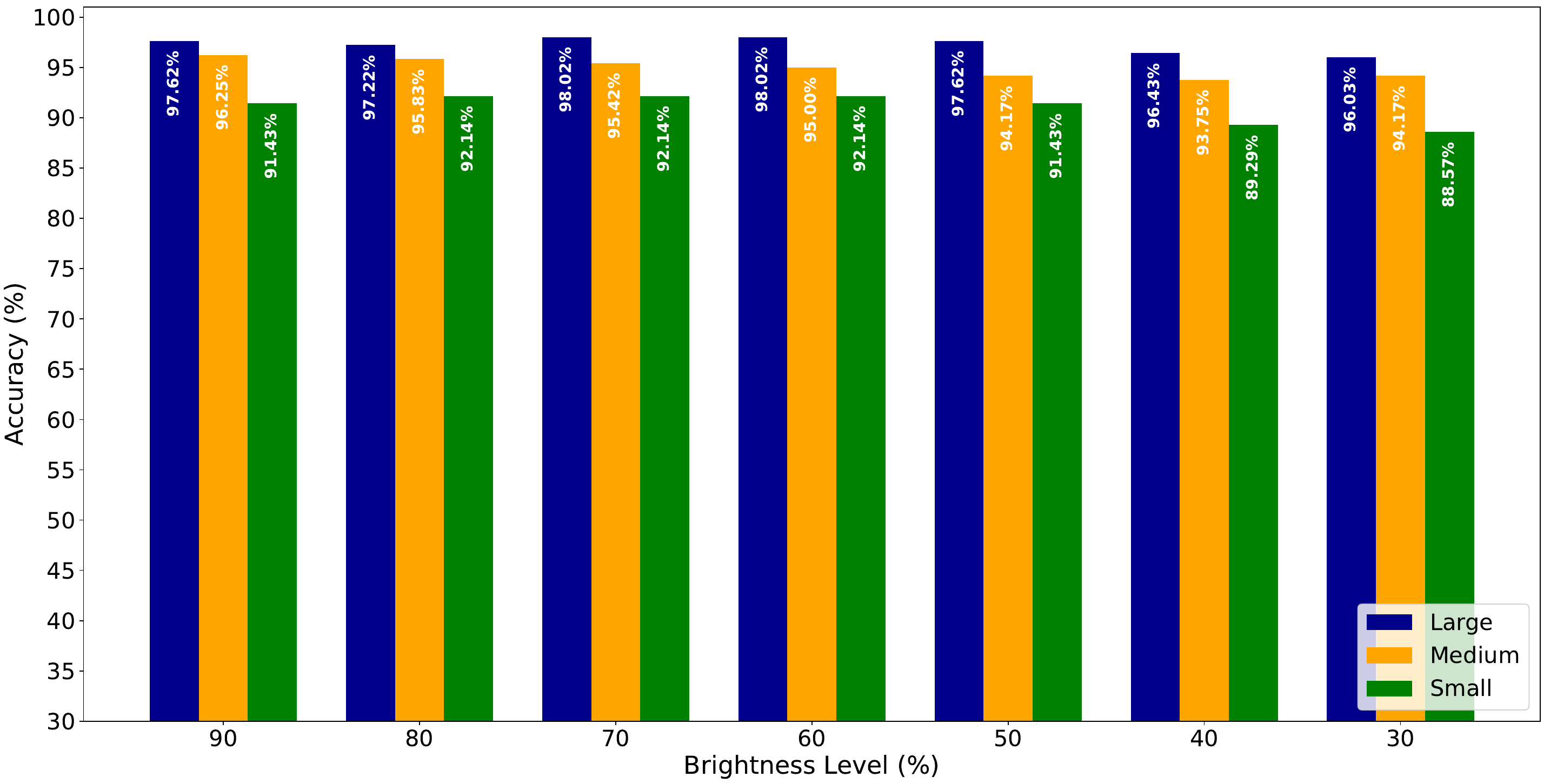}
        \caption{The effect of brightness level on the accuracy of the CNN classifiers.}
        \label{B}
\end{figure}

As seen in Fig. \ref{B}, classifier accuracy across all bird size species remained relatively robust even as brightness declined. Interestingly, the large bird classifier showed exceptional stability, maintaining accuracy above 96\% across all brightness levels tested. Medium and small bird classifiers also exhibited strong performance, with only a slight reduction as images became darker. For instance, medium bird accuracy dropped modestly from 96.25\% at 90\% brightness to 94.17\% at 30\%, while small bird accuracy decreased from 91.43\% to 88.57\%. These results suggested that the CNN models were highly resilient to mild-to-moderate reductions in brightness.
    
\end{itemize}

\subsection{The Performance of the Two-stage Classification Framework for Flock Type Identification}
\label{FTI}
\begin{figure}[H]
    \centering
        \includegraphics[width=16.5 cm, height = 8 cm]{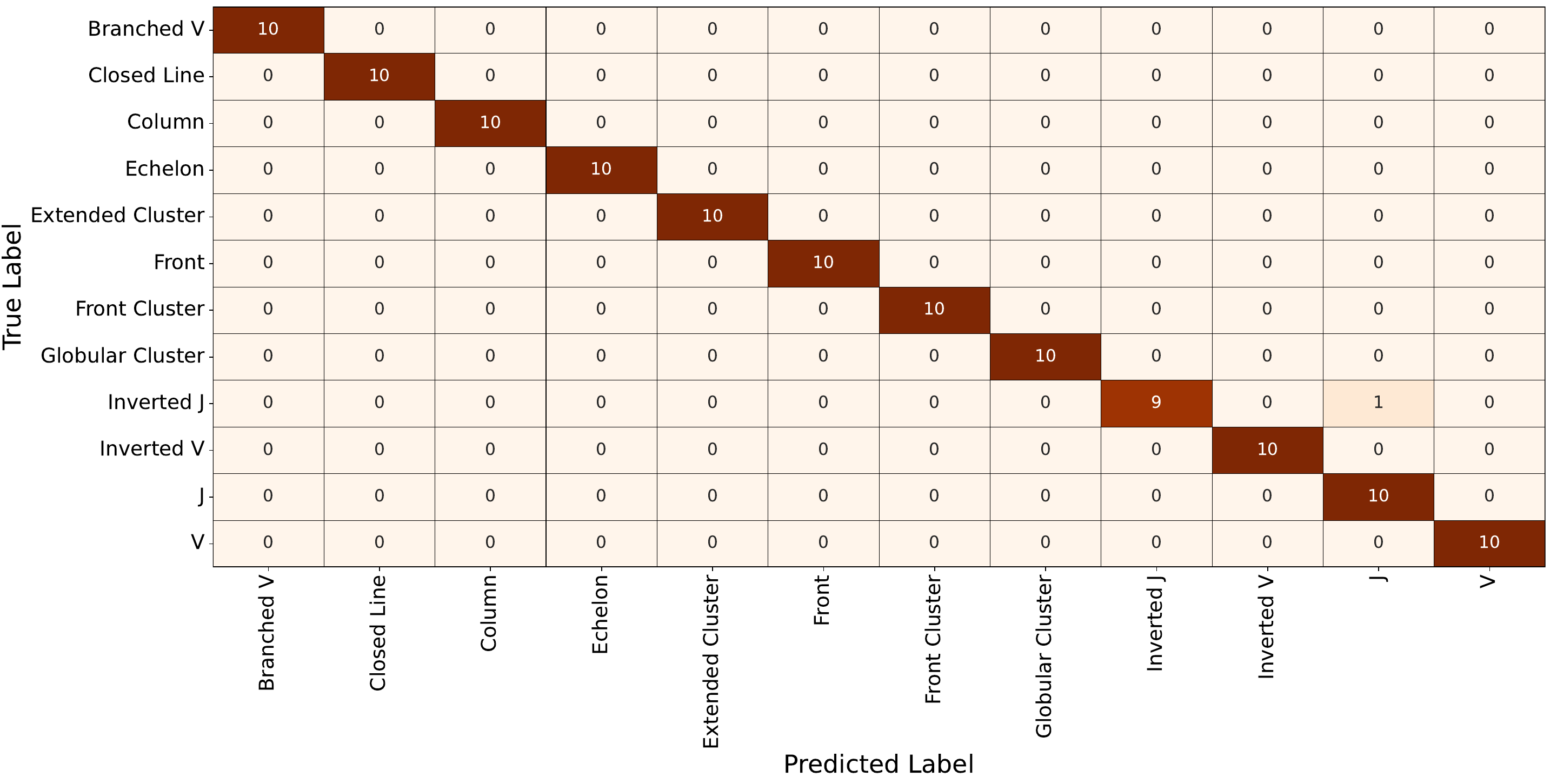}
        \caption{Confusion matrix of CNN for identifying the type of flock (Bottom view).}
        \label{CM-FlockType1}
\end{figure}

To evaluate the proposed two-stage classification framework for flock type identification, we first constructed the required image datasets. Because labeled images of bird flocks with formation annotations are not publicly available, we generated synthetic flock images using a single bird template. Specifically, for the bottom-view formation classifier (Classifier~1 in Fig.~\ref{Flock-Flowchart}), we used a bottom-view image of a pigeon and programmatically generated flock patterns corresponding to the twelve horizontal formations considered in this work. The generated images were rendered on a blue background to simulate the sky, consistent with the assumption that the sensing system can provide bottom-view imagery (e.g., from ground-based cameras). These synthetic bottom-view images were then used to train a CNN classifier based on the ResNet architecture to predict the horizontal flock formation. The confusion matrix of the first classifier (determining the flock type from bottom view) is provided in Fig.~\ref{CM-FlockType1}. In addition, the class-wise precision, recall, and F1-score for each class are reported in Table~\ref{Tab-FlockType1}.

\renewcommand{\arraystretch}{0.6}
\begin{table}[H]
\centering
\caption{Class-wise precision, recall, and F1-score of CNN for identifying birds' flock type.}
\begin{tabular}{lccc}
\toprule
\textbf{Class} & \textbf{Precision} & \textbf{Recall} & \textbf{F1-score} \\
\midrule
Branched V          & 1.00 & 1.00 & 1.00 \\
Closed Line         & 1.00 & 1.00 & 1.00 \\
Column              & 1.00 & 1.00 & 1.00 \\
Echelon             & 1.00 & 1.00 & 1.00 \\
Extended Cluster    & 1.00 & 1.00 & 1.00 \\
Front               & 1.00 & 1.00 & 1.00 \\
Front Cluster       & 1.00 & 1.00 & 1.00 \\
Globular Cluster    & 1.00 & 1.00 & 1.00 \\
Inverted J          & 1.00 & 0.90 & 0.95 \\
Inverted V          & 1.00 & 1.00 & 1.00 \\
J                   & 0.91 & 1.00 & 0.95 \\
V                   & 1.00 & 1.00 & 1.00 \\
\midrule
\textbf{Accuracy} & \multicolumn{3}{c}{\textbf{99.17\%}} \\
\bottomrule
\end{tabular}
\label{Tab-FlockType1}
\end{table}

\begin{figure}[H]
    \centering
        \includegraphics[width=16.5 cm, height = 8 cm]{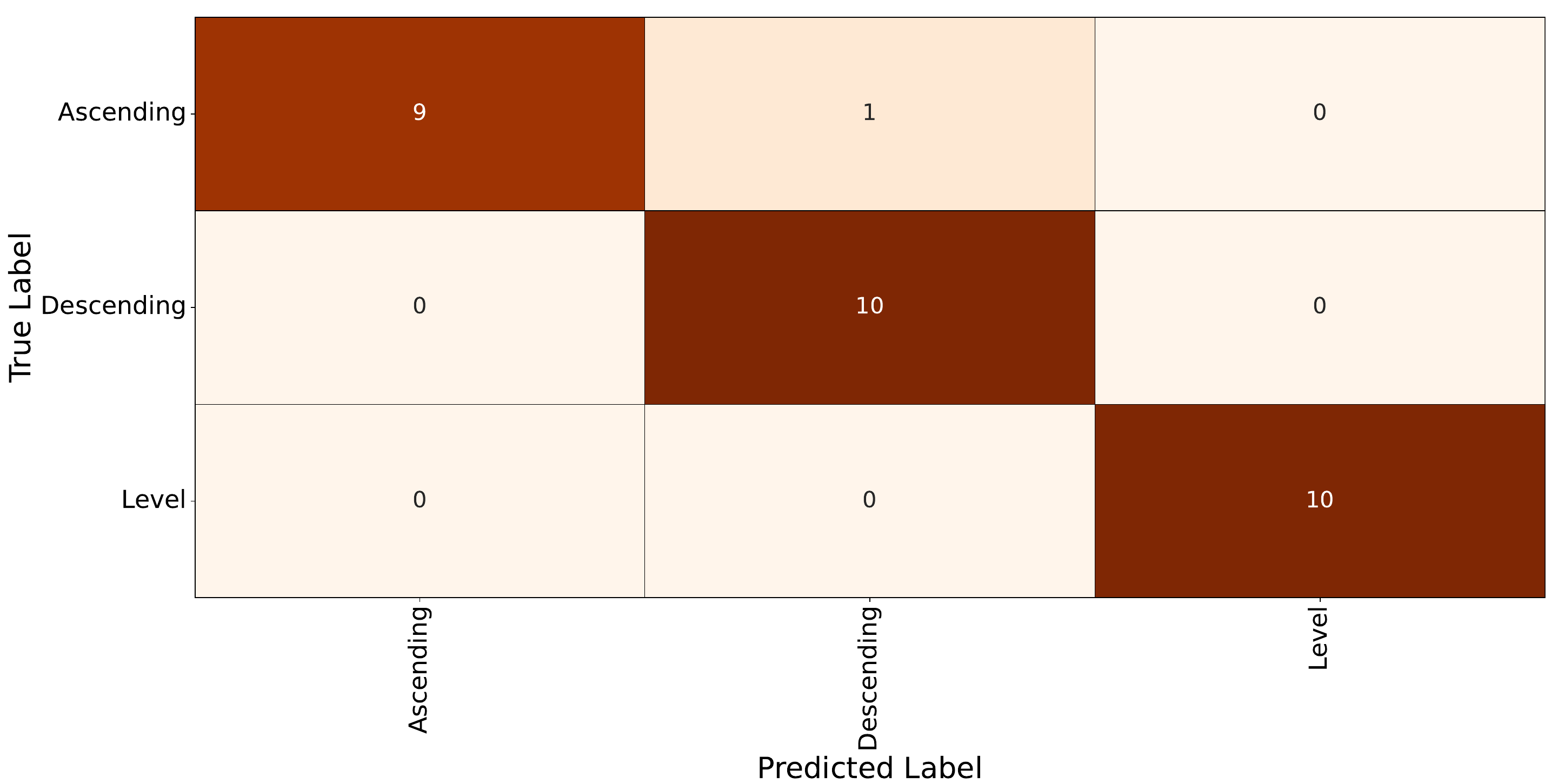}
                \caption{Confusion matrix of CNN for identifying the type of flock (Side view).}
        \label{CM-FlockType2}
\end{figure}

For the second stage, which is only activated when the formation is classified as a Column, we generated a separate dataset using a side-view image of a pigeon. Using this template, we created Column-formation images with three vertical alignment labels: Ascending, Descending, and Level (Classifier~2 in Fig.~\ref{Flock-Flowchart}). A ResNet-based CNN was then trained to predict the vertical alignment from side-view inputs. The confusion matrix of the second classifier (determining the vertical alignment from side view) is provided in Fig.~\ref{CM-FlockType2}. In addition, the class-wise precision, recall, and F1-score for each class are reported in Table~\ref{tab:FlockTypeSide}.

\renewcommand{\arraystretch}{0.6}
\begin{table}[H]
\centering
\caption{Class-wise precision, recall, and F1-score of CNN for identifying flock type (side view).}
\begin{tabular}{lccc}
\toprule
\textbf{Class} & \textbf{Precision} & \textbf{Recall} & \textbf{F1-score} \\
\midrule
Ascending   & 1.00 & 0.90 & 0.95 \\
Descending  & 0.91 & 1.00 & 0.95 \\
Level       & 1.00 & 1.00 & 1.00 \\
\midrule
\textbf{Accuracy} & \multicolumn{3}{c}{\textbf{96.67\%}} \\
\bottomrule
\end{tabular}
\label{tab:FlockTypeSide}
\end{table}

Because the proposed framework operates in a conditional two-stage manner, the overall system accuracy depends on the class-wise behavior of the first-stage classifier. For all non-Column formations, the final prediction is determined solely by the bottom-view classifier. However, for Column formations, a correct end-to-end prediction requires both correct detection of the Column class in the first stage and correct identification of the vertical alignment in the second stage. Therefore, the effective accuracy for Column samples is given by the product of the Column recall (1.00) in the first-stage classifier and the accuracy of the second-stage classifier (96.67\%).

\subsection{The Performance of the Flock Size Classification}
\label{FSC}
\begin{figure}[H]
    \centering
        \includegraphics[width=14.5 cm, height = 7 cm]{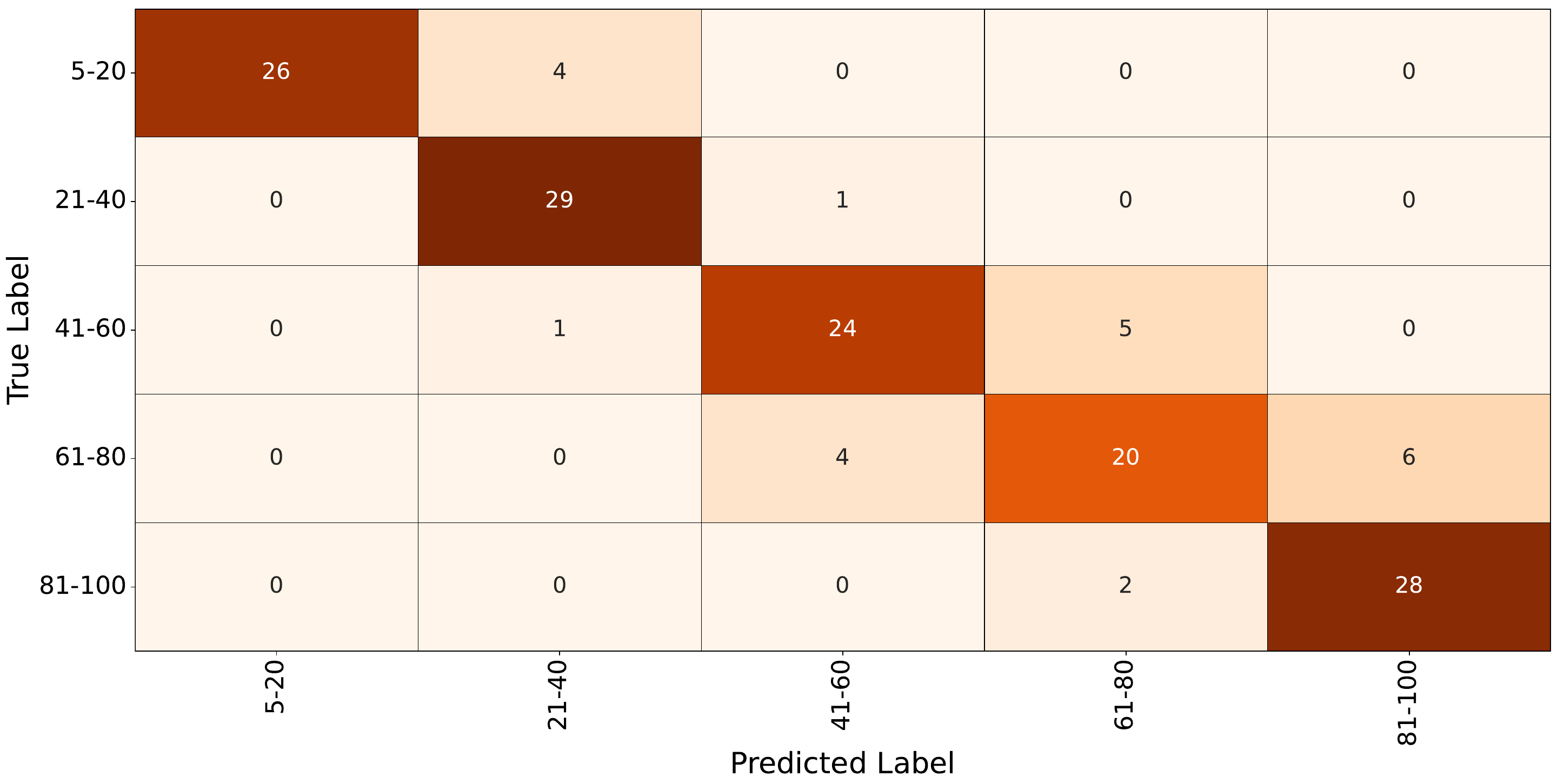}
                \caption{Confusion matrix of CNN for identifying the flock size.}
        \label{Flock-Size}
\end{figure}

In addition to flock type identification, we evaluated flock size classification using the same synthetic image generation process described in Section~\ref{FTI}. During dataset construction, each generated flock image was labeled according to the number of birds present. These labels were then mapped into five flock size categories: 5--20 birds, 21--40 birds, 41--60 birds, 61--80 birds, and 81--100 birds. Using this labeled dataset, we trained a multi-class CNN classifier (ResNet-based) to predict the flock size category from an input image. The confusion matrix of the classifier to specify the size of flock is provided in Fig.~\ref{Flock-Size}. In addition, the class-wise precision, recall, and F1-score for each class are reported in Table~\ref{Tab-FlockSize}.

\renewcommand{\arraystretch}{0.6}
\begin{table}[H]
\centering
\caption{Class-wise precision, recall, and F1-score of CNN for identifying flock size ranges.}
\begin{tabular}{lcccc}
\toprule
\textbf{Flock Size Range} & \textbf{Precision} & \textbf{Recall} & \textbf{F1-score} & \textbf{Support} \\
\midrule
5--20     & 1.00 & 0.87 & 0.93 & 30 \\
21--40    & 0.85 & 0.97 & 0.91 & 30 \\
41--60    & 0.83 & 0.80 & 0.81 & 30 \\
61--80    & 0.74 & 0.67 & 0.70 & 30 \\
81--100   & 0.82 & 0.93 & 0.88 & 30 \\
\midrule
\textbf{Accuracy} & \multicolumn{4}{c}{\textbf{85.00\%}} \\

\bottomrule
\end{tabular}
\label{Tab-FlockSize}
\end{table}

\section{Insights from Experimental Evaluations}
\label{Discussion}
This section highlights the key findings and lessons learned from our experiments on bird species classification using CNN-based models. Through a series of systematic evaluations, we explored how different architectural choices, dataset sizes, and image size influence model performance. The following insights offer practical guidance for optimizing deep learning models in bird classification tasks.

\begin{figure}[H]
    \centering
        \includegraphics[width=10.5cm, height = 5 cm]{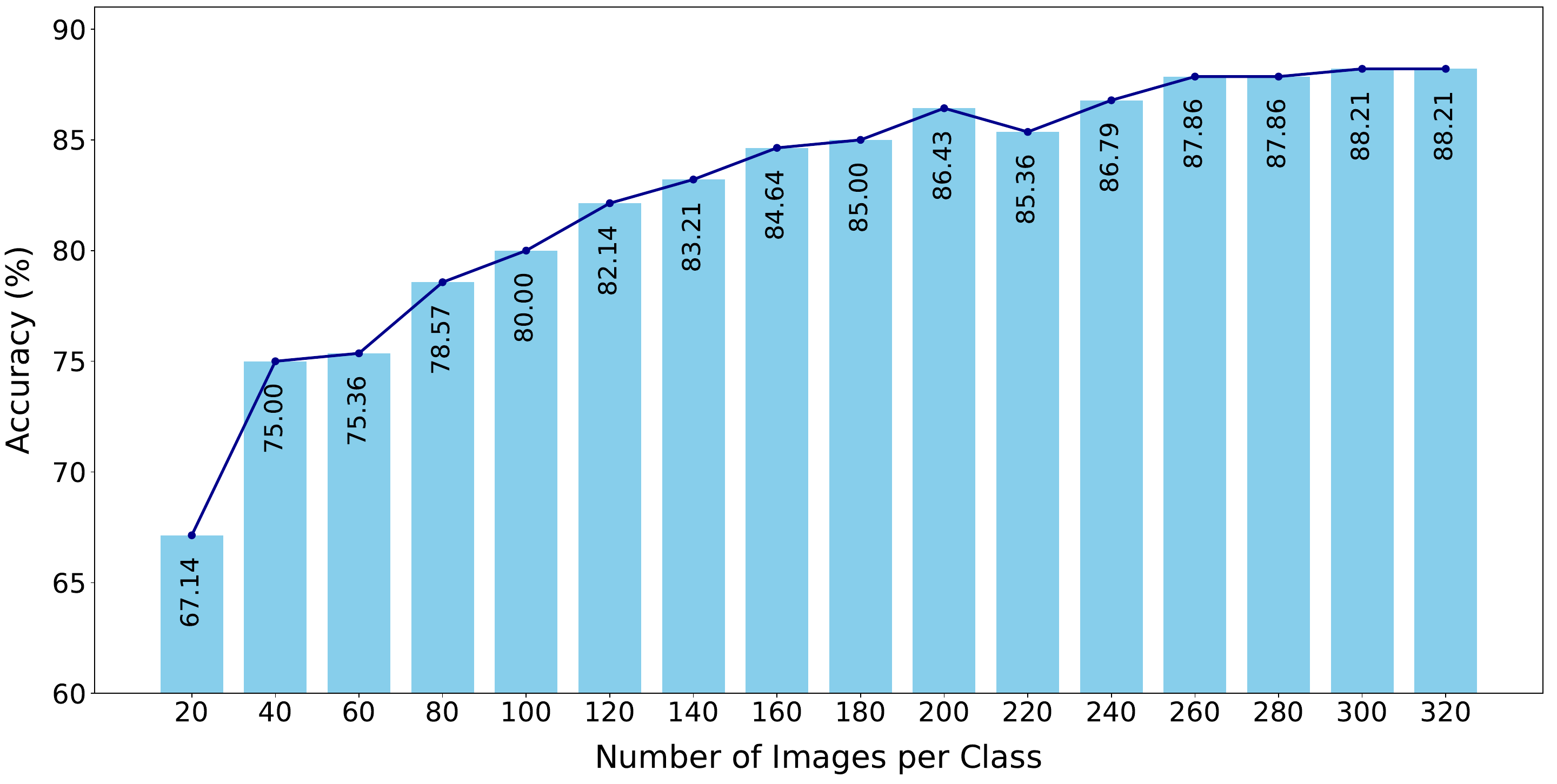}
        \caption{The effect of training and validation data size per class on the accuracy of the CNN classifier.}
        \label{N-I}
\end{figure}

\noindent\textbf{Insight 1: }We observed a general trend indicating that increasing the number of training images per class tends to improve the classification accuracy of the model. While the accuracy does not always increase linearly or consistently with more data, sometimes decreasing or plateauing across certain training sizes, the overall pattern shows that having more examples per class typically leads to better performance. This suggests that additional data helps the CNN learn more representative and discriminative features. To assess this effect, we conducted a series of controlled experiments using a fixed test set of 40 images per class across all runs. For each experiment, we varied the number of training and validation images per class, starting from 20 and increasing incrementally (e.g., 40, 60, etc.). The dataset for each class was sorted to ensure consistent image selection, and 90\% of the data was used for training, with the remaining 10\% for validation. We maintained the same model architecture and hyperparameters throughout to isolate the impact of training data quantity. As shown in Fig \ref{N-I}, the overall trend demonstrates that more training data per class generally contributes to higher model accuracy.

\noindent\textbf{Insight 2: }We noticed that tuning a ResNet model is more difficult than working with a custom CNN. While we can still try different parameters such as learning rate, number of fine-tuned layers, or data augmentation, the accuracy usually does not improve much. ResNet has a fixed structure and pretrained layers, which help with strong initial results but limit flexibility. This means that even though ResNet performs well out of the box, getting noticeably better accuracy is challenging.

\noindent \textbf{Insight 3:} We investigated the impact of using grayscale images versus RGB (Red, Green, Blue) images on classification accuracy across various input sizes. To enable ResNet, which is originally designed for 3-channel RGB inputs, to accept grayscale images, we duplicated the single grayscale channel three times to simulate an RGB-like format. The results of our experiments, summarized in Table \ref{tab:grayscale_rgb_comparison}, revealed that while grayscale images can achieve competitive results, RGB images generally offer slightly better performances. Overall, these findings suggest that grayscale images can be a viable alternative when pretrained models are used and color information is limited. When maximum accuracy is desired, especially in tasks where color conveys meaningful patterns, RGB inputs remain preferable.

\renewcommand{\arraystretch}{0.5}
\begin{table}[H]
\centering
\caption{Comparison of the CNN's accuracy using grayscale and RGB images across different input sizes.}
\begin{tabular}{lcc}
\toprule
\textbf{Size of Image} & \textbf{Grayscale Accuracy (\%)} & \textbf{RGB Accuracy (\%)} \\
\midrule
64×64     & 61.11  & 63.10  \\
128×128   & 81.35  & 81.75  \\
180×180   & 90.08  & 90.48  \\
224×224   & 90.08  & 90.48  \\
256×256   & 92.46  & 94.84  \\
299×299   & 93.25  & 95.24 \\
320×320   & 93.25  & 94.05 \\
384×384   & 95.63  & 96.83  \\
512×512   & 97.22  &  97.62 \\
\bottomrule
\end{tabular}
\label{tab:grayscale_rgb_comparison}
\end{table}

\begin{figure}[H]
    \centering
        \includegraphics[width=15.5 cm, height = 5 cm]{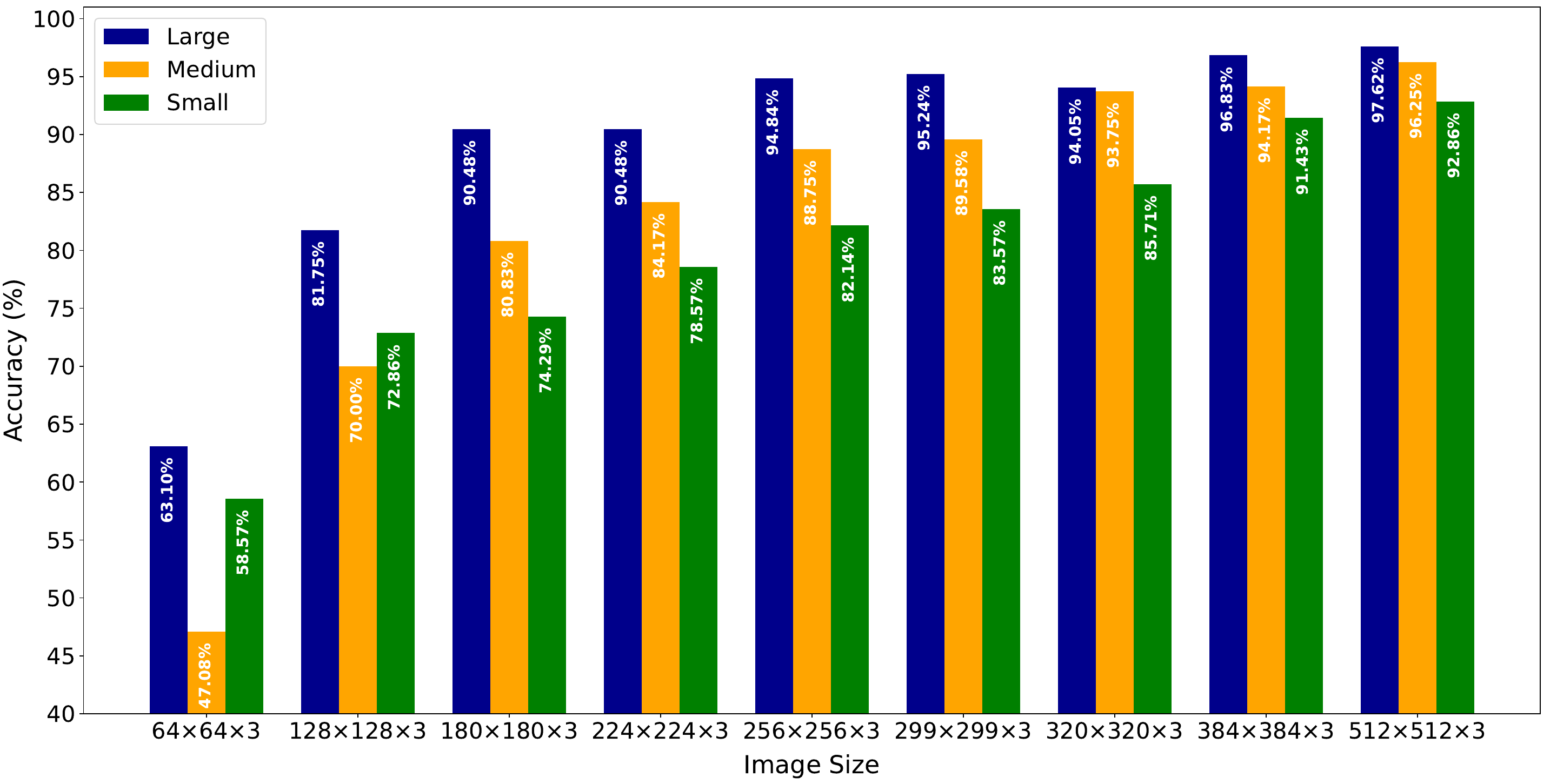}
        \caption{The effect of image size on the accuracy of the CNN classifiers.}
        \label{S-I}
\end{figure}

\noindent\textbf{Insight 4: }We found that the input image size has a noticeable impact on both training time and classification accuracy. Using smaller image dimensions, such as 64×64, significantly speeds up training. However, this often comes at the cost of lower accuracy, likely due to the limited amount of visual detail retained at smaller resolutions. As we increased the image size to 128×128, 180×180, 224×224, 256×256, 299×299, 320×320, 384×384, and 512×512 we observed a general improvement in model accuracy, as shown in Fig. \ref{S-I}. This suggests that higher-resolution images provide richer and more discriminative features for the CNN to learn from, but with longer training times. The trade-off between speed and performance highlights the importance of choosing an appropriate image size based on the specific goals and resource constraints of a project. Also, an interesting observation from our experiments is that the optimal input image size required to achieve a target accuracy (e.g., 90\%) varies with the physical size of the bird species. For large-sized birds, a resolution of at least 180×180×3 was sufficient to reach 90.48\% accuracy. However, medium-sized birds required images of 320×320×3 to surpass 90\% accuracy (93.75\%), while small-sized birds needed even higher resolution images of 384×384×3 to achieve a comparable accuracy (91.43\%). This trend indicates that the smaller the bird, the higher the image resolution required for accurate classification, likely due to the finer details needed to distinguish smaller species.

\section{Conclusion}
\label{Conclusion}

In this study, we explored two different approaches for classifying bird species and aircraft: the CCA and the UCA. The CCA divides the task into three logical steps: bird detection, size classification, and species classification, while the UCA uses a single end-to-end model to classify all bird species and aircraft in one stage. Our results show that the UCA achieves slightly higher overall accuracy, while the CCA still performs well but its final accuracy depends on how well all five models work together. In addition to comparing the two classification strategies, we evaluated the performance of the CNN models against three traditional machine learning methods: SVM, RF, and KNN. This comparison was performed for the species classifiers used in the CCA pipeline (classifier 3 for large birds, classifier 4 for medium birds, and classifier 5 for small birds). For all three classifiers (classifiers 3, 4, 5), CNN consistently outperformed the benchmark models, demonstrating its superior ability to handle complex image classification tasks. We also assessed the robustness of the CNN-based classifiers (classifiers 3, 4, and 5) under challenging conditions, including sensor noise, snow, rain, and low-light (darkness). The results indicate that while performance does degrade under severe conditions, the CNN models maintain relatively high accuracy, showing strong resilience to various real-world disturbances. Overall, our findings support the use of CNNs for effective and robust classification of bird species and aircraft, which can contribute to improving bird strike prevention systems in aviation.
In addition to species-level classification, this work also implemented CNN-based framework for bird flock type identification and flock size categorization, enabling the system to capture higher-level collective flight characteristics beyond individual bird recognition. The performance of these models demonstrates that deep learning approaches can reliably identify both geometric flock formations and size-related attributes from visual data. It should be noted that the flock type and flock size classifiers were trained and evaluated on synthetically generated images due to the lack of publicly available labeled datasets for bird flock formations. While this approach enables controlled evaluation of formation geometry, future work will focus on validating and fine-tuning these models using real-world aerial imagery of bird flocks to assess generalization performance. Future advancements in bird strike prevention systems may include the ability to detect the precise location of birds within each image.

\end{document}